\pdfoutput=1

\documentclass[11pt]{article}
\usepackage{soul}
\usepackage[normalem]{ulem}
\usepackage{EMNLP2023} 

\usepackage{times}
\usepackage{latexsym}
\usepackage{makecell}
\usepackage{float}

\usepackage[T1]{fontenc}

\usepackage[utf8]{inputenc}

\usepackage{microtype}

\usepackage{inconsolata}
\usepackage{listings}
\usepackage{graphicx}               
\usepackage{tabularx}               
\usepackage{booktabs}
\usepackage{amsmath}
\usepackage{stmaryrd}
\usepackage{xcolor}
\usepackage{soul}
\usepackage[linesnumbered,ruled,vlined]{algorithm2e}
\usepackage{multirow}               
\usepackage{diagbox}                
\usepackage{hhline}                 
\usepackage{color, colortbl}                  
\usepackage{amsmath}                
\usepackage{amssymb}                
\usepackage{pifont}
\usepackage{mathtools}              
\usepackage{enumitem}               

\usepackage{subfigure}
\usepackage{booktabs}
\usepackage{extarrows}
\usepackage{makecell}

\newcommand{\jw}[1]{\textcolor{purple}{[\textbf{jack:} #1]}}
\NewDocumentCommand{\zihao}{ mO{} }{\textcolor{red}{\textsuperscript{\textit{zihao}}\textsf{\textbf{\small[#1]}}}}

\usepackage{listings}
\usepackage{xcolor}

\definecolor{codegreen}{rgb}{0,0.6,0}
\definecolor{codegray}{rgb}{0.5,0.5,0.5}
\definecolor{codepurple}{rgb}{0.58,0,0.82}
\definecolor{backcolour}{rgb}{0.95,0.95,0.92}

\lstdefinestyle{mystyle}{
    backgroundcolor=\color{backcolour},   
    commentstyle=\color{codegreen},
    keywordstyle=\color{magenta},
    numberstyle=\tiny\color{codegray},
    stringstyle=\color{codepurple},
    basicstyle=\ttfamily\footnotesize,
    breakatwhitespace=false,         
    breaklines=true,                 
    captionpos=b,                    
    keepspaces=true,                 
    numbers=left,                    
    numbersep=5pt,                  
    showspaces=false,                
    showstringspaces=false,
    showtabs=false,                  
    tabsize=2
}

\title{Persona-SQ: A Personalized  Suggested Question Generation Framework For Real-world Documents}

\author{Zihao Lin$^1$\thanks{\hspace{5pt}Work done during an internship at Adobe.}, Zichao Wang$^2$, Yuanting Pan$^3$, Varun Manjunatha$^2$ \\ {\bf Ryan Rossi$^2$, Angela Lau$^2$, Lifu Huang$^1$, Tong Sun$^2$} \\
$^1$UC Davis \hspace{10pt} $^2$Adobe \hspace{10pt} $^3$Stanford University\\
\texttt{qzlin@ucdavis.edu} \hspace{10pt} \texttt{jackwa@adobe.com}\\
}

\begin{document}
\maketitle
\begin{abstract}
Suggested questions (SQs) provide an effective initial interface for users to engage with their documents in AI-powered reading applications. In practical reading sessions, users have diverse backgrounds and reading goals, yet current SQ features typically ignore such user information, resulting in homogeneous or ineffective questions. We introduce a pipeline that generates personalized SQs by incorporating reader profiles (professions and reading goals) and demonstrate its utility in two ways: 1) as an improved SQ generation pipeline that produces higher quality and more diverse questions compared to current baselines, and 2) as a data generator to fine-tune extremely small models that perform competitively with much larger models on SQ generation. Our approach can not only serve as a drop-in replacement in current SQ systems to immediately improve their performance but also help develop on-device SQ models that can run locally to deliver fast and private SQ experience.

\end{abstract}

\section{Introduction}
Large language models (LLMs) have shown strong promise as document assistants to help users better read and understand their content in the form of AI-powered reading software and applications such as ChatPDF,\footnote{\footnotesize\url{www.chatpdf.com}} NotebookLM,\footnote{\footnotesize\url{notebooklm.google.com}} and Acrobat's AI Assistant.\footnote{\footnotesize\url{www.adobe.com/acrobat/generative-ai-pdf}}
One of the core features of these AI-powered reading applications is automatically generating suggested questions (SQs)~\cite{Wang2019,Huang2023}. These questions are among the first features that users see when they first upload a document and have the potential to help improve user engagement~\cite{Cox2019,10382515}, and guide the user to more effectively navigate documents~\cite{2312.15820}, ultimately leading to improved productivity. These automatically generated SQs could also relieve users from manually typing questions they want to ask, resulting in a more effortless interaction~\cite{Sarwar2020}.

Typically, users with different backgrounds and interests may possess distinct goals and information-seeking needs, even when reading the same document. Ideally, for different users, the AI-powered reading applications would tailor the generated SQs to their backgrounds and needs. Unfortunately, the current SQ feature across reading applications relies mostly on the document as the anchor for generating document-relevant SQs but largely ignores information about users themselves. One challenge lies in the difficulty of obtaining such user profile information during reading, likely because of privacy considerations and because user activities, from which we can draw inferences about the user, are difficult to track and record, especially when the document is in the form of a PDF. As a result, without user information, the generated SQs may appear homogeneous, repetitive, and ineffective. These observations and challenges motivate our work: how to personalize the generated SQs to tailor to the backgrounds and reading goals of different individuals, especially with the absence of user profile information?

\begin{figure*}[t!]
    \centering
    \includegraphics[width=\textwidth]{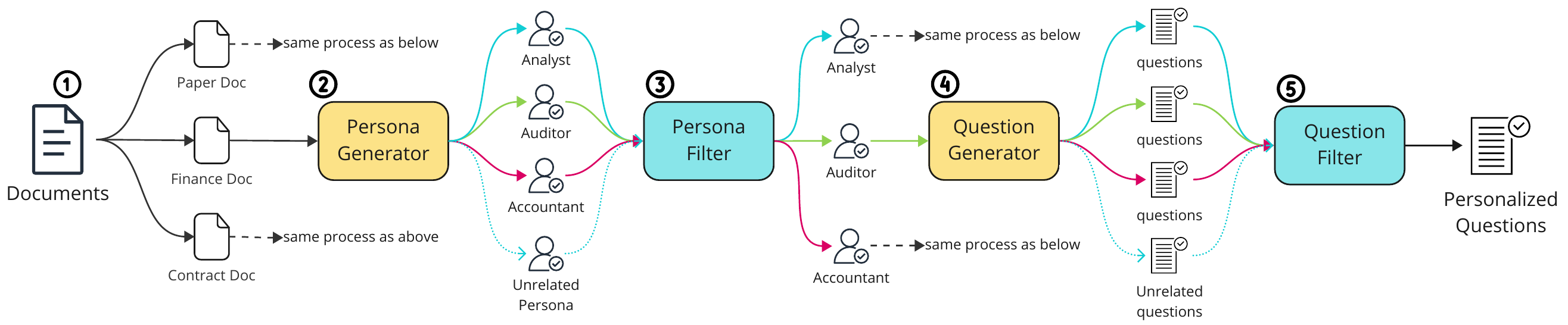}
    \caption{An illustration of \textbf{Persona-SQ}, our personalized suggested question generation pipeline.}
    \label{fig:sq-gen}
\end{figure*}

\subsection{Contributions}
We situate our work in the context of reading in professional work environments and investigate persona-based SQ generation by synthetically injecting user profile information, in the form of their profession and reading goals, into the SQ generation process. We propose a simple framework, \textbf{Persona-SQ} for such a synthetic persona-based SQ generation system, where an LLM first generates a few user profiles and then generates SQs for each user profile. Each generation stage has a scoring process to retain only high-quality and relevant generated profiles and questions.

We validate our approach by generating SQs from three sets of public documents drawn from diverse domains including finance, legal, and academia. Various metrics, including human evaluations and our newly designed diversity metrics, show that our Persona-SQ system instantiated using GPT4o \cite{openai2024gpt4technicalreport}, consistently produces more diverse and higher quality SQs than the GPT4o baseline that generates SQs without using persona information. This encouraging result implies that our method has the potential to be a simple drop-in upgrade to improve existing SQ generators when they are implemented using powerful LLMs through API calls. 

We further showcase the utility of Persona-SQ by instantiating it with an open-source model (\texttt{Llama-3.1-70B} \cite{dubey2024llama}). We utilize it to curate a large synthetic SQ dataset with 100k questions from thousands of diverse, real-world documents which we then use to fine-tune very tiny models of only 360 million parameters for the task of SQ generation. On both automatic and human evaluations, we demonstrate that models fine-tuned using the Persona-SQ dataset outperform models fine-tuned on an SQ dataset without persona and on public question datasets by a large margin, and are a strong contender to their much larger counterparts such as GPT4o. These small models have the potential to be deployed on the user's end device, delivering a fast and private SQ experience when reading documents without API calls and without the document leaving the user's device. 

\section{Persona-SQ Framework}
\label{sec:persona-sq-framework}
\begin{figure*}[!ht]
    \centering
    \subfigure[lawyers]{
        \includegraphics[width=0.30\textwidth]{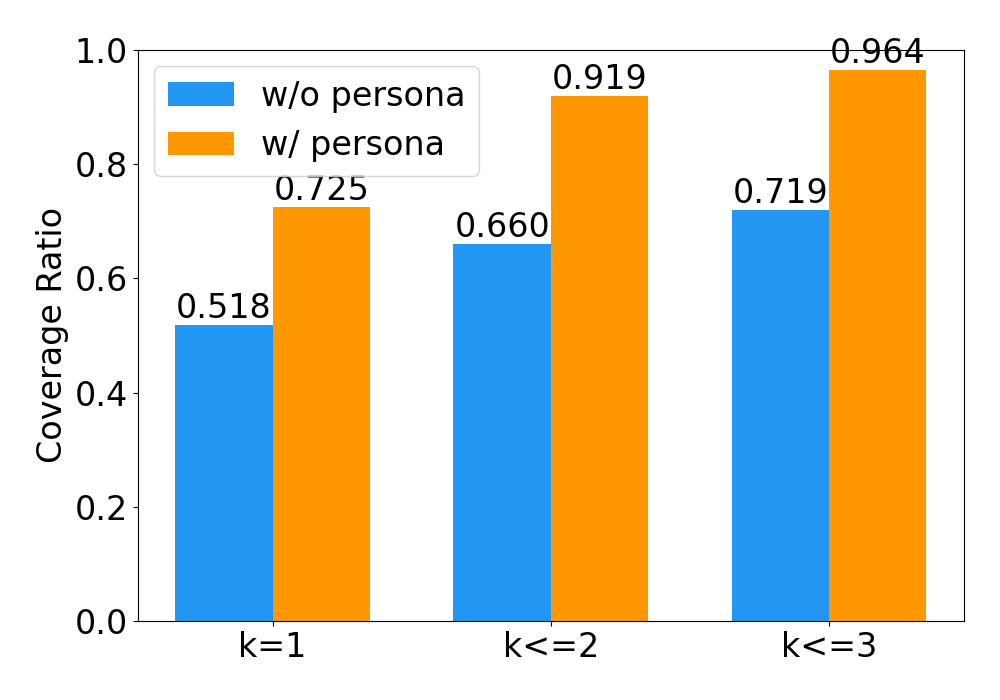}
    }
    \hfill  
    \subfigure[patent attorneys]{
        \includegraphics[width=0.30\textwidth]{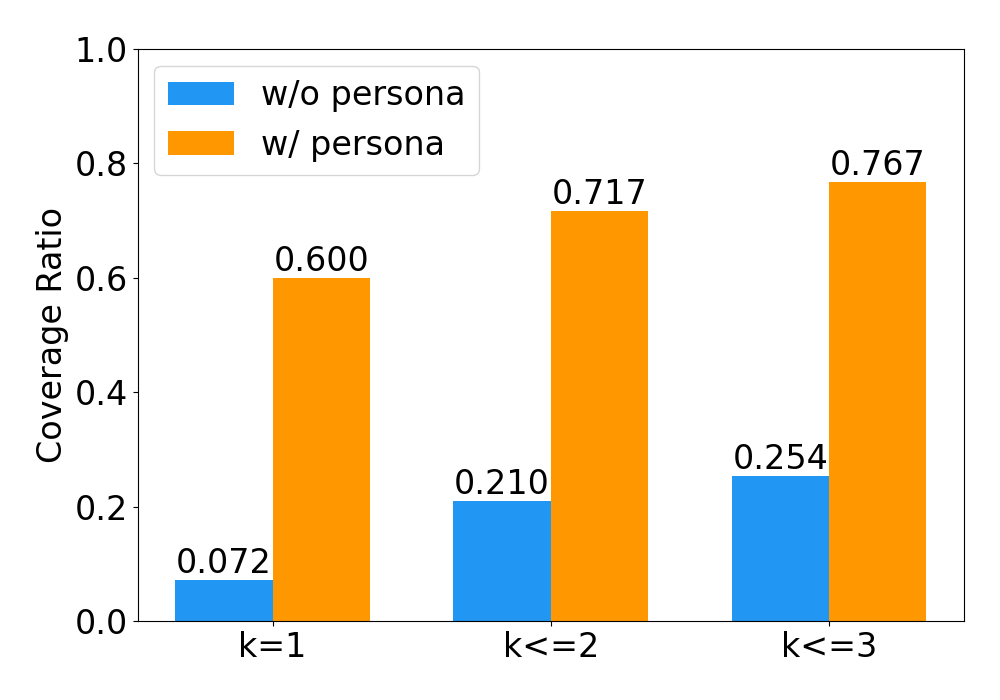}
    }
    \hfill
    \subfigure[finance managers]{
        \includegraphics[width=0.30\textwidth]{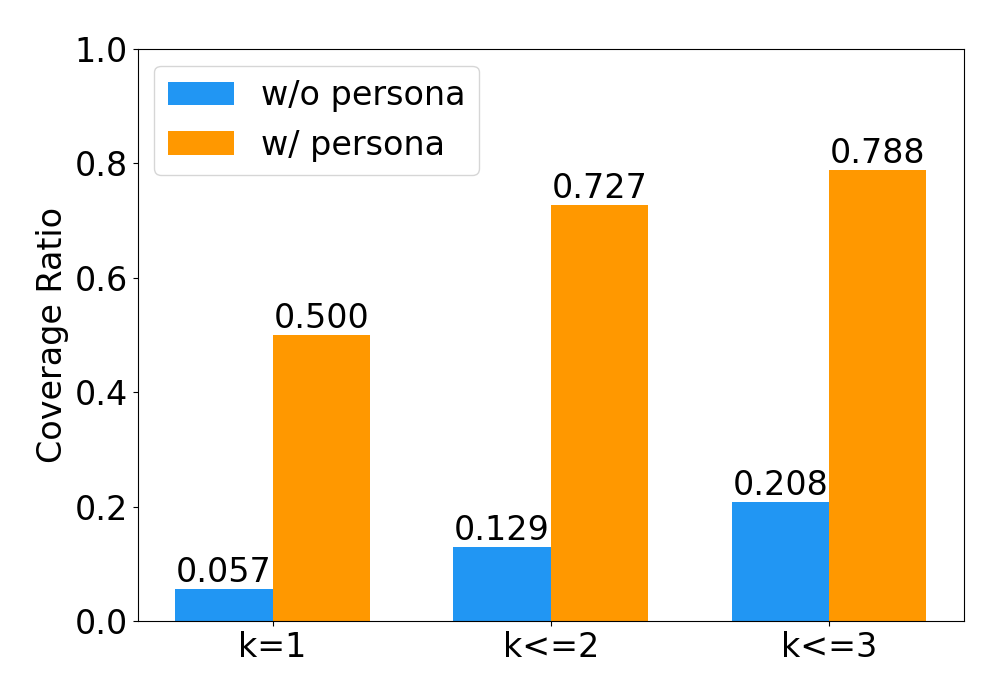}
    }
    \vspace{-8pt}
    \caption{Examples of the persona coverage ratio (legal). The higher scores of SQs generated with persona compared to those generated without persona indicate the personalized SQs are more aligned to the intended personas. 
    }
    \vspace{-5pt}
    \label{fig:cover-ratio}
\end{figure*}

\label{sec:data-generation-pipeline}
We now introduce \textbf{Persona-SQ}, our approach to generate personalized SQs, which is illustrated in Figure \ref{fig:sq-gen}. Persona-SQ consists of five steps: 
document collection, persona generation, question formulation, and robust quality control mechanisms for filtering suboptimal personas and questions. We show how to use our pipeline in Appendix \ref{sec:utilize-persona-sq}.

\paragraph{Step 1: Collect Documents.} We compile sets of open-source documents from public websites and datasets. For a given domain $d$, we denote the corresponding document set as $\mathcal{D}_d = \{ D_{1}, ..., D_{U}\}$, where $U$ is the total number of documents. 

\paragraph{Step 2: Generate Professions and Goals.} For each document $D_i$ within domain $d$, we employ an LLM to generate relevant professional roles $p$ and their corresponding sets of five objectives ${g_1, ..., g_5}$. These objectives represent the potential goals that professionals might aim to achieve through their inquiries. For instance, in the financial domain, the generated personas include ``investors'' seeking to ``evaluate the company's operational performance and profitability,'' and ``regulators'' aiming to ``assess potential future corporate risks.'' This process results in a comprehensive pool of profession-goal pairs for each domain. The specific prompt template used for generating these professions and their associated goals is detailed in Table \ref{tab:prompt-generation}.

\paragraph{Step 3: Quality Control for Professions and Goals.} We implement two distinct strategies to ensure the quality of generated personas.
First, we normalize the profession pool by utilizing the LLM to consolidate overlapping personas generated from different documents within the same domain. For instance, variations such as ``accountants'' and ``financial accountants'' are unified under the single persona ``accountant''. The specific prompt for this consolidation process is presented in Table \ref{tab:prompt-generation}. Subsequently, we aggregate the goals associated with each normalized profession. For each domain $d$, we establish a dictionary of persona-goals pairs, denoted as $\mathcal{H}_d = \{p^1: [g_1^{1}, ..., g_n^{1}], ..., p^m: [g_1^{m}, ..., g_n^{m}] \}$, where $m$ represents the total number of personas in domain $d$, and $n$ denotes the number of goals per domain. For simplification, we ignore the subscript $d$ for personas and goals. 
Second, we implement a quality assessment mechanism for the generated goals associated with each persona. We evaluate each goal on a scale from 1 to 5, based on its relevance to the corresponding persona. Higher scores indicate a greater likelihood that the persona would pursue that goal when reading the document. Goals scoring below 4 are eliminated from the pool. The detailed scoring criteria and associated prompts are documented in Table \ref{tab:prompt-persona-quality-control}. Following this filtration process, we randomly select five goals from the refined goal pool for each persona-document pair to facilitate personalized question generation.

\paragraph{Step 4: Generate Personalized Questions.} For each document, we generate various personalized questions $q$ according to the persona and goals. We can formulate the process as the following equation: $\{q_1^{i}, ..., q_{t_i}^{i}\} = \text{LLM}(P_{gen}, p^i, [g_1^{1}, ..., g_5^{1}])$, where $t_i$ is the number of generated personalized questions giving persona $p_i$. The prompt for generation $P_{gen}$ is shown in Table \ref{tab:prompt-generation}.

\paragraph{Step 5: Quality Control for Personalized Questions.} We implement three quality control mechanisms to ensure the validity of generated questions:
\begin{itemize}
\item Question Length Assessment: We establish length constraints by filtering out questions containing fewer than 5 tokens or exceeding 100 tokens to maintain optimal question complexity and clarity.
\item Quality Evaluation: We employ an LLM-based multi-dimensional scoring system (scale 1-5) based on two critical criteria: (1) relevance between SQs and the given persona with goals, and (2) relevance between SQs and the given document. The generated SQs whose scores are below 4 are excluded. The detailed evaluation criteria and scoring prompt are presented in Table \ref{tab:prompt-question-quality-control}.
\item Answerability Verification: We evaluate the answerability of each question using LLM-based assessment. For questions deemed answerable, we generate both the answer and corresponding reference content from the source document. Unanswerable SQs are excluded. The verification prompt is documented in Table \ref{tab:prompt-question-check-answerability}.
\end{itemize}

We note that the above description of Persona-SQ is more suitable for generating a collection of SQs from a collection of documents. This process 1) ensures that we obtain enough generated SQs and normalized personas for corpus-level analyses (see Section \ref{sec:demo-1}) and 2) simplifies the process of synthesizing training data for fine-tuning models for SQ generation (see Section \ref{sec:demo-2}). However, 
we emphasize that it is straightforward to apply Persona-SQ for a single document, which provides personalized SQs in a single reading session; the only difference is that step 1 is no longer needed because the document would be provided by the user from which personas, goals, and personalized SQs will be generated. We also note that the system is extensible; it can be further enriched and expanded to include not only persona information but other types of information to guide and improve the resulting SQs for use cases in addition to personalization.

\section{Demonstration 1: Persona-SQ improves LLM-based SQ generation}
\label{sec:demo-1}
In this section, we demonstrate the efficacy of Persona-SQ when it is instantiated with the state-of-the-art LLM. We show that Persona-SQ improves SQ generation compared to the regular SQ generation pipeline without persona information. We perform automatic evaluation on a large set of documents and generate SQs in Section \ref{sec:automatic-evaluation} and human evaluation on a small set of documents in Section \ref{sec:human-evaluation}. Demo is built with Gradio (Appendix~\ref{app:demo}).

\begin{table}[t!]
    \centering
    \caption{The corpus-level cosine similarity scores $\downarrow$.}
    \vspace{-5pt}
    \setlength\tabcolsep{4pt}
    \resizebox{0.7\columnwidth}{!}
    {
    \begin{tabular}{c|ccc}
        \toprule
         \textbf{Method} & \textbf{Legal}  & \textbf{Finance} & \textbf{Academia}\\
        \midrule
        Baseline & 95.8 & 96.5 & 91.7 \\
        Persona-SQ & \textbf{84.4} & \textbf{89.9} & \textbf{83.2}\\
        \bottomrule
    \end{tabular}
    }
    \label{tab:semantic-similarity}
\end{table}

\subsection{Automatic evaluation}
\label{sec:automatic-evaluation}
The automatic evaluation seeks to demonstrate that personalized SQs are distinctly differentiated across varying objectives, and appropriately tailored to each persona. We introduce five novel evaluation criteria, including question semantic diversity, question persona alignment, and question quality, which are introduced in this section; along with persona distribution and persona alignment distribution skewness, which will be introduced in Appendix \ref{app:metric-2-persona-distribution} and \ref{app:metric-4-coverage-ratio-distribution-skewness} separately. 

\paragraph{Dataset}
We first randomly collect a total of 250 documents from three domains, including finance, legal, and academia. We apply GPT4o-based Persona-SQ to generate persona-specific SQs, and generic SQs without giving persona information which are used as baselines. Table \ref{tab:stats-dataset} displays the statistics on the source documents and the generated personas and SQs. 

\paragraph{Question Semantic Diversity} We assess whether our persona-based approach generates truly distinct questions for different personas by measuring the cosine similarity between questions generated for the same document. Using the \texttt{gte-Qwen2-1.5B-instruct} embedding model \cite{li2023towards}, we compute pairwise similarities between questions generated for different personas and average them to obtain document-level and dataset-level diversity scores. Appendix \ref{app:metric-1-question-semantic-diversity} provides further details on the implementation of this metric. Table~\ref{tab:semantic-similarity} shows that, on the corpus level, Persona-SQ generates more diverse questions compared to the baseline without personas across domains, demonstrating that incorporating personas leads to more differentiated questions targeting different user interests. 
Additional visualizations in Appendix~\ref{app:more-eval-results-question-semantic-diversity} illustrate this increased diversity through similarity heatmaps on a document level.

\begin{table}[t!]
    \centering
    \caption{The corpus-level coverage ratio scores $\uparrow$.}
    \vspace{-5pt}
    \setlength\tabcolsep{4pt}
    \resizebox{0.8\columnwidth}{!}{
    \begin{tabular}{cc|ccc}
        \toprule
       \textbf{Top K} & \textbf{Persona} & \textbf{Legal}  & \textbf{Finance} & \textbf{Academia}\\
        \midrule
       \multirow{2}{*}{Top 1} &  Baseline & 9.6 & 9.2 & 20.8\\
         & Persona-SQ & \textbf{35.1} & \textbf{32.2} & \textbf{31.9} \\
        \midrule
       \multirow{2}{*}{Top 2} & Baseline & 21.6 & 20.6 & 34.3\\
        & Persona-SQ & \textbf{55.9} & \textbf{50.4} & \textbf{50.7} \\
        \midrule
       \multirow{2}{*}{Top 3} &  Baseline & 30.6 & 27.9 & 43.7\\
        & Persona-SQ & \textbf{67.7} & \textbf{61.4} & \textbf{61.2} \\
        \bottomrule
    \end{tabular}
    }
    \label{tab:coverage_ratio}
\end{table}

\paragraph{Question Persona Alignment} We assess whether questions generated by Persona-SQ appropriately reflect their intended personas through a novel ``reverse ranking method'' where an LLM ranks personas based on their relevance to each generated question. The details of the ``reverse ranking method'' is shown in Appendix \ref{app:metric-2-persona-distribution}.
Using this ranking method, we compute a coverage ratio that measures how well questions align with their intended personas. For each question generated using persona $p^i$, we calculate the proportion of times $p^i$ appears as the most relevant persona in the LLM's ranking. Higher ratios indicate better alignment between generated questions and their target personas. More details are available in Appendix \ref{app:metric-3-coverage-ratio}. We report the average coverage of all personas in Table \ref{tab:coverage_ratio}. It is illustrated that questions generated by Persona-SQ achieve significantly higher coverage ratios compared to the baseline without personas, demonstrating that our approach generates questions that better reflect their intended personas. The coverage ratios of three persons in the legal domain are shown in Figure \ref{fig:cover-ratio} as examples.

\paragraph{Question quality} We use GPT4o as a judge~\cite{zheng2023judging} to evaluate a small sample of questions from both Persona-SQ and baselines. The metrics include relevance, readability, importance, and answerability, following the suggestions of recent work~\cite{oh-etal-2023-evaluation,fu-etal-2024-qgeval} and our own observations that traditional question generation metrics such as ROUGE~\cite{lin-2004-rouge} are inappropriate to capture the nuances in a question and it is better to resort to human evaluation (see next paragraph) and LLM-based evaluations. Using GPT4o as evaluator (sample prompts are in Appendix~\ref{app:eval-prompt}), we show in Table~\ref{tab:auto-eval-results} that Persona-SQ significantly improves question importance while mostly maintaining the performance of other metrics. 


\subsection{Human evaluation}
\label{sec:human-evaluation}
We also conduct a preliminary user study in a more realistic scenario where a user uploads a document and observes a set of SQs. We conduct an A/B style test where the user sees a total of six questions, three generated by Persona-SQ and the rest by the baseline, along with a document. The user's task is to rank all six questions in terms of their preferences in decreasing order, i.e., SQ ranked 1 is the most preferred SQ, without knowing which question is generated by which process. We use a subset of 14 documents and recruit 400 users for this study. We then compute the mean and median rankings of questions from both Persona-SQ and baseline, respectively. Both use GPT4o as the LLM to generate SQs. Results in Table~\ref{tab:ranking-results} reveal strong early signal that users prefer SQ generated by our Persona-SQ system compared to baseline. These results further validate the usefulness of Persona-SQ in improving SQs in real-world scenarios.

\section{Demonstration 2: Persona-SQ results in powerful small model for SQ generation}
\label{sec:demo-2}
We additionally demonstrate Persona-SQ's utility in generating synthetic training data to fine-tune extremely small models (less than 400 million parameters) for the task of SQ generation. The reason for choosing models of such extreme small scale is twofold. First, the smaller the model size, the easier it is to implement and run the model within an AI-powered reading application in an actual production environment (more in Appendix~\ref{app:on-device}). Second, there is a growing interest in finding practical use cases for extremely small models.\footnote{For example, see {\small\url{https://shorturl.at/Vs7xn}} and {\small\url{https://shorturl.at/HEHFu}} for relevant discussions.} Both of these motivate us to focus on scaling down model sizes and to contribute a new practical use case, i.e., SQ generation, for these small models. We build this model demo with Gradio (Appendix~\ref{app:demo}). 

\begin{table}[t!]
\centering
\caption{Users rank SQs generated by Persona-SQ system more favorably than SQs generated baseline system.}
\vspace{-5pt}
\resizebox{1\columnwidth}{!}{%
\begin{tabular}{lccc}
\toprule
{\bf Method} & \textbf{Avg. Rank $\downarrow$} & \textbf{Win Ratio $\uparrow$} & \textbf{MRR $\uparrow$} \\
\midrule
Baseline & 4.12 & 24.2\% & 0.313 \\
Persona-SQ & {\bf 2.88} & {\bf 75.8\%} & {\bf 0.504} \\
\bottomrule
\end{tabular}
}
\label{tab:ranking-results}
\end{table}

\paragraph{Dataset.} We instantiate Persona-SQ with an open-source LLM, namely \texttt{Llama-3.1-70B}, and apply it on a large set of diverse documents to generate between 9 and 16 SQs per document. We split the dataset according to document IDs into training, validation, and test sets. Table~\ref{tab:data-stat} shows the resulting dataset's statistics. 

\begin{table}[]
\centering
\caption{Persona-SQ, both using GPT4o and fine-tuned SmolLM, generates higher-quality SQs across most metrics (relevance, readability, importance, and answerability) than SQs generated by baselines without persona information.}
\label{tab:auto-eval-results}
\vspace{-5pt}
\resizebox{\columnwidth}{!}{%
\begin{tabular}{@{}lcccc@{}}
\toprule
{\bf Model/Method}             & {\bf Rel.} & {\bf Read.} & {\bf Imp.} & {\bf Ans.} \\ \midrule
{Baseline (GPT4o)}         & 4.94      & 5.00        & 3.97       & {\bf 4.86}                 \\
{Persona-SQ (GPT4o)}       & {\bf 4.94}      & {\bf 5.00}        & {\bf 4.97}       & 4.75                 \\\midrule
{Baseline (SmolLM 360M)}   & 4.25          & 4.69            &     4.14       &  3.86                  \\
{Persona-SQ (SmolLM 360M)} & {\bf 4.63}      & {\bf 4.77}        & {\bf 4.77}       & {\bf 4.17}                 \\ \bottomrule
\end{tabular}%
}
\end{table}

\paragraph{Models and baselines.}
We fine-tune the SmolLM 360M Instruct model\footnote{We have also attempted an even smaller one, SmolLM 135M Instruct, but the results were not competitive; we leave improving the SQ generation performance for even smaller models as a valuable future direction.} on the SQ dataset synthesized by Persona-SQ as well as by the baseline (without using persona). We also fine-tune them with SQuAD, an open-source QA dataset that we re-purpose for the SQ generation task. More details are in Appendix~\ref{app:model-details}.
 

\paragraph{Evaluations} We conduct a series of evaluations similar to the previous section. {\bf For automatic evaluation}, we first compute question semantic diversity and question persona alignment, comparing the model fine-tuned on the Persona-SQ generated dataset versus the model fine-tuned on an SQ dataset without persona. Table~\ref{tab:auto-eval-score-smalllm-360} succinctly summarizes the results, suggesting that Persona-SQ results in a model capable of generating more diverse questions. We then compare the SQs generated by our Persona-SQ fine-tuned model with those generated by GPT4o and with those generated by non-Persona-SQ fine-tuned model. Results in Table~\ref{tab:auto-eval-results} further confirms that model fine-tuned on Persona-SQ dataset outperforms the baseline model across the board, and approahes the performance of Persona-SQ instantiated with GPT4o. {\bf For human evaluation}, we largely follow the procedure outlined in the previous section, comparing Persona-SQ fine-tuned model with GPT4o baseline without persona. Results in Table~\ref{tab:auto-eval-results} show promising signal that users prefer the Persona-SQ fine-tuned small model over GPT4o baseline, even though our model is perhaps hundred times smaller than GPT4o. More evaluation results are available in Appendix~\ref{app:additional-smollm-results}.

\begin{table}[t!]
    \centering
    \small
    \setlength\tabcolsep{4pt}
    \caption{The evaluation scores of SmolLM-360M and the baseline. \textbf{Sim.} represents the question semantic diversity and \textbf{coverage ratio topK} represents the question persona alignment.}
    \begin{tabular}{c|cccc}
        \toprule
         & & \multicolumn{3}{c}{\textbf{Coverage Ratio}} \\
        \cmidrule(lr){3-5}
        \textbf{Method} & \textbf{Sim. $\downarrow$} & \textbf{Top 1 $\uparrow$} & \textbf{Top 2 $\uparrow$} & \textbf{Top 3 $\uparrow$} \\
        \midrule
        Baseline & 69.3 & 50.0	& 77.3	& 83.8	\\
        Persona-SQ & \textbf{68.1} & \textbf{55.8}	& \textbf{81.7}	& \textbf{88.1}	\\
        
        \bottomrule
    \end{tabular}
    
    \label{tab:auto-eval-score-smalllm-360}
\end{table}


\begin{table}[t!]
\centering
\caption{Users in general prefer SQs generated by our model fine-tuned on the Persona-SQ dataset to those generated by GPT4o without persona.}
\vspace{-5pt}
\resizebox{1\columnwidth}{!}{%
\begin{tabular}{lccc}
\toprule
{\bf Method} & \textbf{Avg. Rank $\downarrow$} & \textbf{Win Ratio $\uparrow$} & \textbf{MRR $\uparrow$} \\
\midrule
Baseline (GPT4o) & 4.38 & 16.7\% & 0.301 \\
Persona-SQ (SmolLM) & {\bf 2.62} & {\bf 83.3\%} & {\bf 0.515} \\
\bottomrule
\end{tabular}
}
\label{tab:ranking-results-smollm}
\end{table}

\paragraph{Deployment considerations} Given its tiny size, the model takes only about 760 megabytes on-device with fp16 weights. With further optimization such as quantization aware training, we can potentially further reduce this model to around 200 megabytes with 4bit weight quantization. Latency when running an un-optimized, un-quantized model on a commercial CPU laptop (MacBook M2) is around 0.5 seconds for model loading and around 10 seconds for generating a persona and question. Further optimization techniques could potentially yield substantial improvements in both storage efficiency and computational performance. The exploration of such optimization strategies presents a promising direction for future research.

\section{Related Work}

\subsection{Question Generation}
Prior work on question generation focuses primarily on the educational use case \cite{wang2018qg, xu2022fantastic, luo2024chain, li2024planning, kumar2024improving}. Those works will result in a question generation pipeline or a model optimized for educational use cases, specifically, generating questions that require students to answer to improve their learning outcomes. In contrast, our work aims to improve the question quality suggested by the AI assistant / chatbot, which helps users to better interact with the assistant and understand documents more easily. Recent works have demonstrated the capability of LLMs to generate high-quality questions \cite{yuan2022selecting, li2024planning, wang2022towards}, which is already implemented in the current AI Assistant. However, those works lack investigations with the personalized question generation. Our Persona-SQ framework bridges this gap by leveraging personas to generate more personalized questions.

\subsection{Personalized Large Language Models}
Personalized LLMs can be divided into two types:
(1) LLM personalization, in which LLMs need to take care of users’ personas (e.g., background information, or historical behaviors) to meet customized needs \cite{salemi-etal-2024-lamp, kumar2024longlamp}; and (2) LLM Role-play, in which LLMs play the assigned personas (i.e., roles) and act in accordance with environmental feedback \cite{shao2023character, shanahan2023role}. Our work belongs to the former type. The definition of {persona} in LLM personalization is different in various works. For example, \citet{sun2024persona} utilizes three personas: distilled persona, induced persona, and historical action to customize LLM's output. Some works define personas as characteristics 
, general facts 
, and historical action 
to customize the dialogue between AI Assistant and users \cite{kim2024commonsense, zhang2018personalizing, tang2023enhancing}. In the personalized healthcare domain, \citet{zhang2024llm} takes the patient profile (e.g., the patient with diabetes) as the persona. Persona-SQ, on the contrary, defines "persona" in two aspects: (1) the profession of the users and (2) the reading goals. We posit that different professions and goals lead to different interests as part of the same document, thus leading to more personalized and diverse questions. 

\section{Conclusion}
We introduced Persona-SQ, an approach to improve suggested questions (SQs) in AI-powered reading applications by incorporating synthetic user profiles consisting of professions and reading goals. Through extensive experiments on documents from diverse domains, we demonstrated that Persona-SQ improves SQ quality and diversity compared to traditional non-personalized approaches. We further showed that Persona-SQ can be used to generate synthetic training data to fine-tune extremely small models (360M parameters) that perform competitively with much larger models on SQ generation. These results suggest two promising directions for improving current AI-powered reading applications: 1) as an immediate drop-in upgrade to existing cloud-based SQ generators to produce more diverse and targeted questions, and 2) as a pipeline to train small, efficient models that can generate high-quality personalized SQs directly on users' devices. We hope our work spurs further research into making AI-powered reading assistants more personalized and accessible.

\section*{Limitations}
We acknowledge two limitations of our work. First, Persona-SQ uses synthetically generated personas (professions and goals) rather than actual user profiles. While our experiments show that even synthetic personas improve SQ quality and diversity over non-personalized baselines, this approach does not yet achieve true personalization. However, the synthetic personas provide natural anchor points for collecting user preference signals - if a user frequently clicks on questions associated with certain personas, this interaction data could be used to infer the user's actual professional background and interests. Once real user profiles become available through such interaction logging or other methods, they can directly replace the synthetic personas in our pipeline without architectural or system changes.

Second, Persona-SQ introduces additional computation from persona generation and multiple quality control steps, potentially increasing system latency. For cloud deployments where the models are accessed through APIs, emerging specialized hardware can help mitigate this overhead. For on-device deployments, our results with extremely small models suggest that the entire pipeline can run efficiently on local devices - the small models can generate SQs quickly while maintaining competitive quality against much larger models, and the quality control steps can be simplified or removed since the model is specifically trained for generating high-quality questions.

\bibliography{emnlp2023}
\bibliographystyle{acl_natbib}

\appendix

\section{Demo details}
\label{app:demo}
We use Gradio to build the demos.

\begin{figure*}
    \centering
    \includegraphics[width=\linewidth]{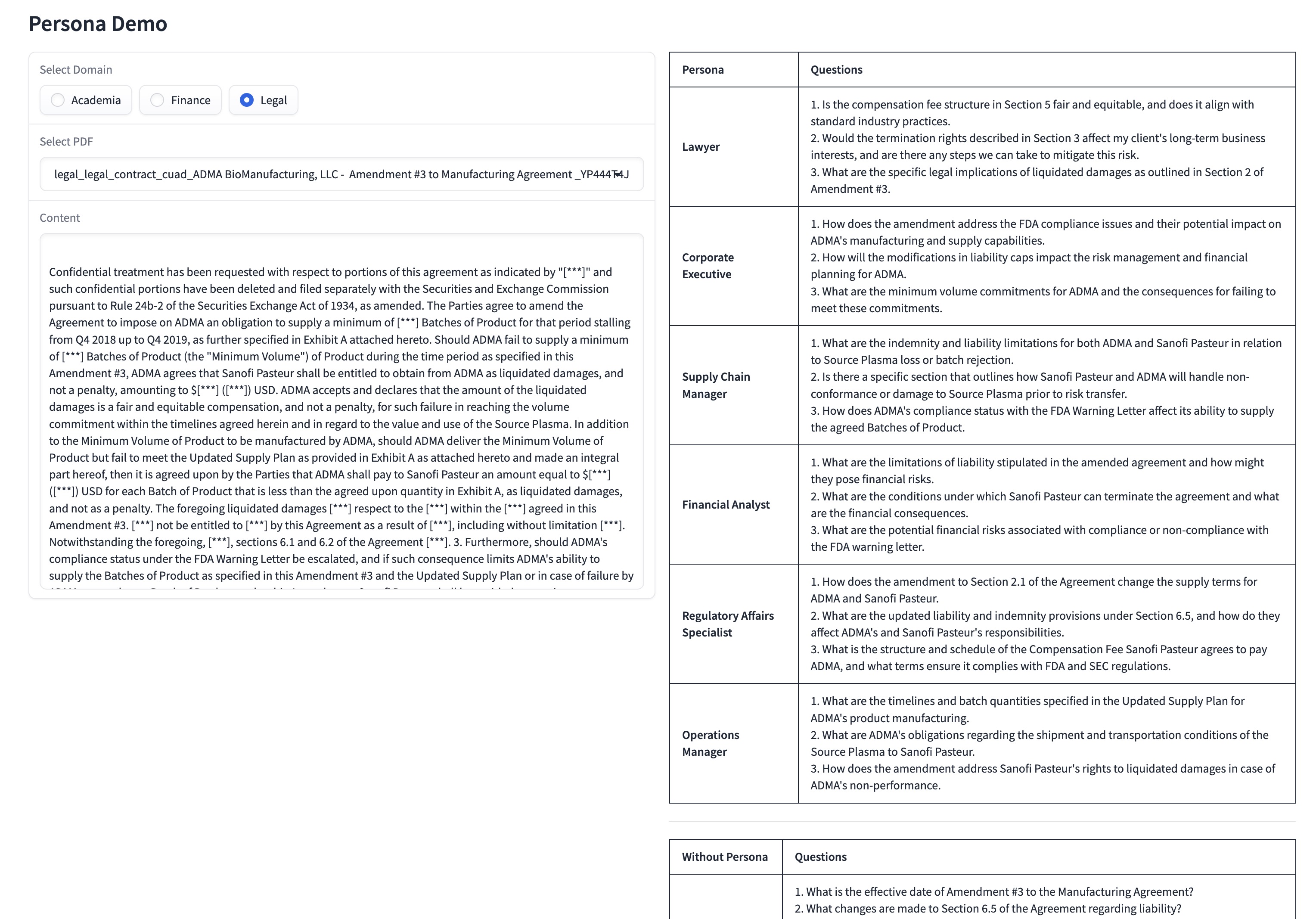}
    \caption{Screenshot01 of the Persona-SQ GPT-4o demo.}
    \label{fig:demo-1-screenshot-1}
\end{figure*}

\begin{figure*}
    \centering
    \includegraphics[width=\linewidth]{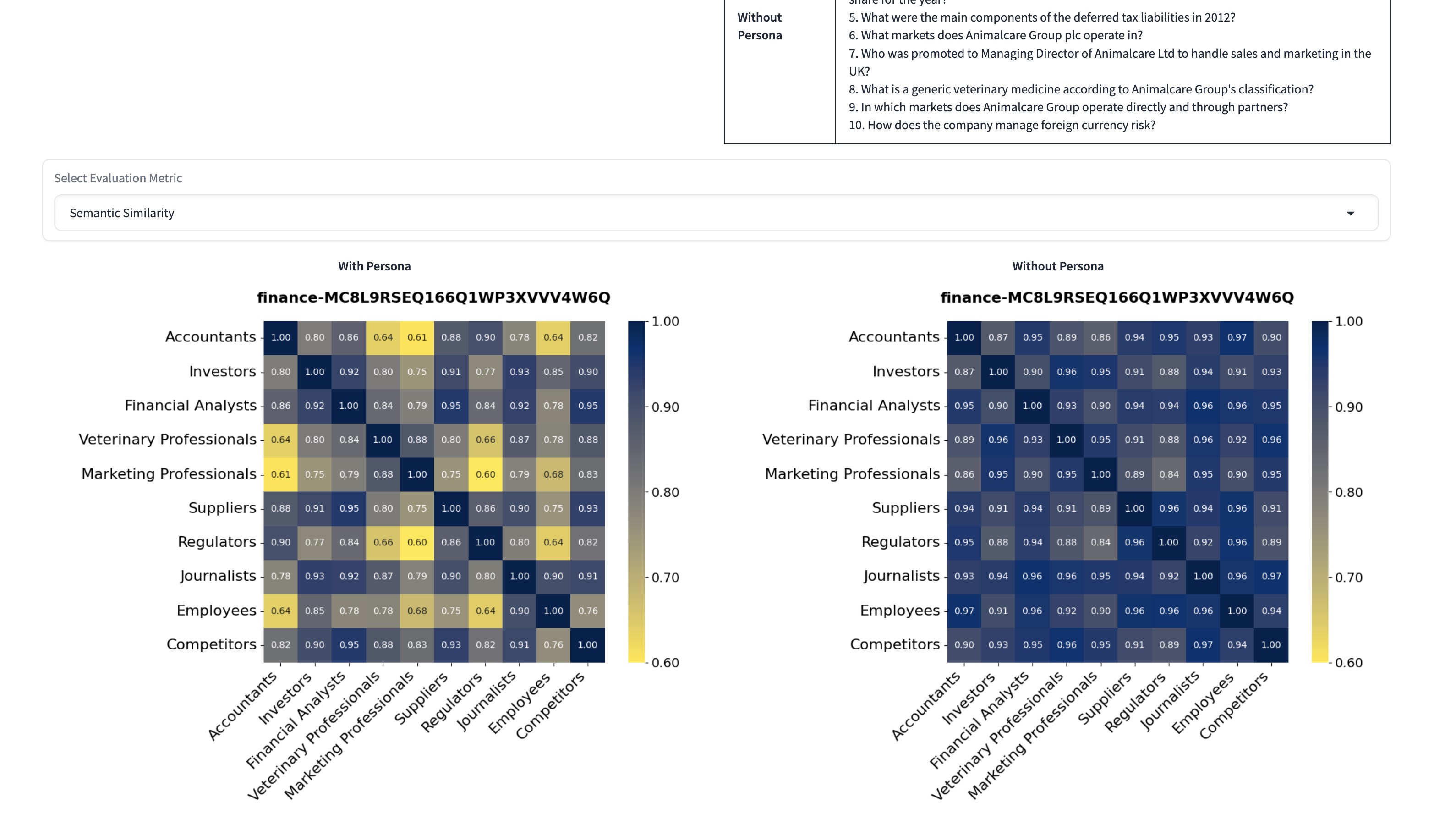}
    \caption{Screenshot-2 of the Persona-SQ GPT-4o demo.}
    \label{fig:demo-1-screenshot-2}
\end{figure*}

\paragraph{Persona-SQ with GPT4o demo} 
Figure ~\ref{fig:demo-1-screenshot-1} presents a visual representation of the demonstration interface. This interactive demonstration showcases the capabilities of Persona-SQ powered by GPT-4o. The interface comprises two primary components: a document selection panel in the upper left section, where users can specify both the domain and target document, and a dual-pane display area. The left pane presents the selected document, while the right pane displays both personalized and generalized self-questions. As illustrated in Figure ~\ref{fig:demo-1-screenshot-2}, the interface incorporates evaluation metric selection functionality. Upon metric selection, the system generates comparative visualizations, juxtaposing the performance analyses of personalized and generalized SQs through distinct graphical representations.

\begin{figure*}
    \centering
    \includegraphics[width=\linewidth]{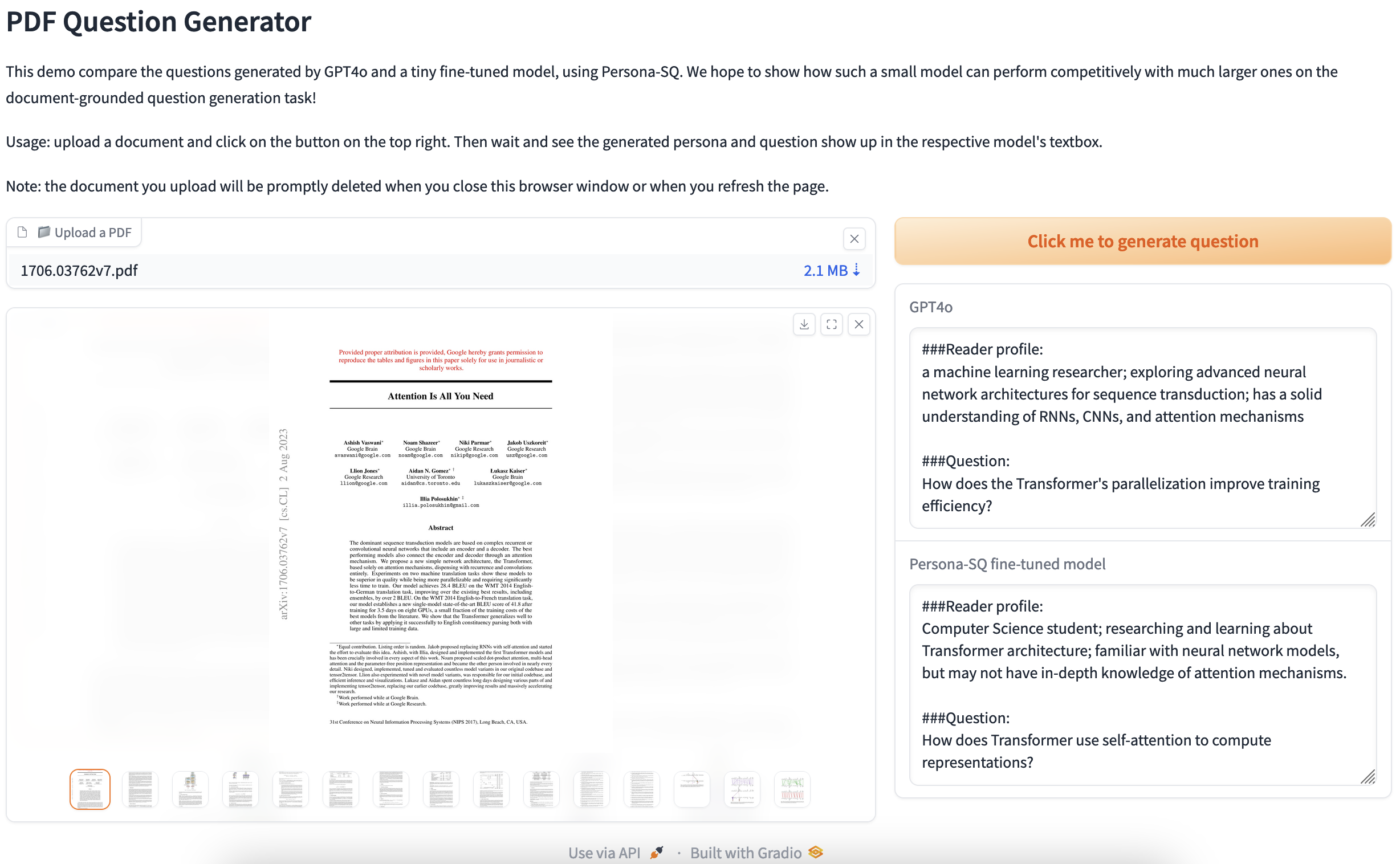}
    \caption{Screenshot of the Persona-SQ fine-tuned demo interface.}
    \label{fig:demo-2-screenshot}
\end{figure*}

\paragraph{Persona-SQ fine-tuned model demo}
A screenshot of the demo is shown in Figure~\ref{fig:demo-2-screenshot}. After uploading the document, we preprocess the document using \texttt{PyMuPDF} to extract the textual content and select the first 1500 tokens if the document is too long as the input to the models. A document preview is shown on the left side of the demo interface. We then feed the extracted document content into the Persona-SQ fine-tuned model and GPT4o. The generated personas and questions are compared side by side on the right side of the demo interface.

\section{Utilize Persona-SQ}
\label{sec:utilize-persona-sq}
In this section, we show how to use our Persona-SQ to generate personalized SQs. We build a python tool that helps to efficiently generate personalized SQs. 
As discussed in Section \ref{sec:persona-sq-framework}, after collecting documents, users can first generate high-quality diverse personas and goals in Step 1-3 using the following code:

\begin{lstlisting}[language=Python, caption=Generate Personas]
# Generate Personas and Goals
generate_persona_and_goal(domain, subdomain, dataset_name, save_base_folder, document)

# Normalize Personas and Goals
classify_personas(domain, subdomain, dataset_name, save_base_folder, persona_and_goal)

# Quality Control for Personas and Goals
control_quality_of_persona_and_goal(domain, subdomain, dataset_name, save_base_folder, persona_and_goals)
\end{lstlisting}

Then, users may generate personalized questions based on previously generated personas and goals for each document and evaluate all the SQs:

\begin{lstlisting}[language=Python, caption=Generate Personas]
# Generate Personalized SQs
generate_questions_raw(domain, subdomain, dataset_name, save_base_folder, persona_and_goal, document)

# Evaluate Quality of SQs
control_generated_question_quality(generated_questions, documents, prompt_for_eval_quality)

# Evaluate Answerability of SQs
evaluate_generated_question_answerability(generated_questions, documents, prompt_for_eval_answerability)

# Quality Control for SQs
filter_generated_question(generated_questions, eval_quality_scores, eval_answerability_results)
\end{lstlisting}

\section{Details on metrics}
\label{sec:appendix}

\subsection{Questions Semantic Diversity}
\label{app:metric-1-question-semantic-diversity}
Specifically, given a document $D_{u}$, where $u$ denotes the document index within domain $d$, \textbf{Persona-SQ} generates questions for $m$ distinct personas, represented as ${p^1, ..., p^m}$. For each persona $p^i$, \textbf{Persona-SQ} produces $t_i$ questions, denoted as ${q_1^{i}, ..., q_{t_i}^{i}}$. We employ an embedding model to transform all questions into vector representations and compute the cosine similarity between the questions generated for different personas. For any two personas $p^i$ and $p^j$, the mean question similarity is computed by:
\begin{equation}
\text{SIM}(p^i, p^j) = \frac{\sum \limits_{e=1}^{t_i}\sum \limits_{f=1}^{t_j}(\text{COS}(q_e^{i}, q_f^{j}))}{t_i * t_j}
\end{equation}
Subsequently, we aggregate all pairwise SIM scores between personas to obtain a comprehensive measure of SQ diversity for document $D_u$. The aggregate similarity is calculated by:
\begin{equation}
\text{SIM}_{D_u} = \frac{\sum \limits
_{i=1}^{m}\sum \limits_{j=i+1}^{m}\text{SIM}(p^i, p^j)}{m(m-1)}
\end{equation}
A higher $\text{SIM}_{D_u}$ value indicates greater similarity among SQs generated for different personas. We average them to get the dataset level score, denoted as $\text{SIM}_{\mathcal{D}} = \sum \limits_{u=1}^{U} \text{SIM}_{D_u}$. Our framework is designed to yield a lower $\text{SIM}_{\mathcal{D}}$, reflecting greater differentiation between persona-specific questions.


\subsection{Metric 2: Persona Distribution}
\label{app:metric-2-persona-distribution}
Persona distribution assesses the distribution of the persona that is related to the SQs generated by giving one document.
We introduce a novel "reverse" evaluation method that uses an LLM to rank personas based on their relevance to each generated question. For instance, given the SQ "What's the profit of the company this year" in the finance domain, and personas "investor," "auditor," and "manager," the LLM will return the ordered list ["investor," "manager," "auditor"], indicating decreasing relevance to the question. Table \ref{tab:prompt-question-check-answerability} provides example inputs and outputs for this process. 

\begin{figure}[tbp]
    \centering
    \includegraphics[width=\linewidth]{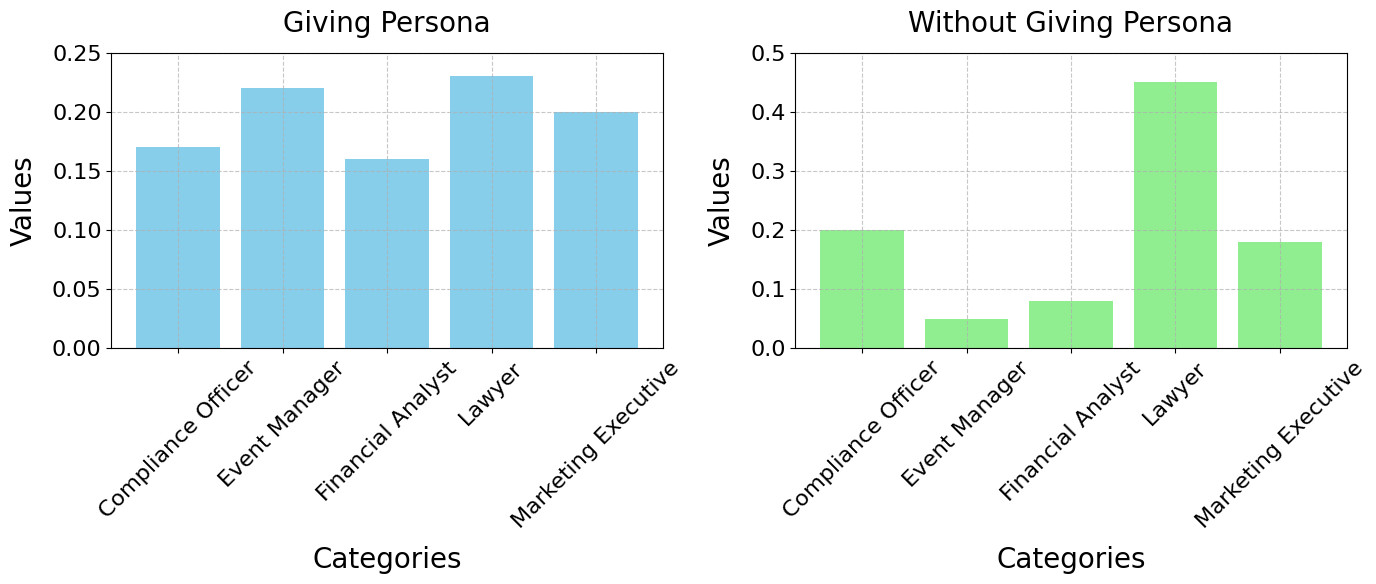}
    \vspace{-5pt}
    \caption{The persona distribution.}
    \label{fig:persona-distribution}
\end{figure}

For each question, we select the first rank persona as the corresponding persona. We calculate the ratio of all corresponding personas of the questions generated for one document. We show the persona distribution of one document by bar plot. As illustrated in Figure \ref{fig:persona-distribution}, the introduction of personas resulted in a more uniform distribution of persona-related questions compared to the baseline generation without persona assignments. Notably, we observed that in the legal domain, suggested questions (SQs) generated without persona guidance tend to converge toward a "lawyer" persona. This phenomenon suggests the existence of domain-specific dominant personas that implicitly influence question generation, which potentially limits the personalization capabilities of AI-powered reading applications. 

\subsection{Metric 3: Question Persona Alignment}
\label{app:metric-3-coverage-ratio}
We utilize Coverage Ratio which assesses how well the generated SQs align with their intended personas at the domain dataset level, to evaluate the question persona alignment. We apply the same "reverse" evaluation aforementioned.

To quantify the coverage ratio, we first define several key metrics. Let $T_i = \sum_{u=1}^{N_d}t_{i, u}$ represent the total number of questions generated for persona $p^i$ across all documents in domain $d$, where: $t_{i, u}$ is the number of questions generated for persona $p^i$ from the $u$-th document, and 
$N_d$ is the total number of documents in domain $d$.

We then define $NUM_{u,(p^i, p^j)}^k$ as the number of questions that satisfy two conditions: (1) They were generated by giving persona $p^i$; (2) In the LLM's ranking, persona $p^j$ appears at position $k$. The coverage ratio is then calculated by:
\begin{equation}
R^k_{(p^i, p^j)} = \frac{NUM_{u,(p^i, p^j)}^k}{t_{i,u}}
\end{equation}
A higher value of $R^k_{d, (p^i, p^j)}$ indicates stronger relevance between the generated questions and their target personas. In our evaluation, we particularly focus on $R^k_{d, (p^i, p^i)}$, which measures how often questions generated for a specific persona are indeed most relevant to that same persona. 
Higher values of this metric indicate better persona-specific question generation.
\subsection{Metric 4: Coverage Ratio Distribution Skewness}
\label{app:metric-4-coverage-ratio-distribution-skewness}
Coverage Ratio Distribution Skewness (CRDS) extends the coverage ratio metric to evaluate how effectively Persona-SQ handles less frequent personas. While the basic coverage ratio measures persona-question alignment, CRDS specifically assesses the system's ability to generate relevant questions across all personas, including those that appear less frequently in the dataset. We construct a distribution using the set of coverage ratios $R^k_{(p^i, p^i)}$ for all personas $i \in {1, ..., m }$ and calculate its statistical skewness—a measure that quantifies the distribution's asymmetry around its mean. An absolute skewness value close to zero indicates a more symmetric distribution, suggesting that \textbf{Persona-SQ} generates questions with similar relevance across all personas, regardless of their frequency in the dataset. This metric is particularly important for ensuring the system maintains high performance even for underrepresented personas. 

\begin{figure*}[!ht]
    \centering
    \subfigure[With Persona]{
        \includegraphics[width=0.48\textwidth]{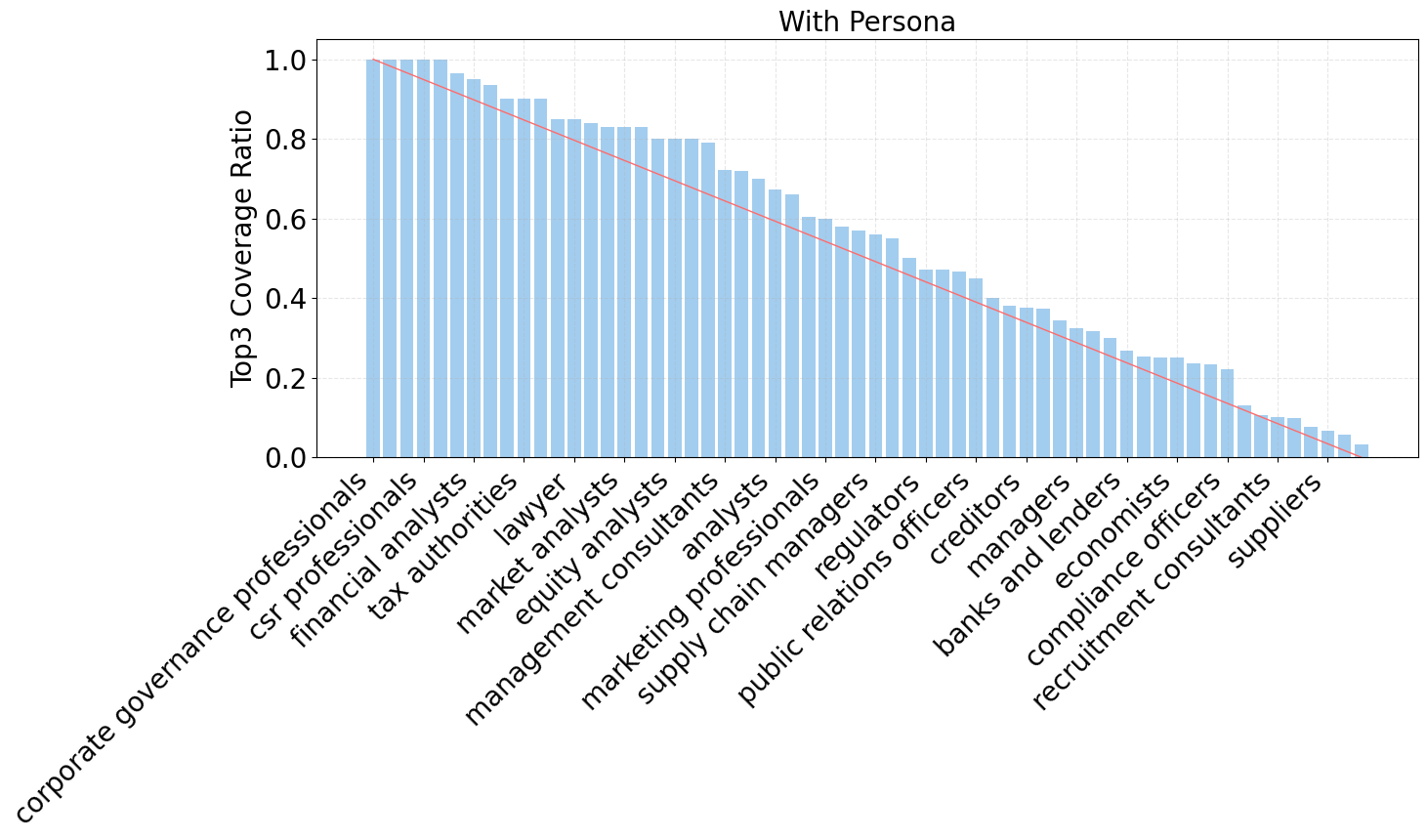}
    }
    \hfill  
    \subfigure[Without Persona]{
        \includegraphics[width=0.48\textwidth]{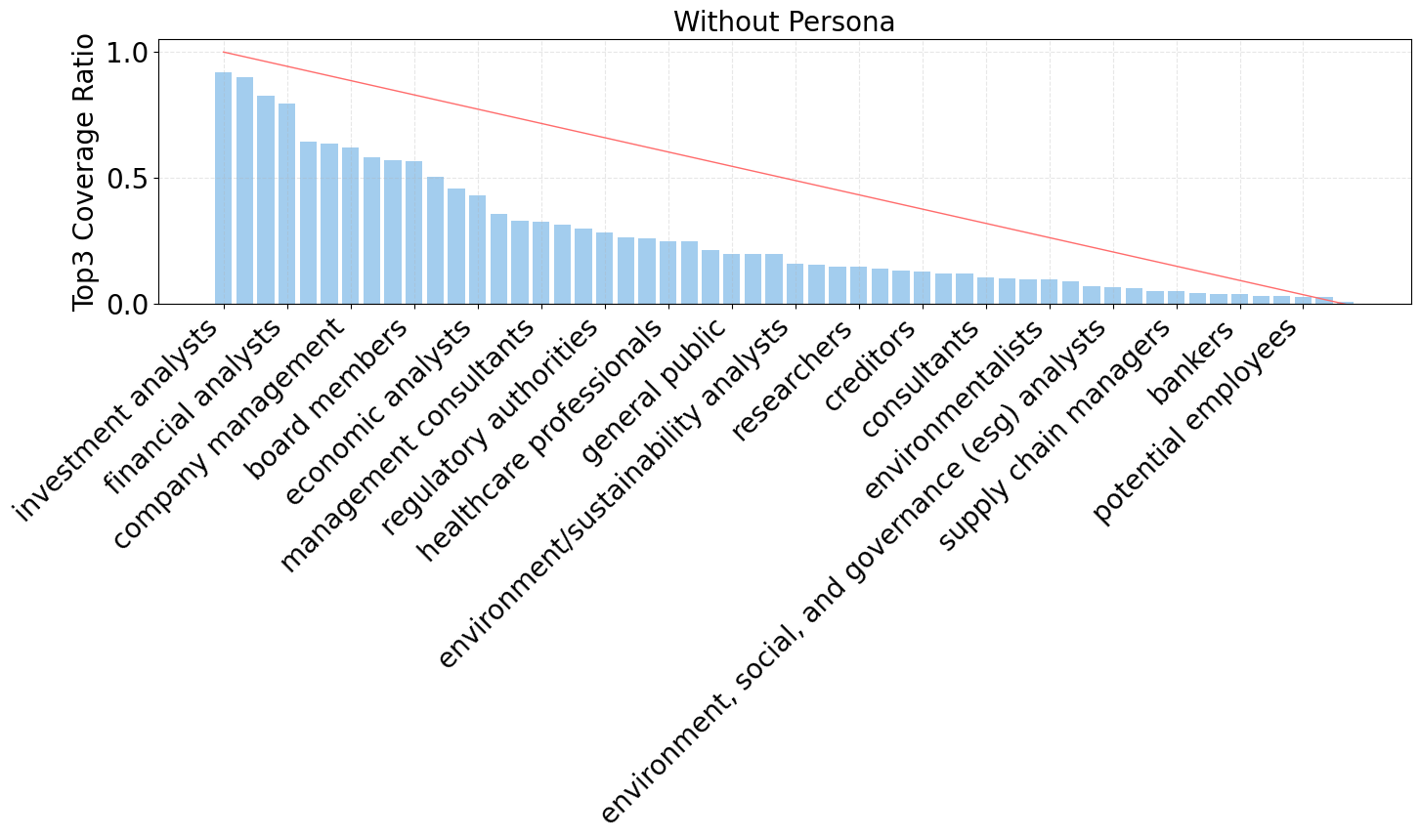}
    }
    \vspace{-5pt}
    \caption{The coverage ratio distribution, showing that Persona-SQ covers more diverse questions than the baseline. For clarity, we print the persona label every three presonas in x-axis.}
    \label{fig:cover-ratio-distribution}
\end{figure*}

The distribution characteristics are visualized in Figure \ref{fig:cover-ratio-distribution}. It reveals that persona-guided generation results in significantly less skewed distributions compared to non-guided generation. This finding indicates that Persona-SQ successfully generates questions that encompass a broader range of personas, including those less frequently represented in the dataset.

\section{More Evaluation Results}
\subsection{Questions Semantic Diversity}
\label{app:more-eval-results-question-semantic-diversity}
\begin{figure*}[!ht]
    \centering
    \subfigure[With Persona (Case 1)]{
        \includegraphics[width=0.48\textwidth]{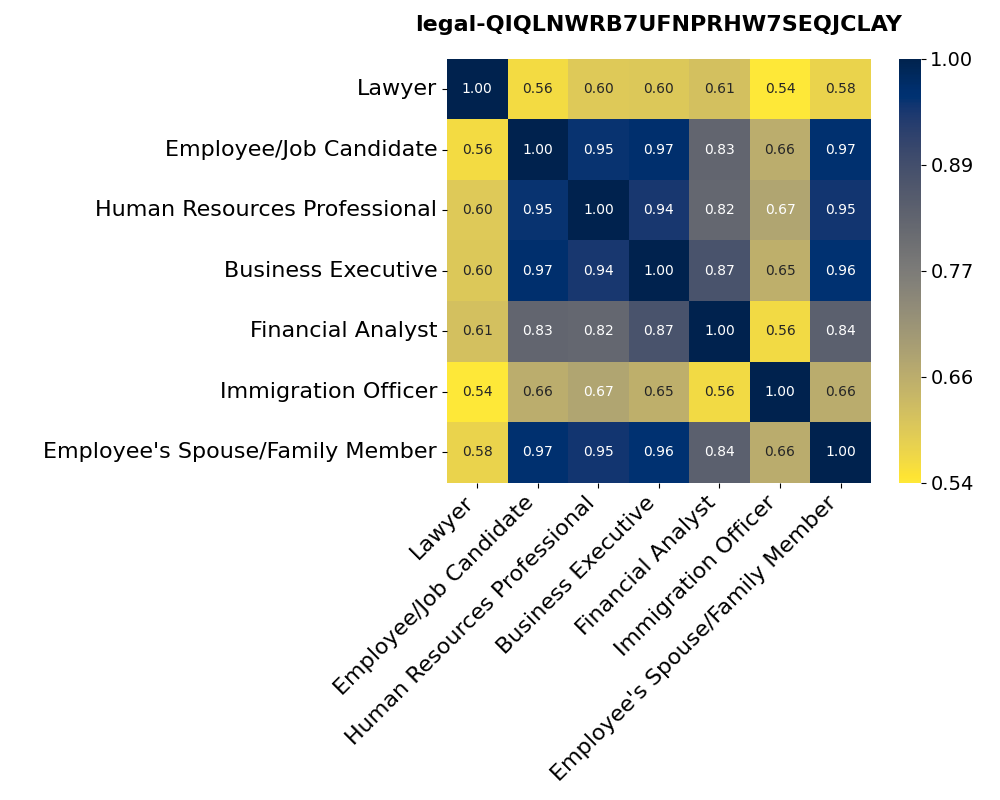}
    }
    \hfill
    \subfigure[Without Persona (Case 1)]{
        \includegraphics[width=0.48\textwidth]{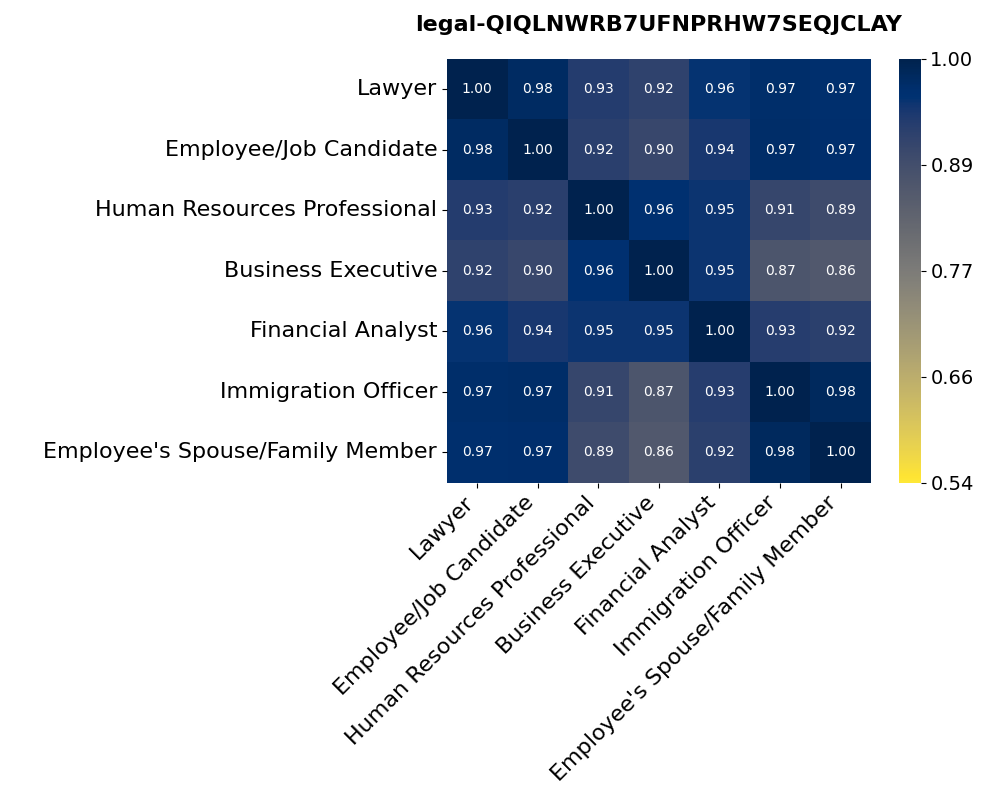}
    }
    
    \subfigure[With Persona (Case 2)]{
        \includegraphics[width=0.48\textwidth]{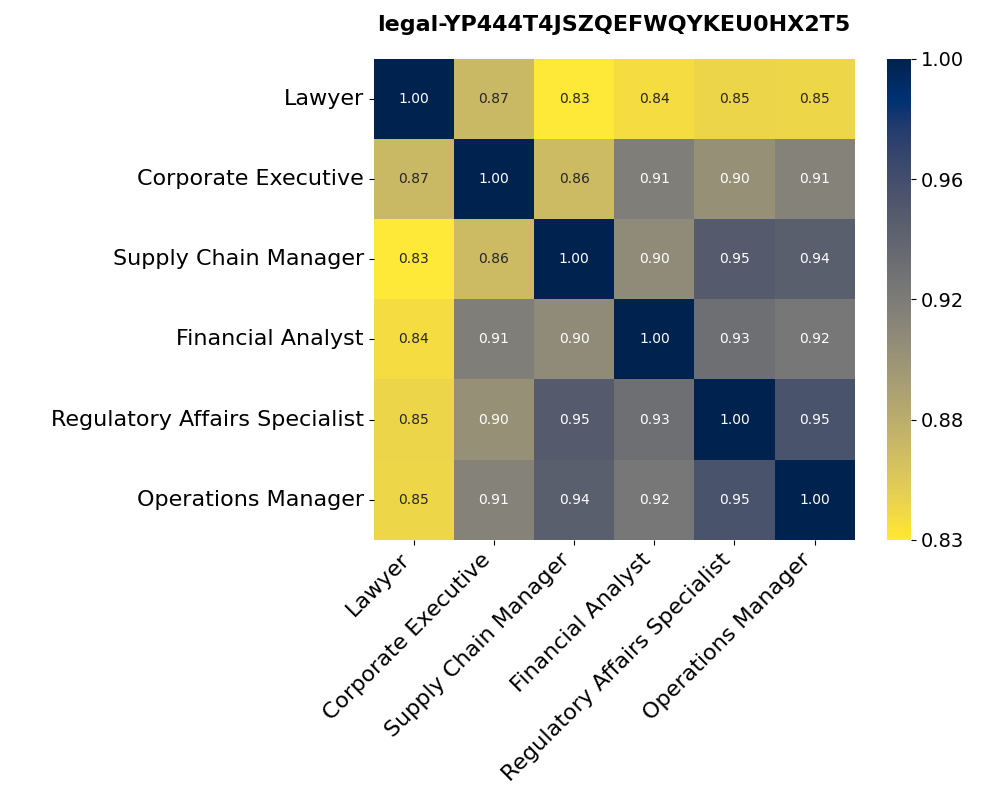}
    }
    \hfill
    \subfigure[Without Persona (Case 2)]{
        \includegraphics[width=0.48\textwidth]{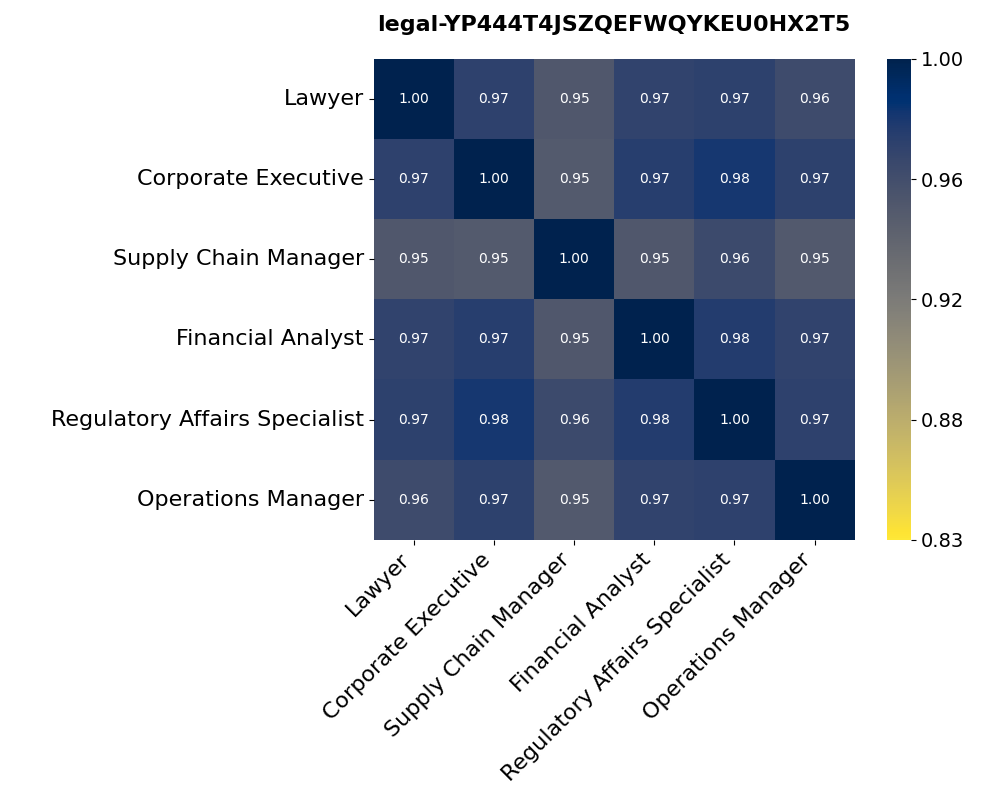}
    }
    
    \subfigure[With Persona (Case 3)]{
        \includegraphics[width=0.48\textwidth]{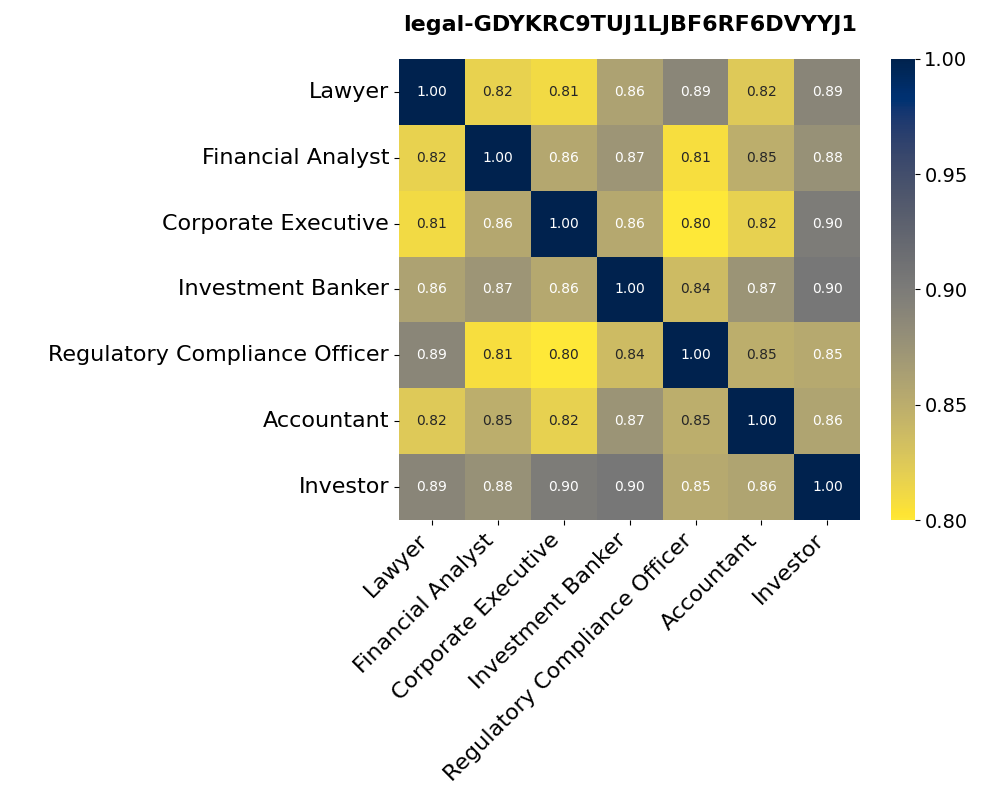}
    }
    \hfill
    \subfigure[Without Persona (Case 3)]{
        \includegraphics[width=0.48\textwidth]{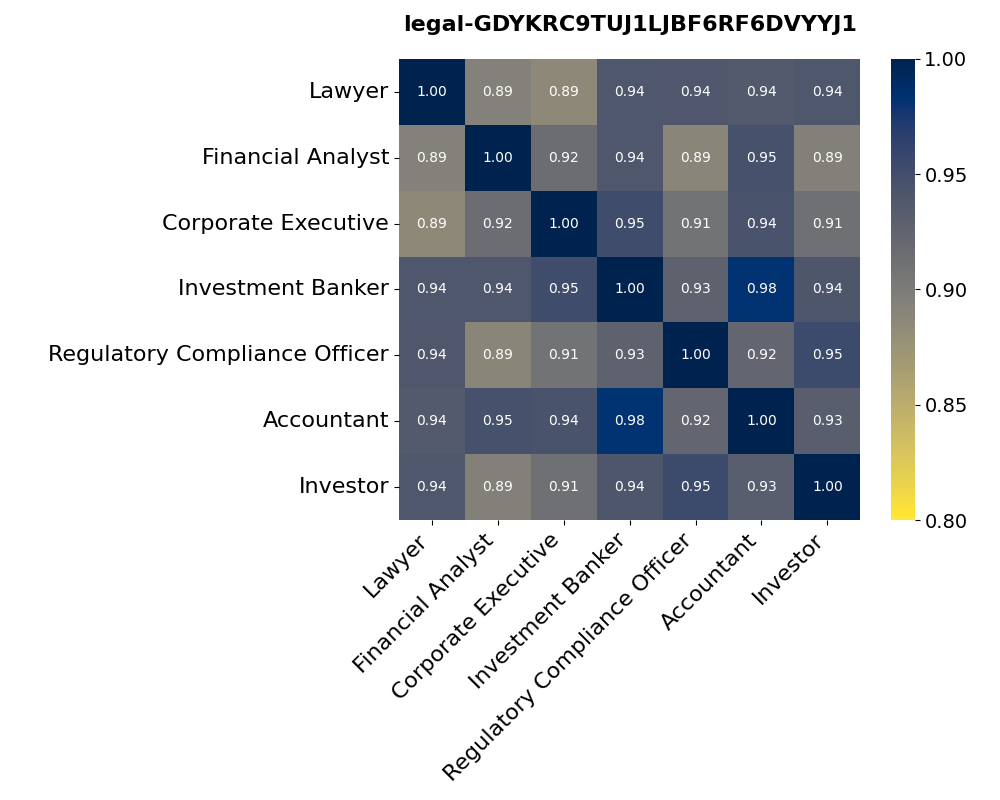}
    }
    
    \vspace{-5pt}
    \caption{Case 1-3: Document-level comparison of semantic similarities between SQs generated with and without persona across three different cases in \textbf{legal} domain.}
    \label{fig:similarity_comparison_app_legal_case_1_3}
\end{figure*}

\begin{figure*}[!ht]
    \centering
    \subfigure[With Persona (Case 4)]{
        \includegraphics[width=0.48\textwidth]{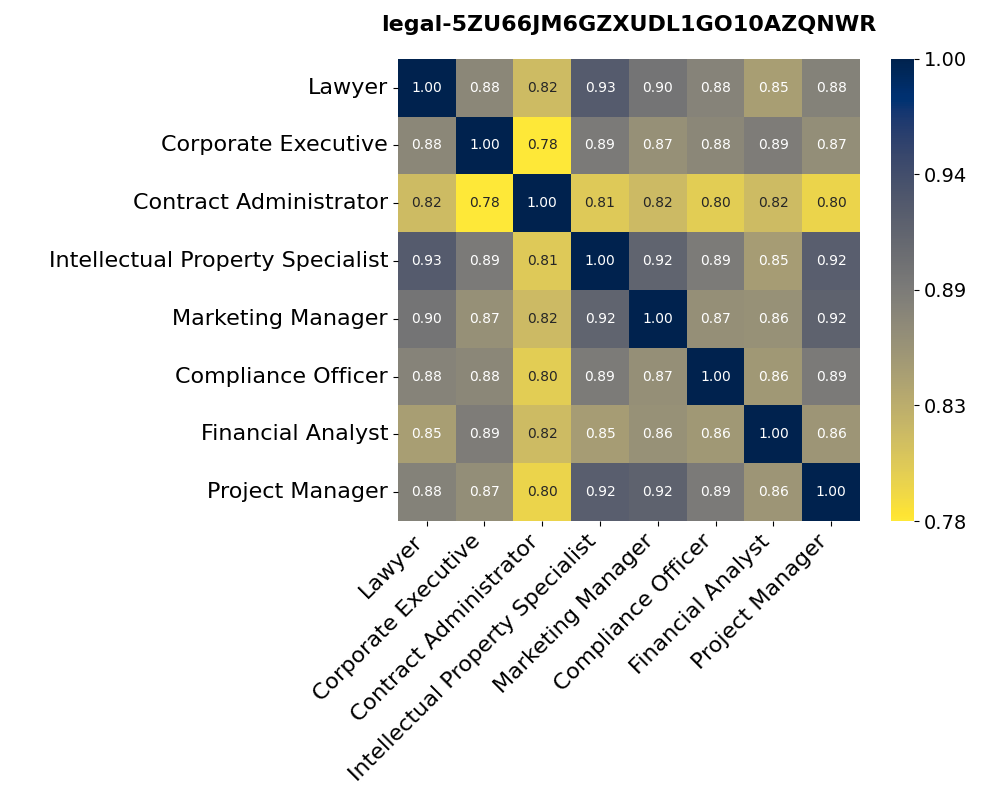}
    }
    \hfill
    \subfigure[Without Persona (Case 4)]{
        \includegraphics[width=0.48\textwidth]{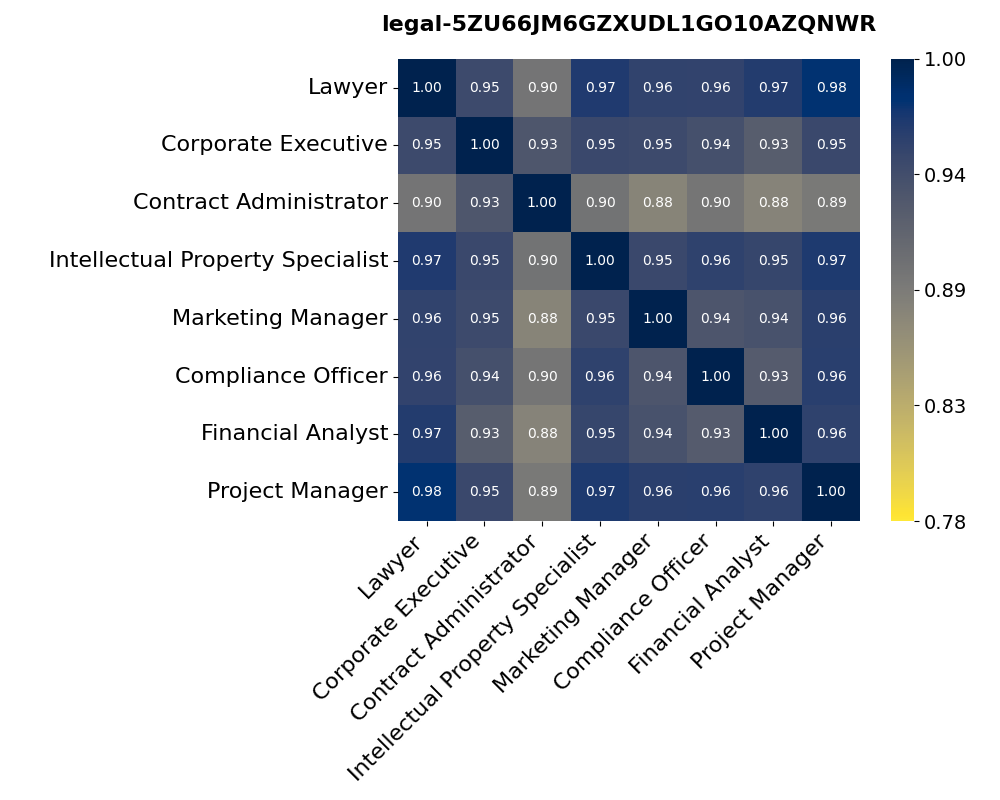}
    }
    
    \subfigure[With Persona (Case 5)]{
        \includegraphics[width=0.48\textwidth]{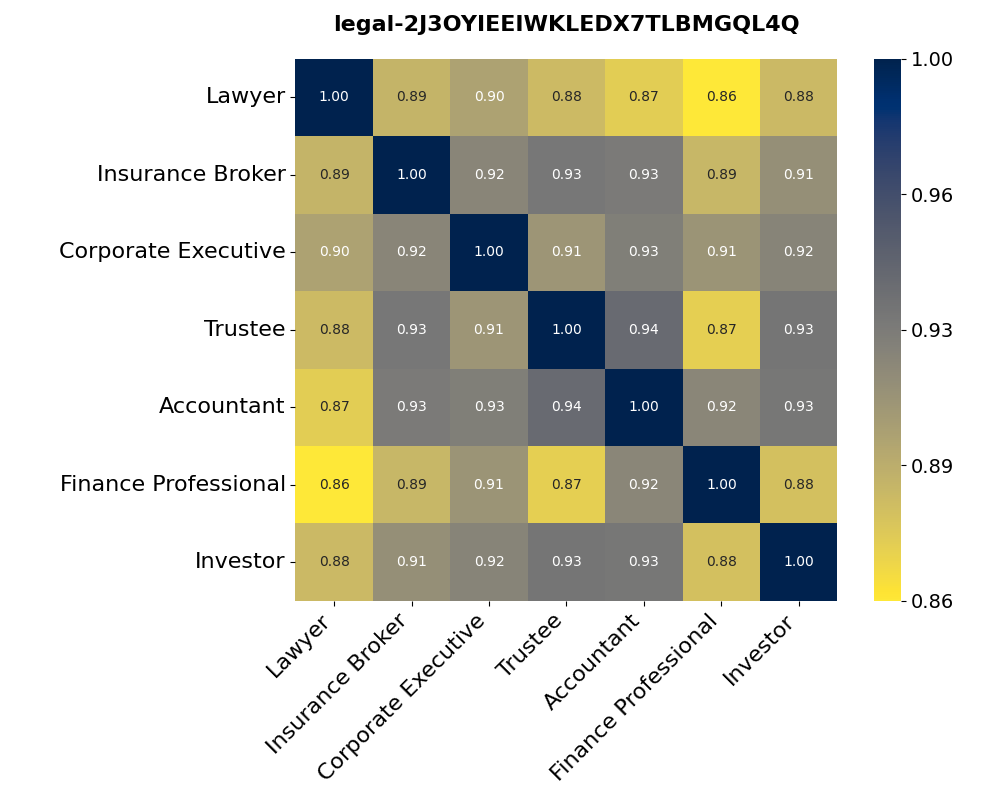}
    }
    \hfill
    \subfigure[Without Persona (Case 5)]{
        \includegraphics[width=0.48\textwidth]{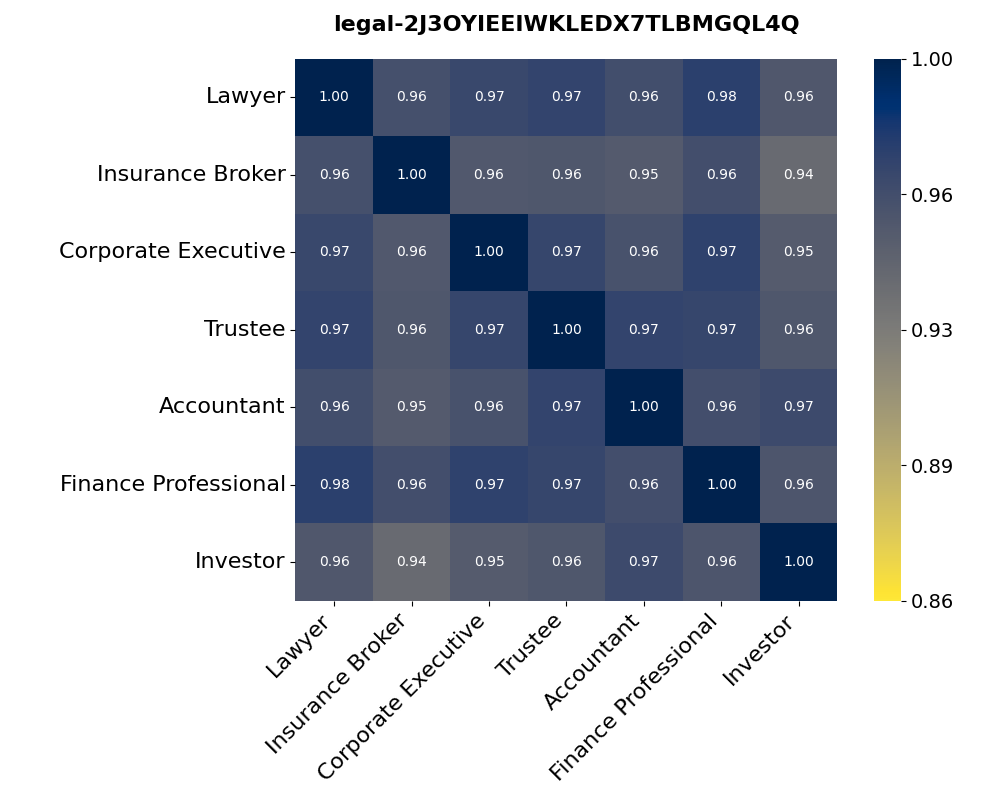}
    }
    
    \subfigure[With Persona (Case 6)]{
        \includegraphics[width=0.48\textwidth]{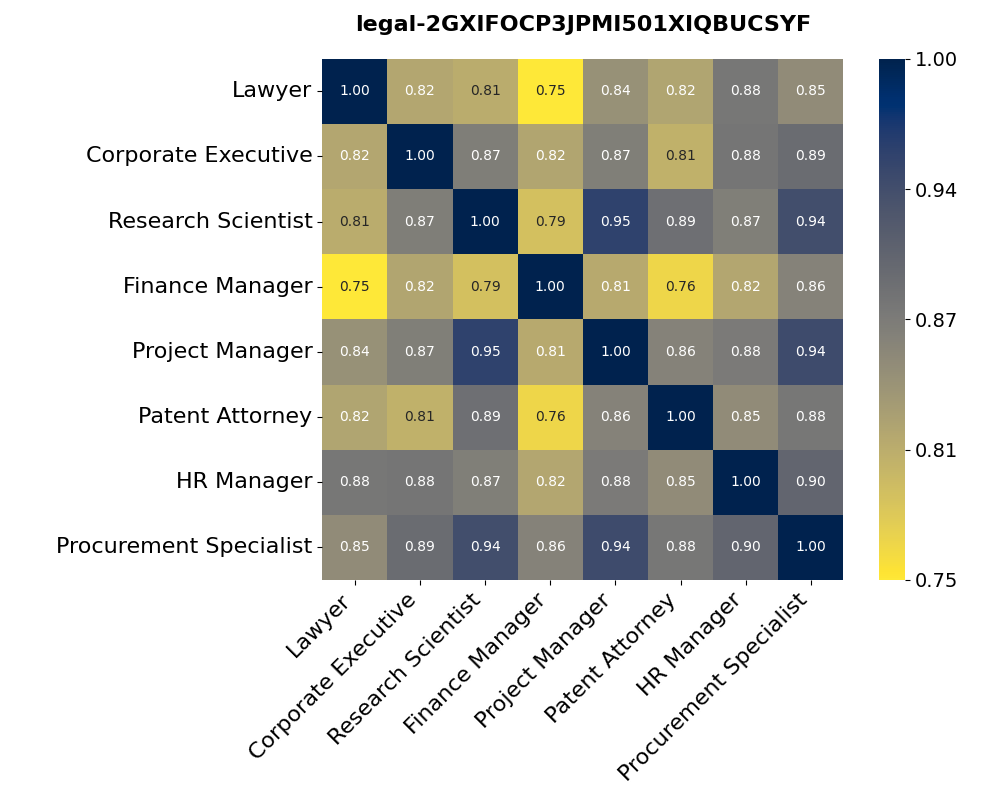}
    }
    \hfill
    \subfigure[Without Persona (Case 6)]{
        \includegraphics[width=0.48\textwidth]{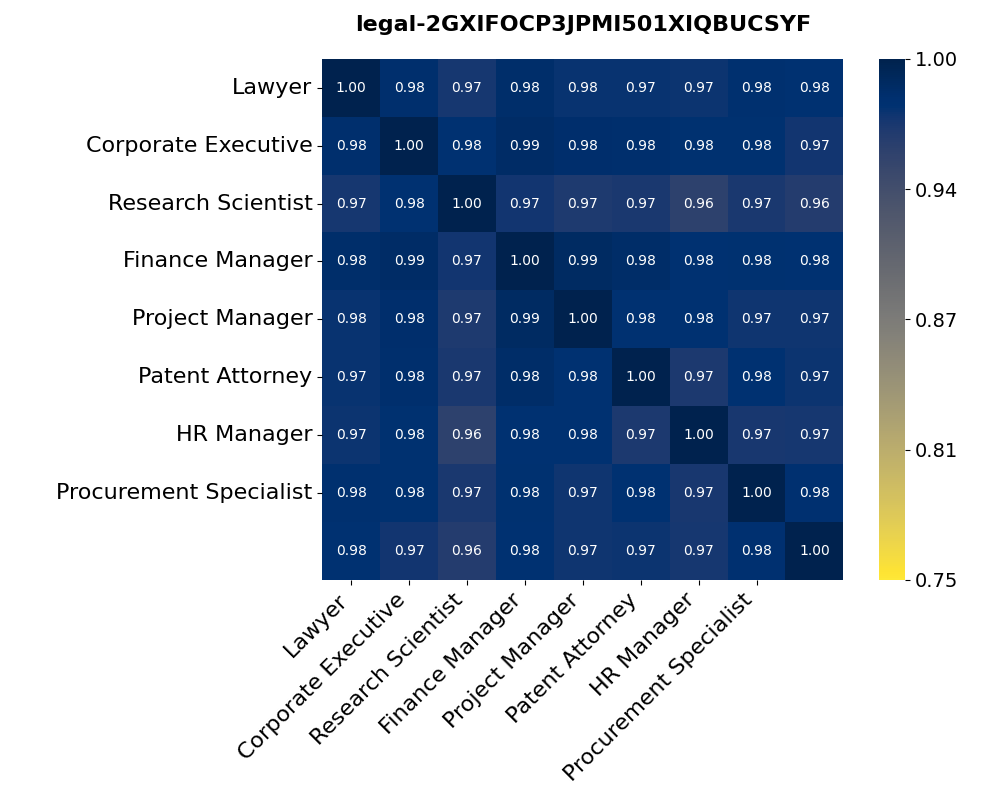}
    }
    
    \vspace{-5pt}
    \caption{Case 4-6: Document-level comparison of semantic similarities between SQs generated with and without persona across three different cases in \textbf{legal} domain.}
    \label{fig:similarity_comparison_app_legal_case_4_6}
\end{figure*}

\begin{figure*}[!ht]
    \centering
    \subfigure[With Persona (Case 1)]{
        \includegraphics[width=0.48\textwidth]{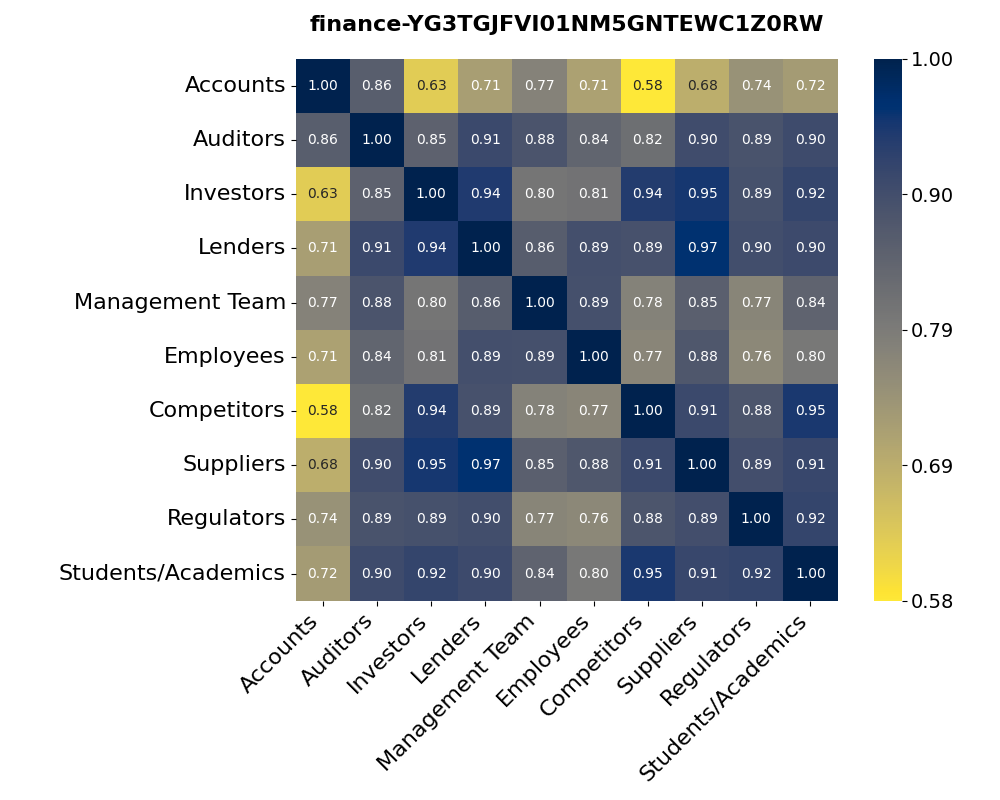}
    }
    \hfill
    \subfigure[Without Persona (Case 1)]{
        \includegraphics[width=0.48\textwidth]{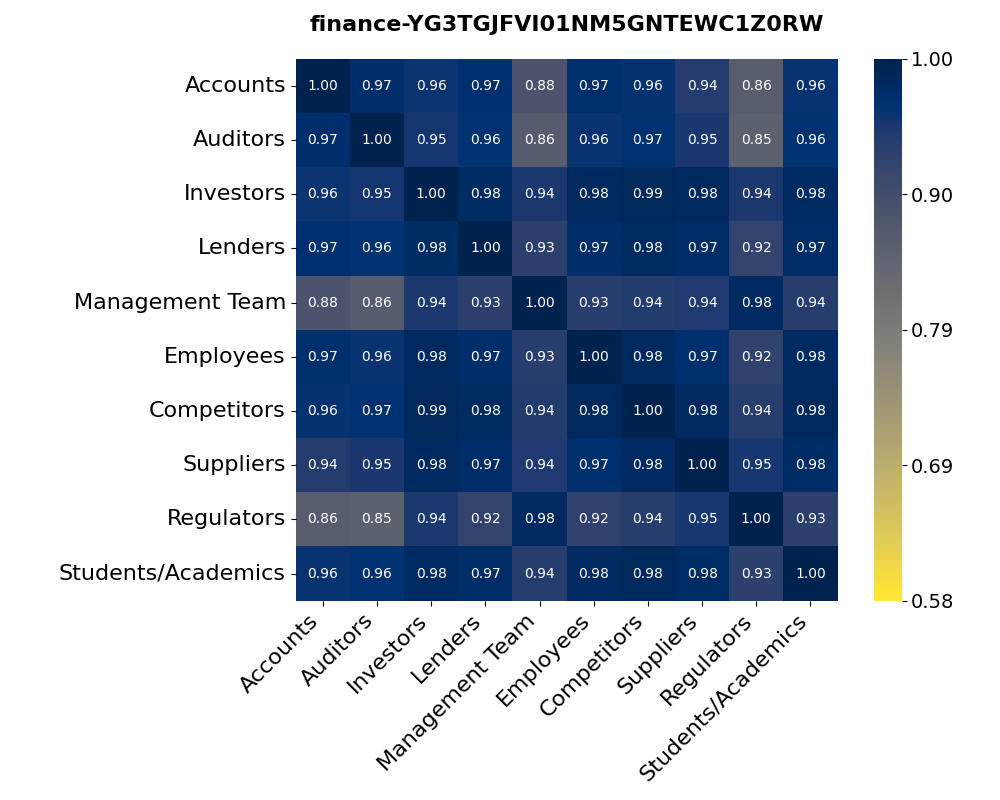}
    }
    
    \subfigure[With Persona (Case 2)]{
        \includegraphics[width=0.48\textwidth]{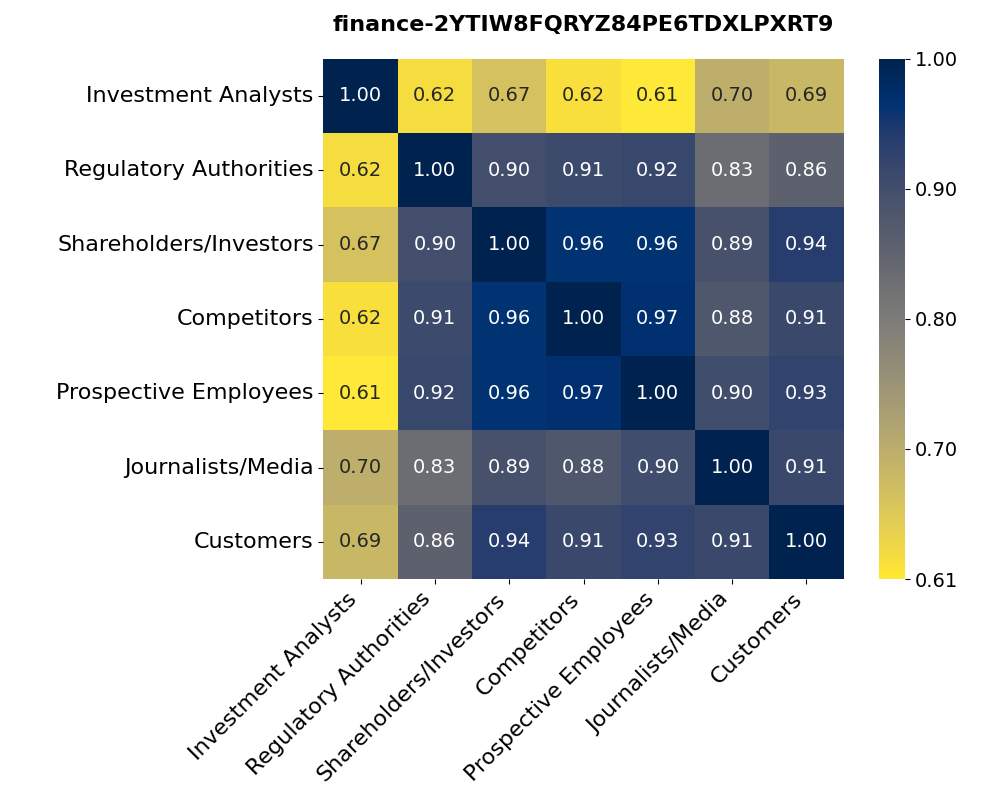}
    }
    \hfill
    \subfigure[Without Persona (Case 2)]{
        \includegraphics[width=0.48\textwidth]{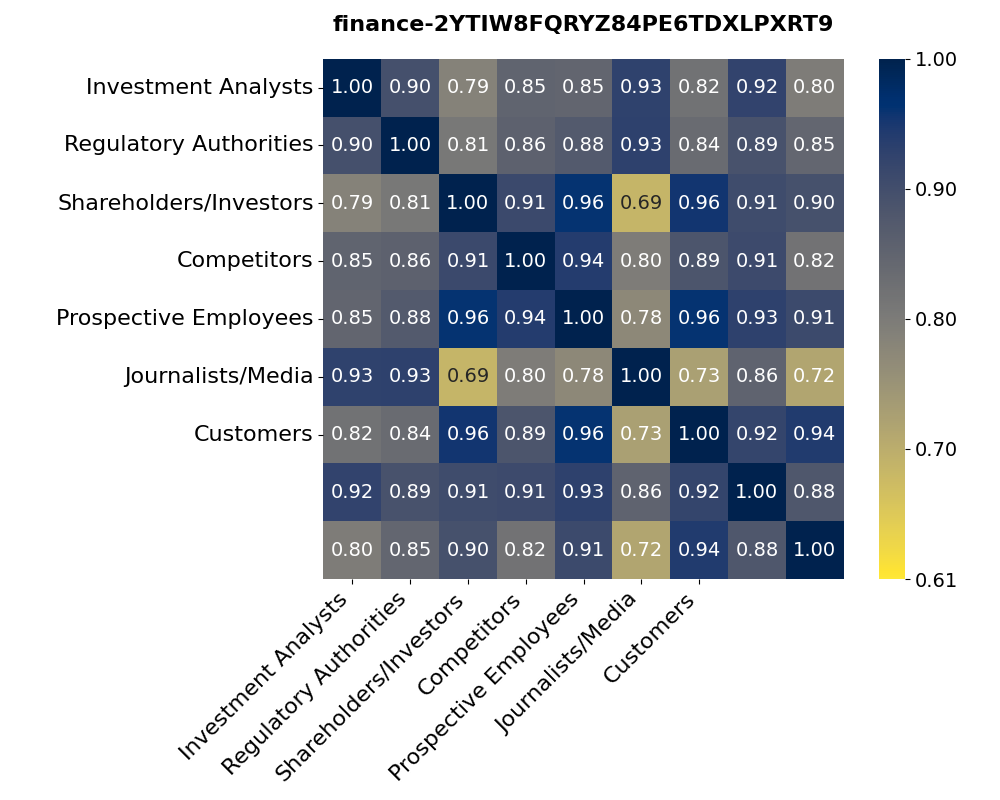}
    }
    
    \subfigure[With Persona (Case 3)]{
        \includegraphics[width=0.48\textwidth]{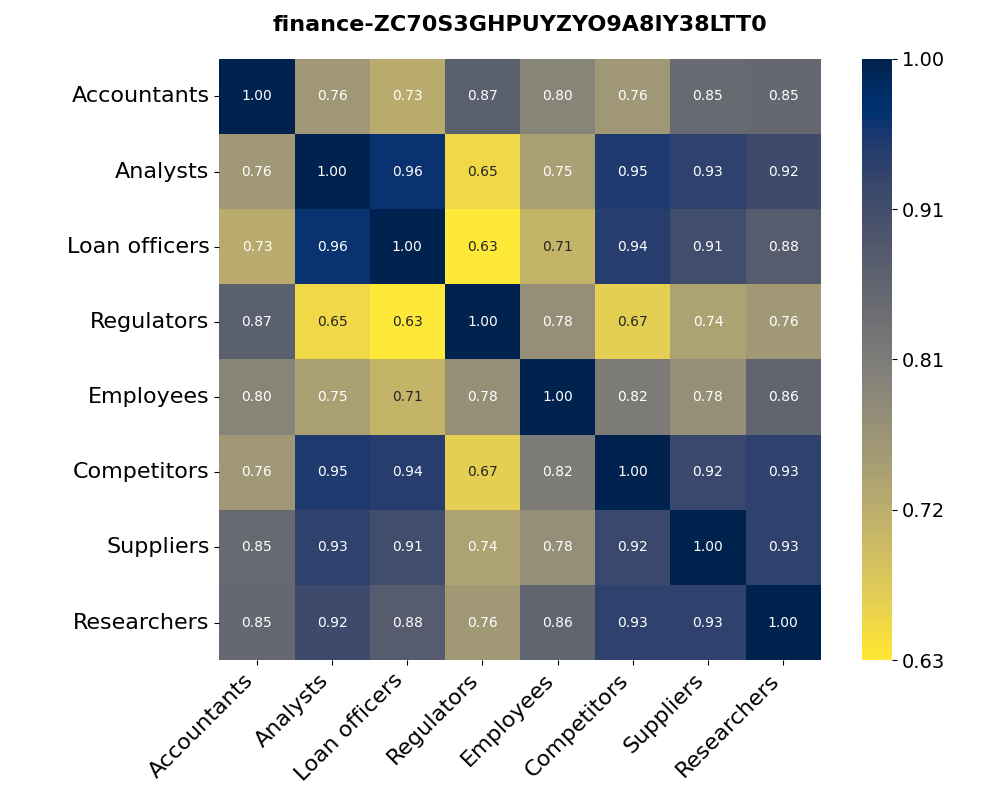}
    }
    \hfill
    \subfigure[Without Persona (Case 3)]{
        \includegraphics[width=0.48\textwidth]{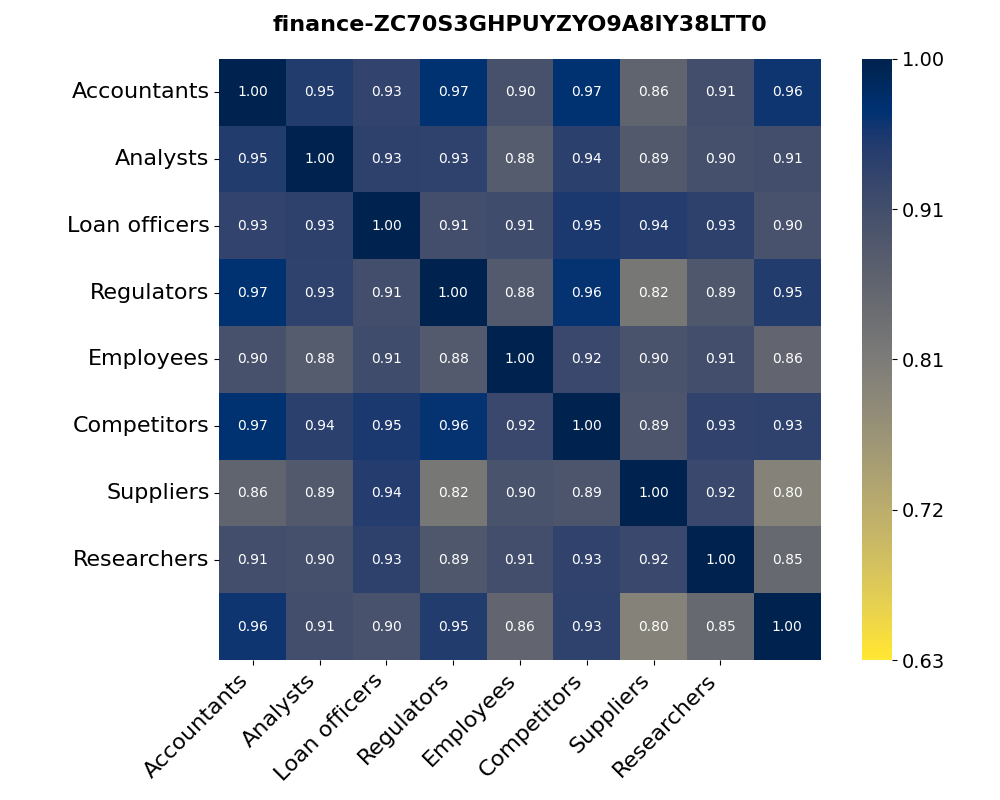}
    }
    
    \vspace{-5pt}
    \caption{Case 1-3: Document-level comparison of semantic similarities between SQs generated with and without persona across three different cases in \textbf{finance} domain.}
    \label{fig:similarity_comparison_app_finance_case_1_3}
\end{figure*}

\begin{figure*}[!ht]
    \centering
    \subfigure[With Persona (Case 4)]{
        \includegraphics[width=0.48\textwidth]{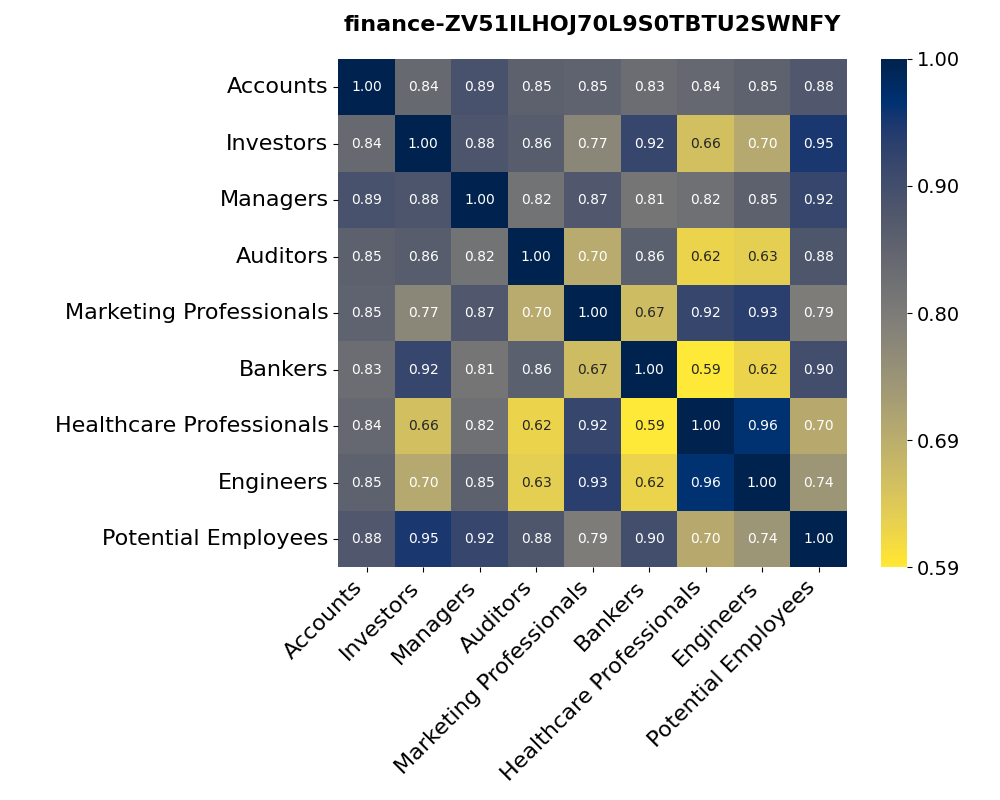}
    }
    \hfill
    \subfigure[Without Persona (Case 4)]{
        \includegraphics[width=0.48\textwidth]{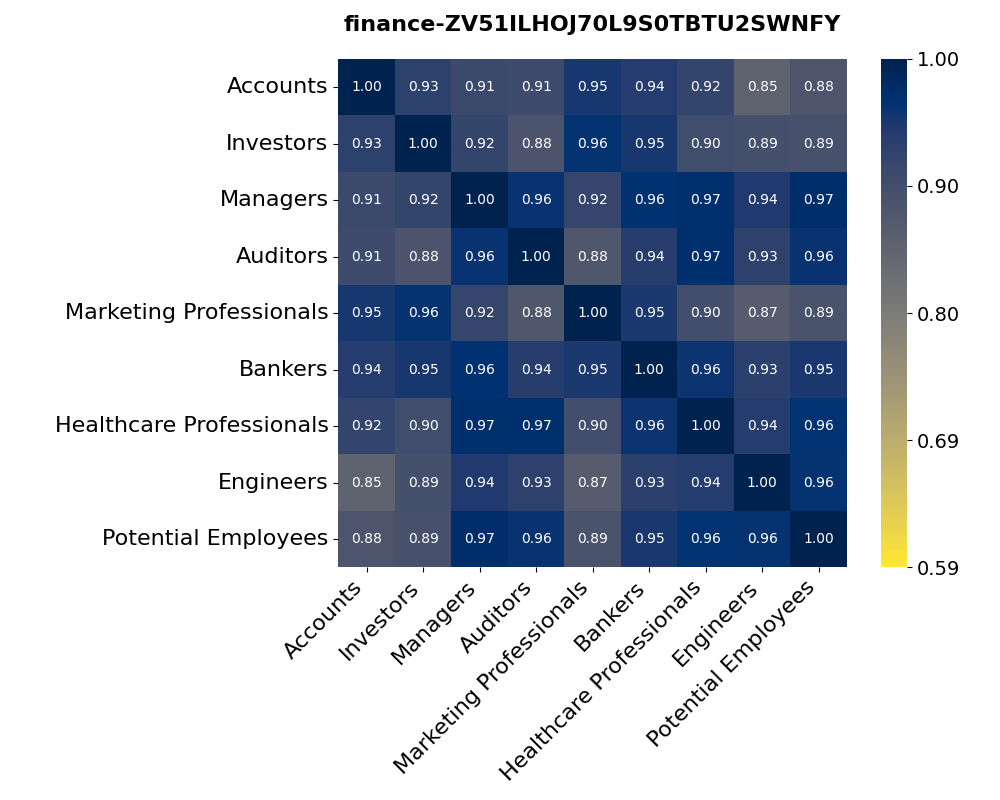}
    }
    
    \subfigure[With Persona (Case 5)]{
        \includegraphics[width=0.48\textwidth]{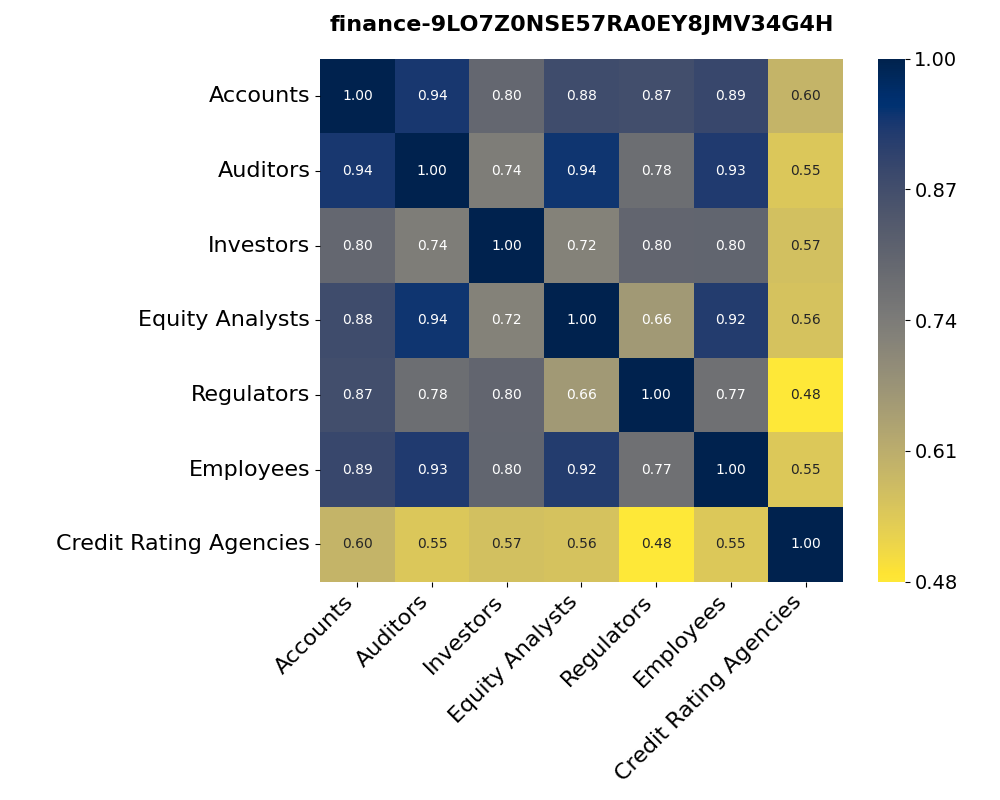}
    }
    \hfill
    \subfigure[Without Persona (Case 5)]{
        \includegraphics[width=0.48\textwidth]{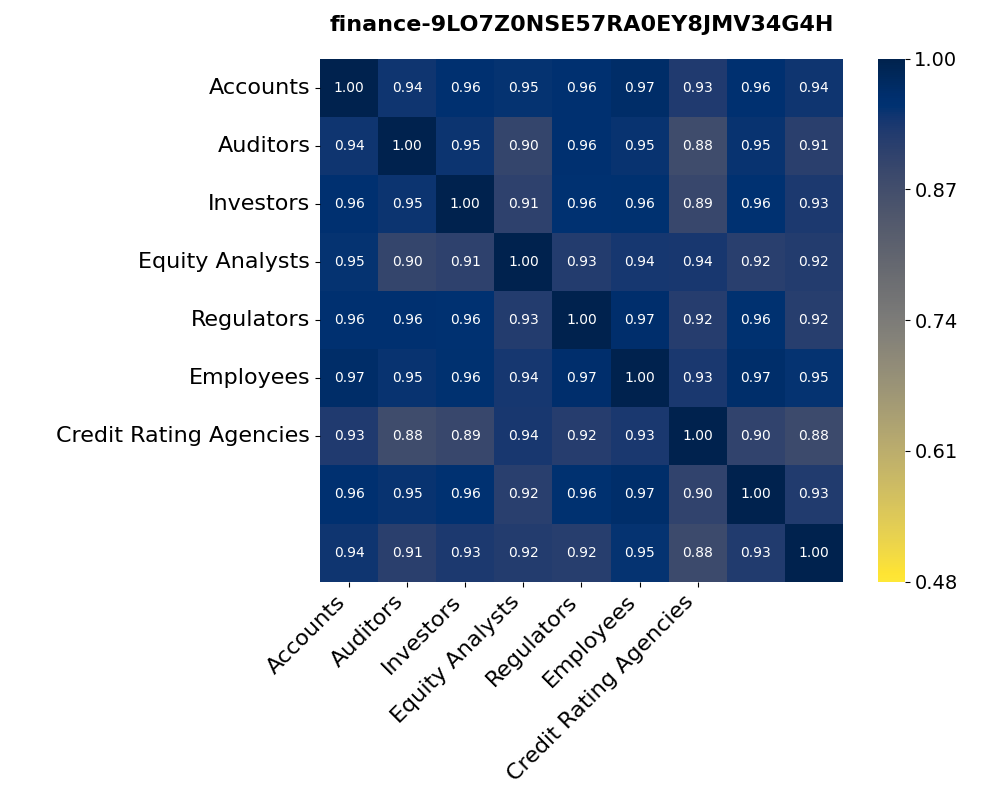}
    }
    
    \subfigure[With Persona (Case 6)]{
        \includegraphics[width=0.48\textwidth]{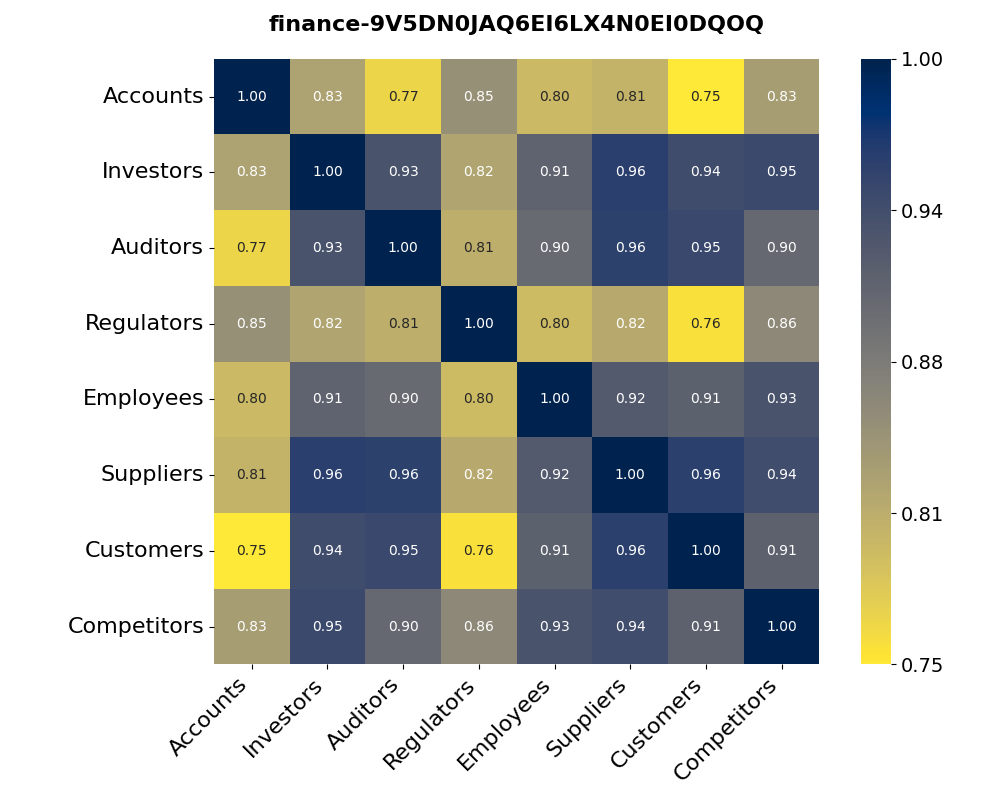}
    }
    \hfill
    \subfigure[Without Persona (Case 6)]{
        \includegraphics[width=0.48\textwidth]{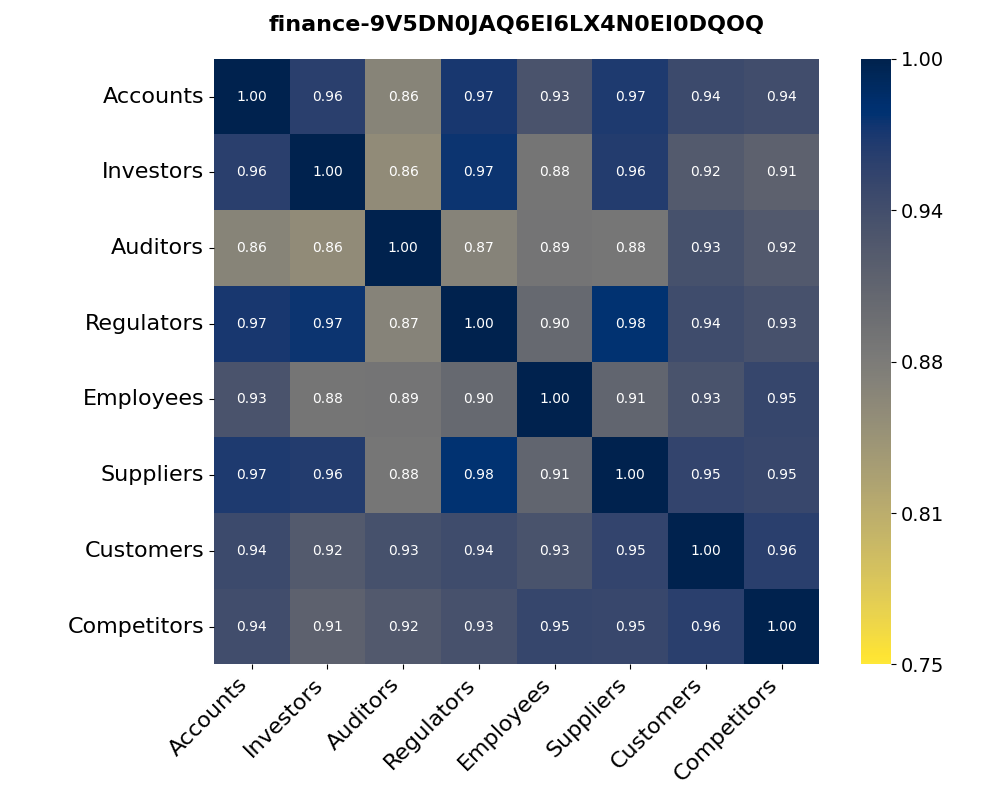}
    }
    
    \vspace{-5pt}
    \caption{Case 4-6: Document-level comparison of semantic similarities between SQs generated with and without persona across three different cases in \textbf{finance} domain.}
    \label{fig:similarity_comparison_app_finance_case_4_6}
\end{figure*}

\begin{figure*}[!ht]
    \centering
    \subfigure[With Persona (Case 1)]{
        \includegraphics[width=0.48\textwidth]{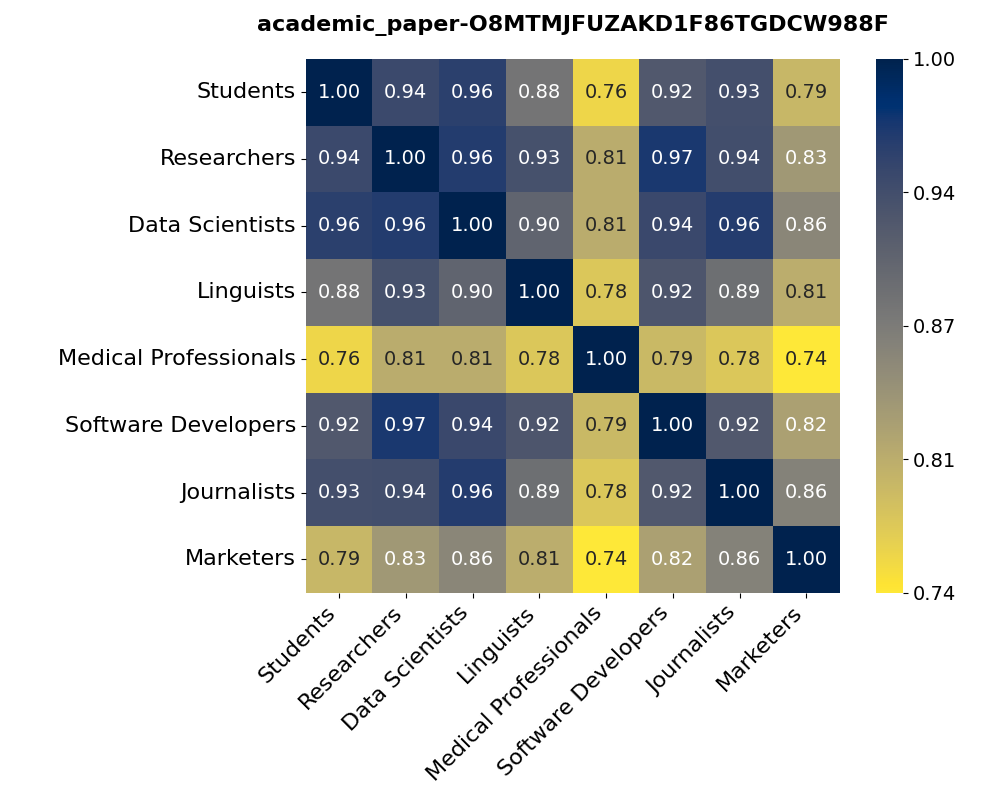}
    }
    \hfill
    \subfigure[Without Persona (Case 1)]{
        \includegraphics[width=0.48\textwidth]{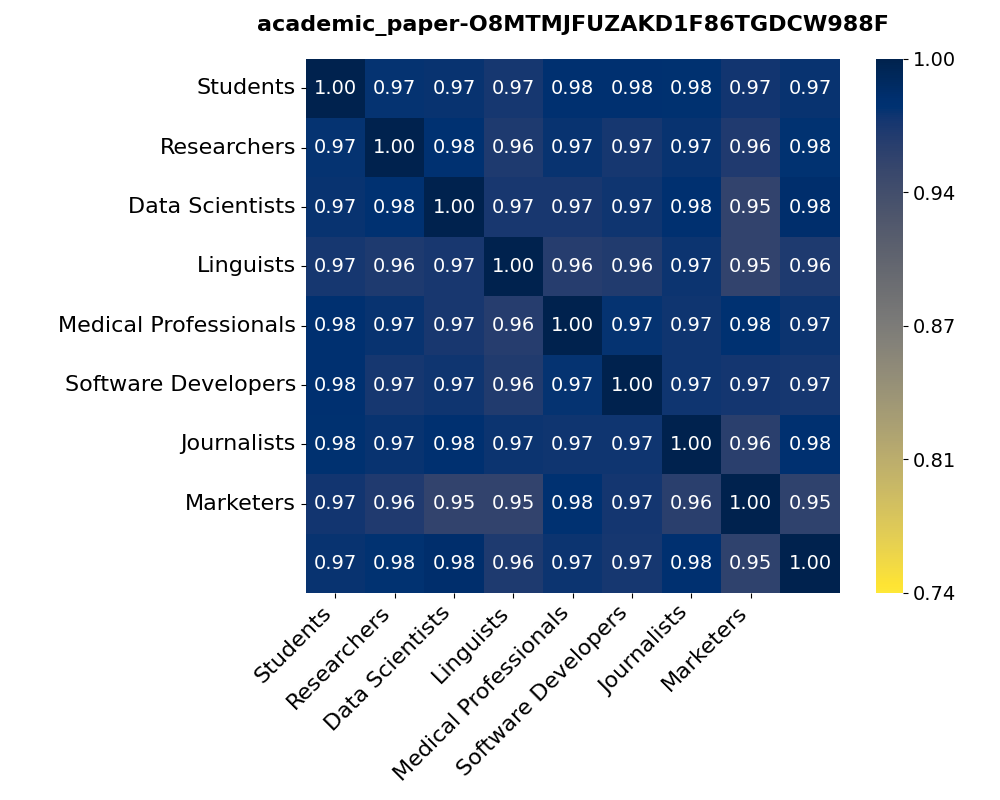}
    }
    
    \subfigure[With Persona (Case 2)]{
        \includegraphics[width=0.48\textwidth]{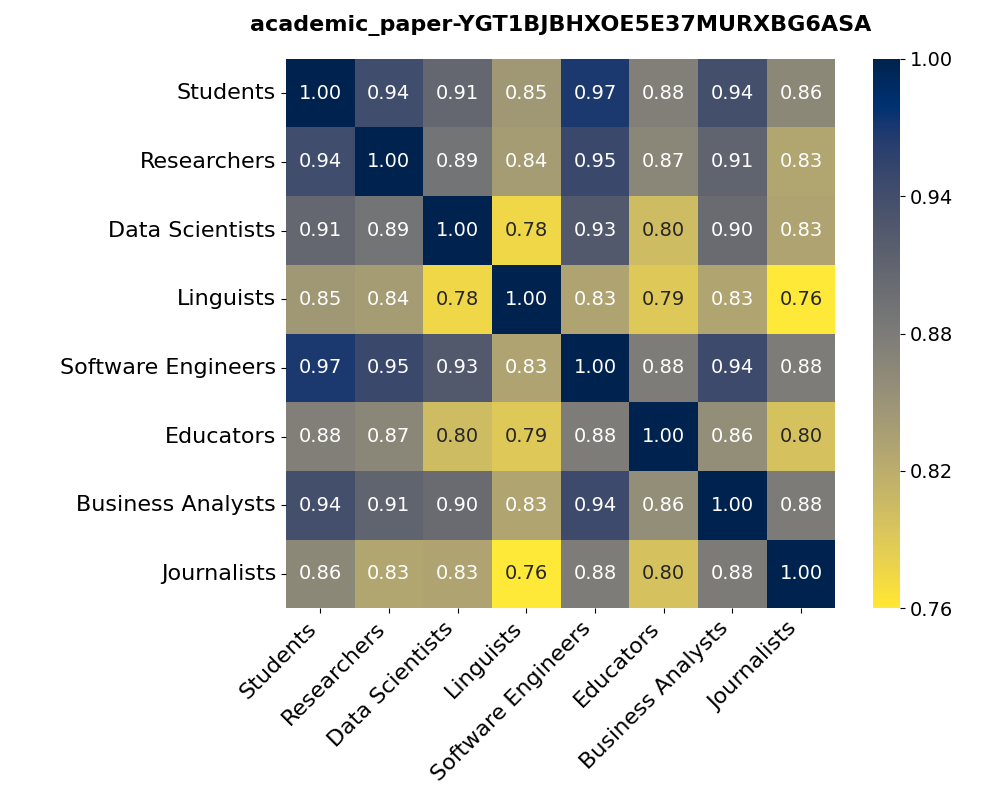}
    }
    \hfill
    \subfigure[Without Persona (Case 2)]{
        \includegraphics[width=0.48\textwidth]{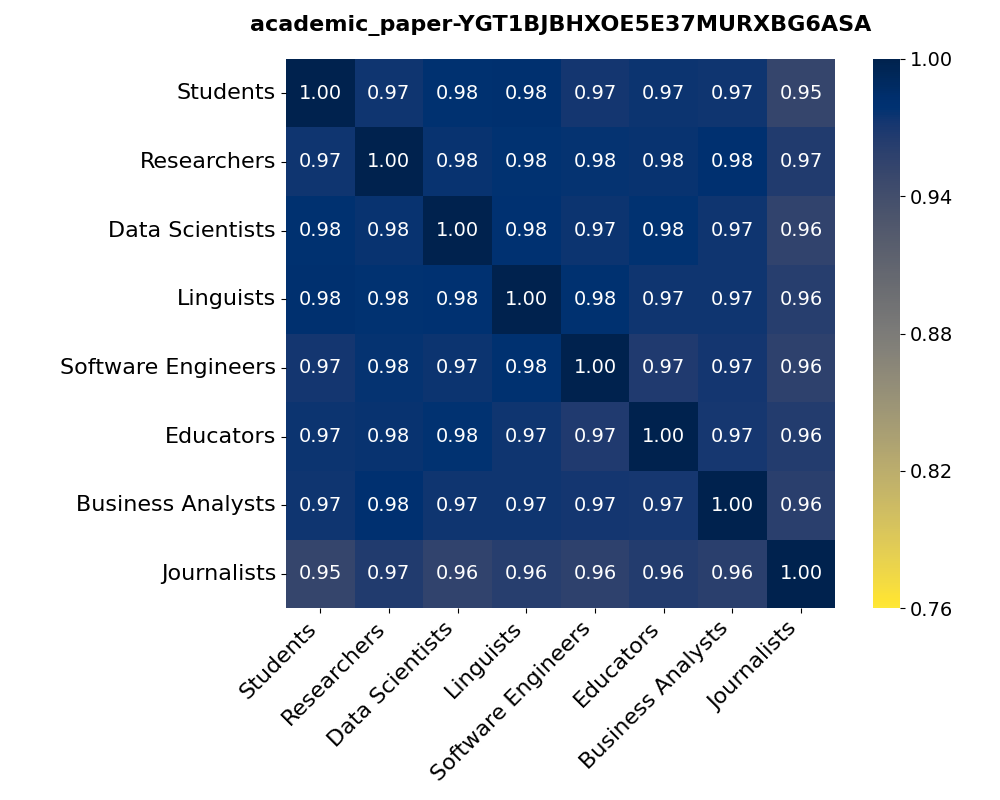}
    }
    
    \subfigure[With Persona (Case 3)]{
        \includegraphics[width=0.48\textwidth]{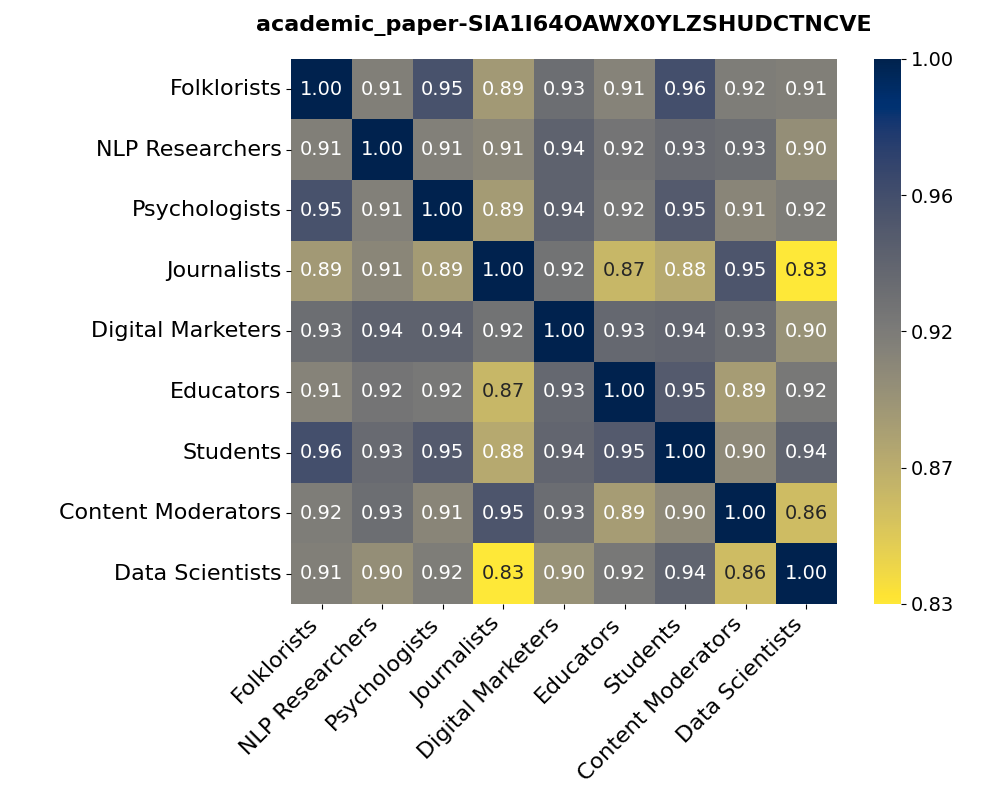}
    }
    \hfill
    \subfigure[Without Persona (Case 3)]{
        \includegraphics[width=0.48\textwidth]{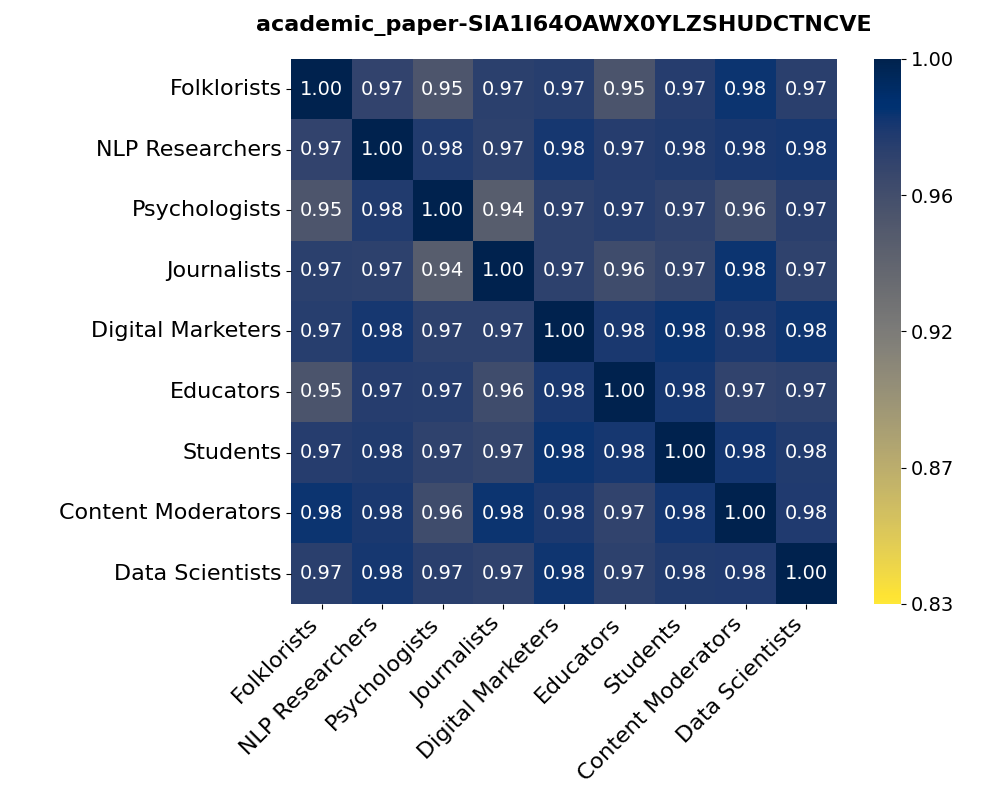}
    }
    
    \vspace{-5pt}
    \caption{Case 1-3: Document-level comparison of semantic similarities between SQs generated with and without persona across three different cases in \textbf{academia} domain.}
    \label{fig:similarity_comparison_app_academia_case_1_3}
\end{figure*}

\begin{figure*}[!ht]
    \centering
    \subfigure[With Persona (Case 4)]{
        \includegraphics[width=0.48\textwidth]{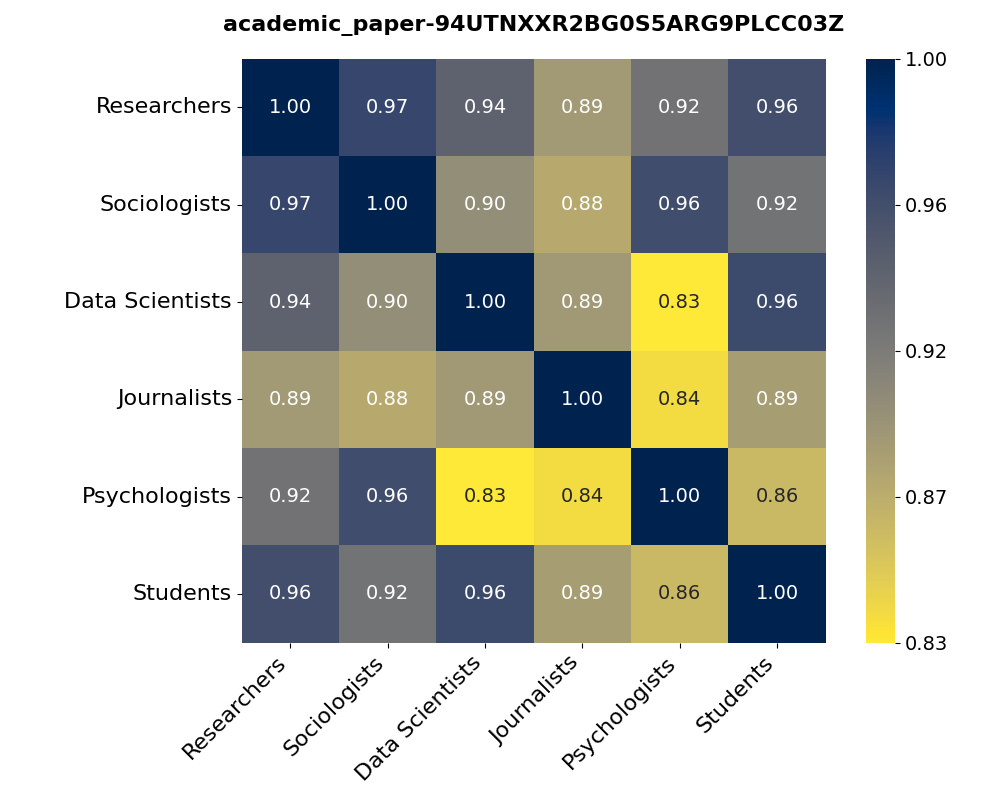}
    }
    \hfill
    \subfigure[Without Persona (Case 4)]{
        \includegraphics[width=0.48\textwidth]{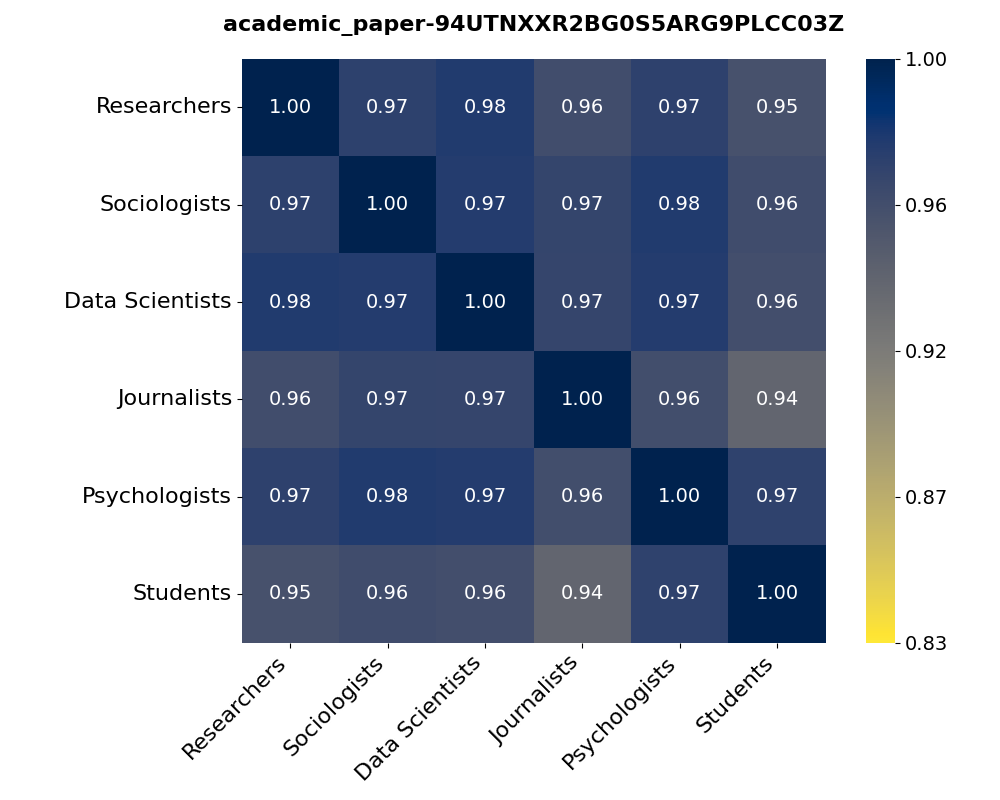}
    }
    
    \subfigure[With Persona (Case 5)]{
        \includegraphics[width=0.48\textwidth]{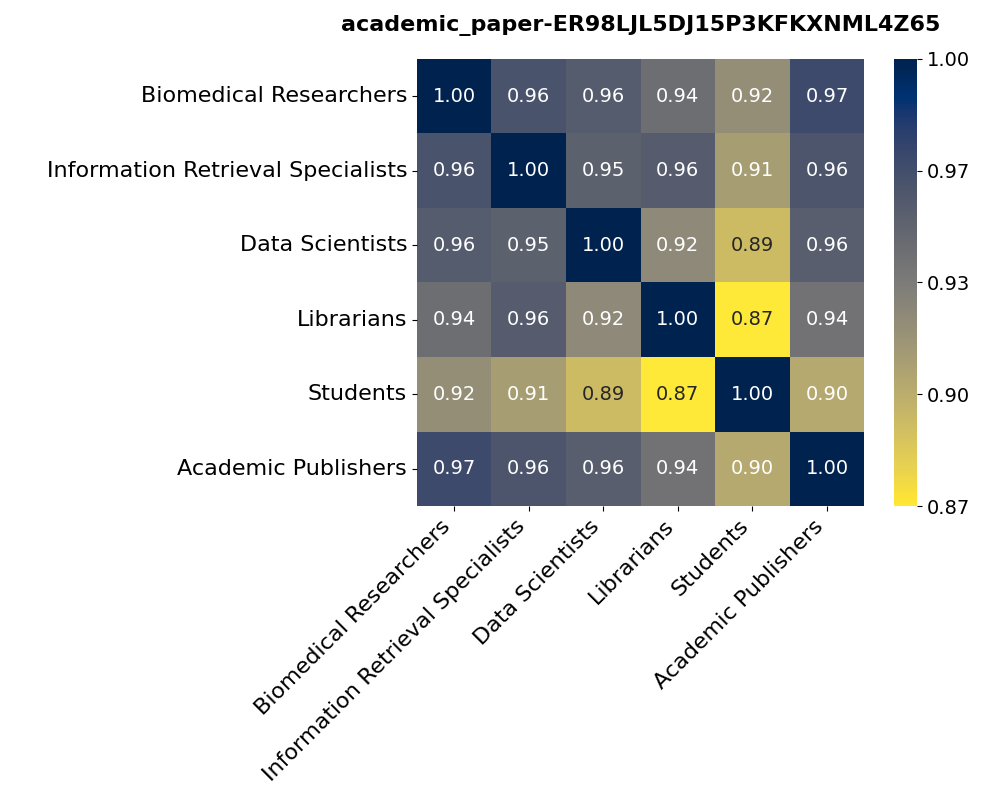}
    }
    \hfill
    \subfigure[Without Persona (Case 5)]{
        \includegraphics[width=0.48\textwidth]{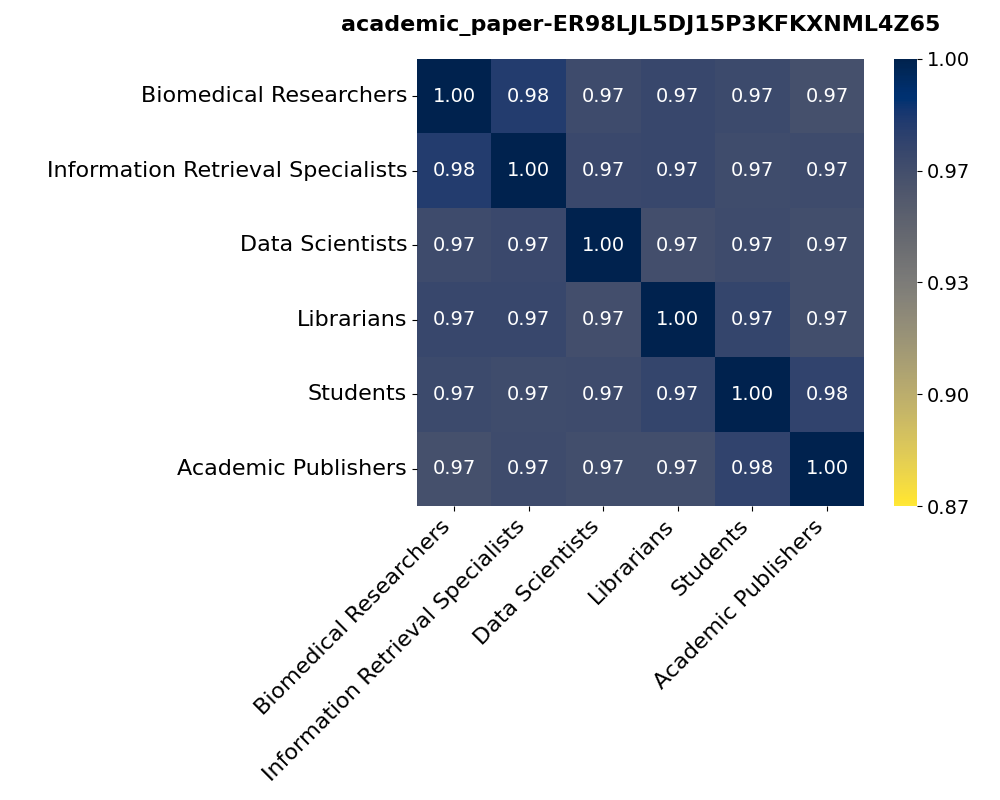}
    }
    
    \subfigure[With Persona (Case 6)]{
        \includegraphics[width=0.48\textwidth]{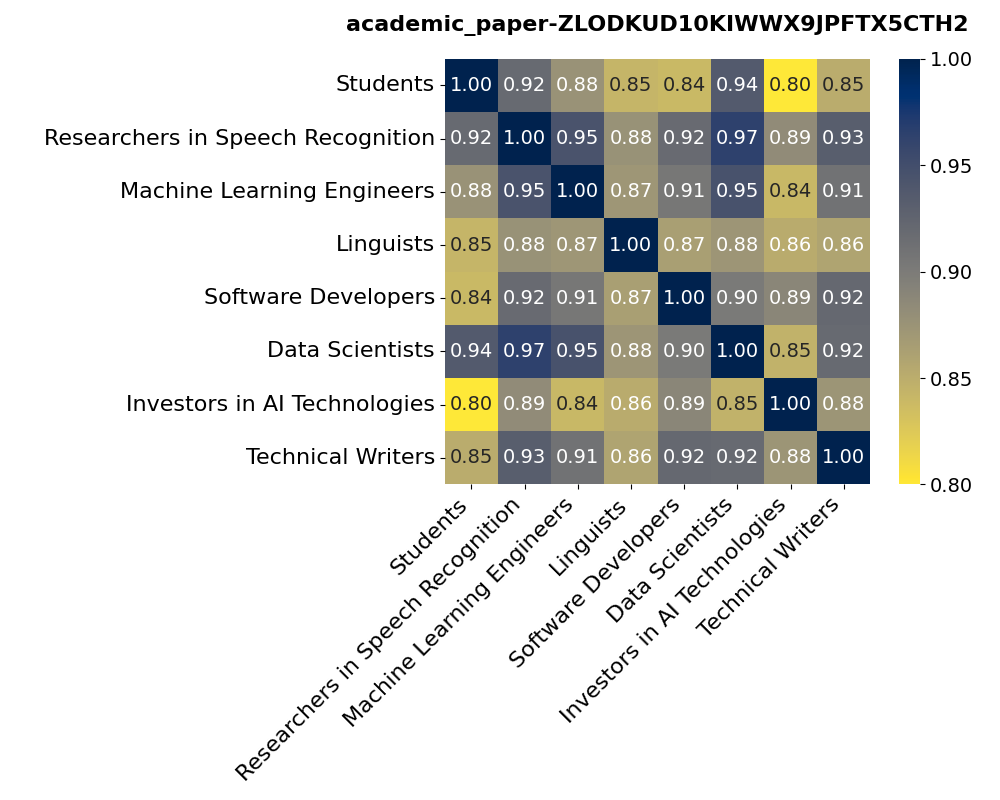}
    }
    \hfill
    \subfigure[Without Persona (Case 6)]{
        \includegraphics[width=0.48\textwidth]{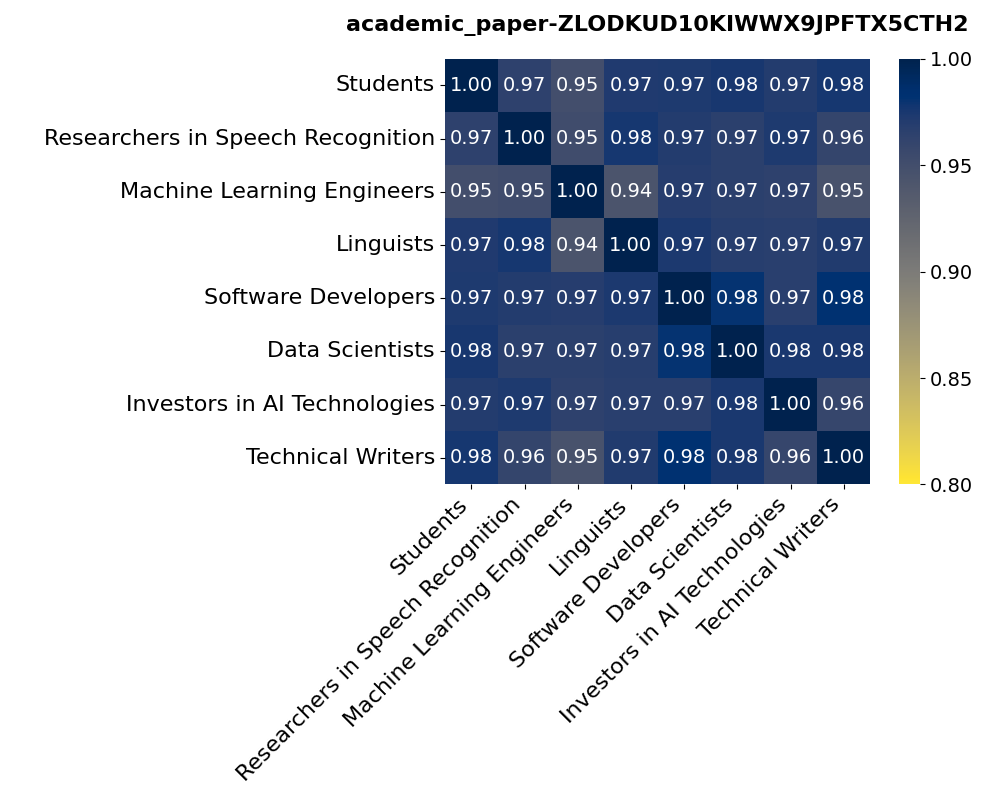}
    }
    
    \vspace{-5pt}
    \caption{Case 4-6: Document-level comparison of semantic similarities between SQs generated with and without persona across three different cases in \textbf{academia} domain.}
    \label{fig:similarity_comparison_app_academia_case_4_6}
\end{figure*}

We display more examples of visualized question semantic diversity from the three domains of legal, finance, and academia in Figure \ref{fig:similarity_comparison_app_legal_case_1_3}, \ref{fig:similarity_comparison_app_legal_case_4_6}, Figure \ref{fig:similarity_comparison_app_finance_case_1_3}, \ref{fig:similarity_comparison_app_finance_case_4_6}, and Figure \ref{fig:similarity_comparison_app_academia_case_1_3}, \ref{fig:similarity_comparison_app_academia_case_4_6} respectively. 

\subsection{Persona Distribution}
\begin{figure*}[!ht]
    \centering
    \subfigure[Case 1]{
        \includegraphics[width=0.8\textwidth]{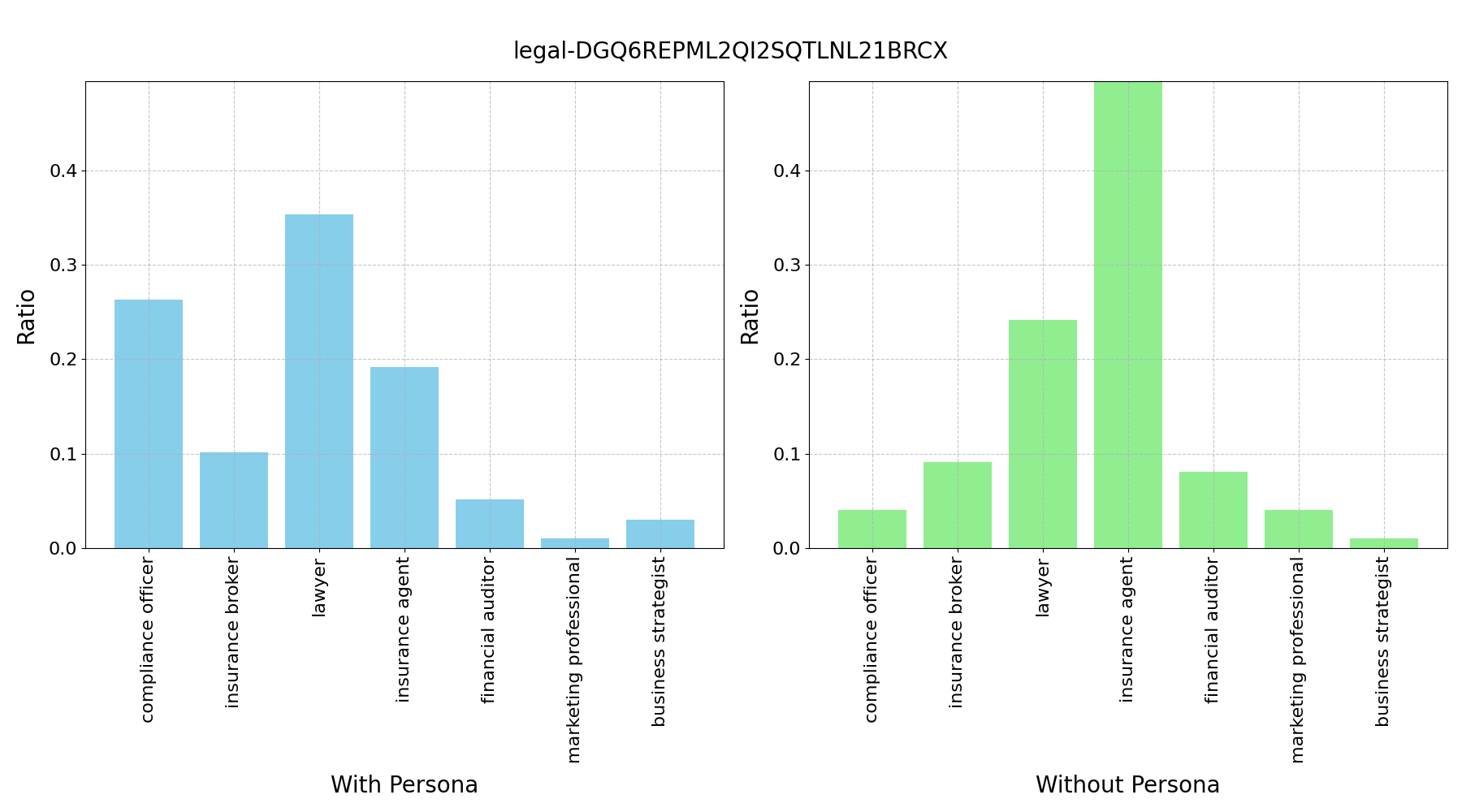}
    }
    \hfill
    \subfigure[Case 2]{
        \includegraphics[width=0.8\textwidth]{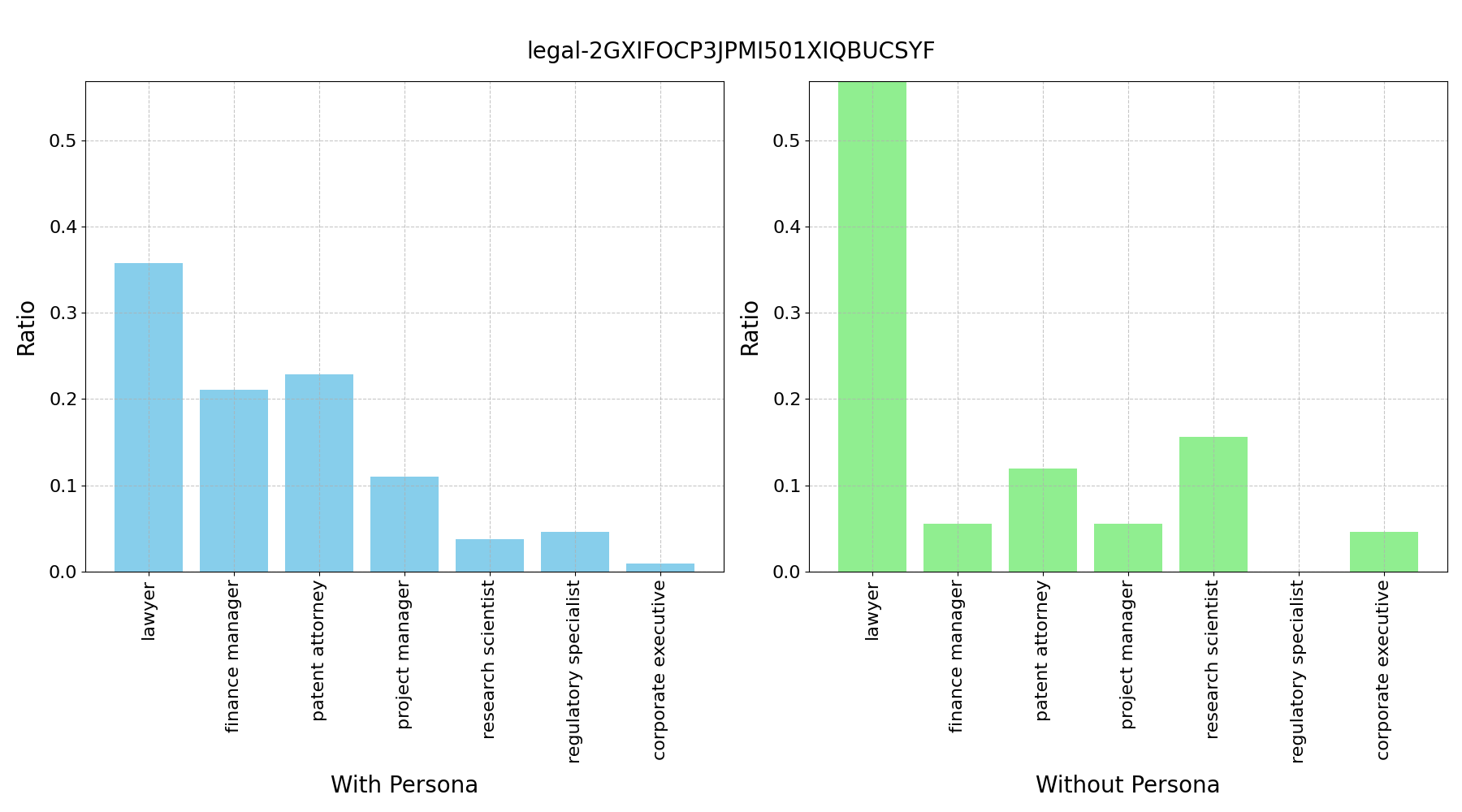}
    }
    
    \subfigure[Case 3]{
        \includegraphics[width=0.8\textwidth]{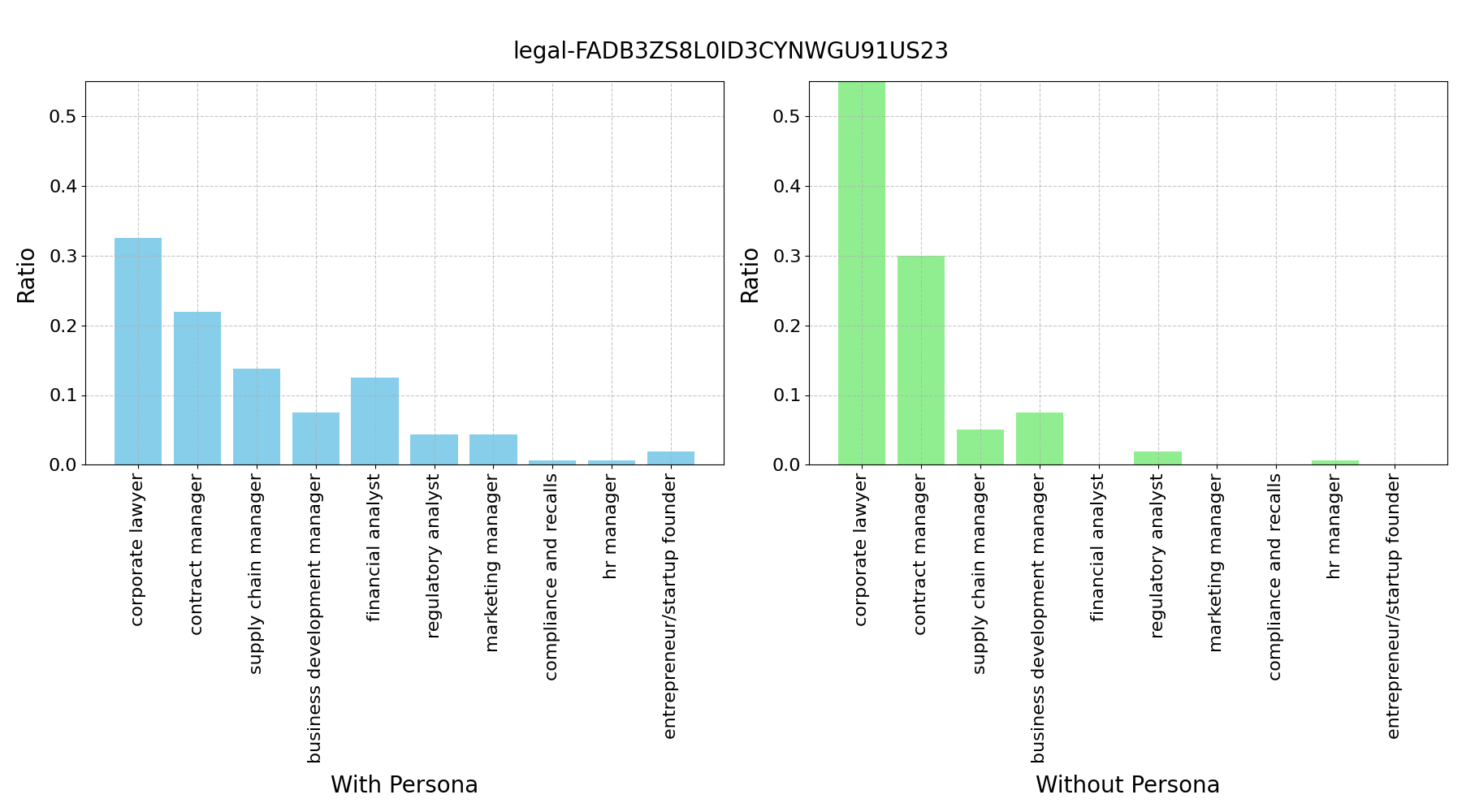}
    }
    
    \vspace{-5pt}
    \caption{Case 1-3: Document-level comparison of persona distribution between SQs generated with and without persona across three different cases in \textbf{legal} domain.}
    \label{fig:persona-distribution-legal-1-3}
\end{figure*}

\begin{figure*}[!ht]
    \centering
    \subfigure[Case 4]{
        \includegraphics[width=0.8\textwidth]{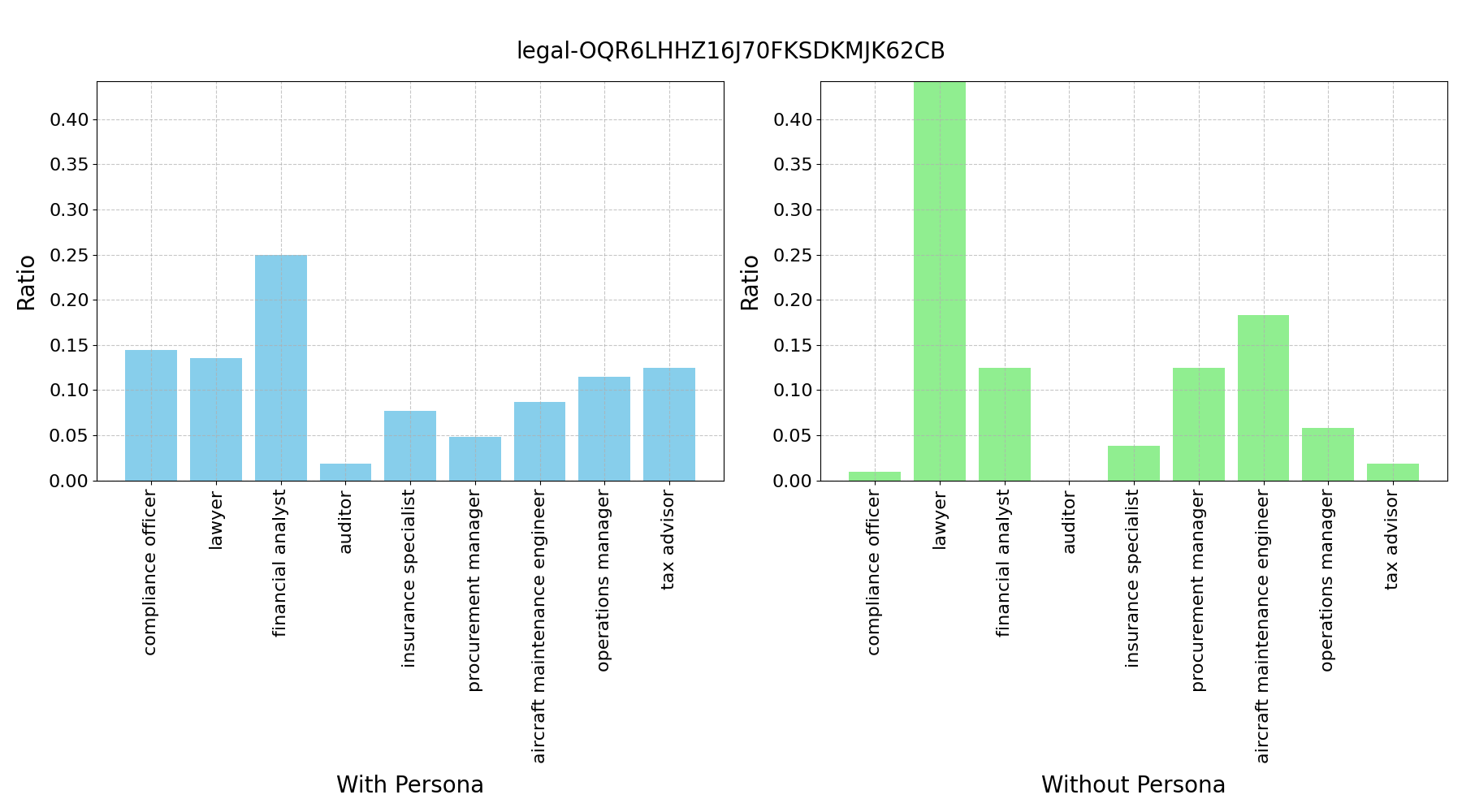}
    }

    \subfigure[Case 5)]{
        \includegraphics[width=0.8\textwidth]{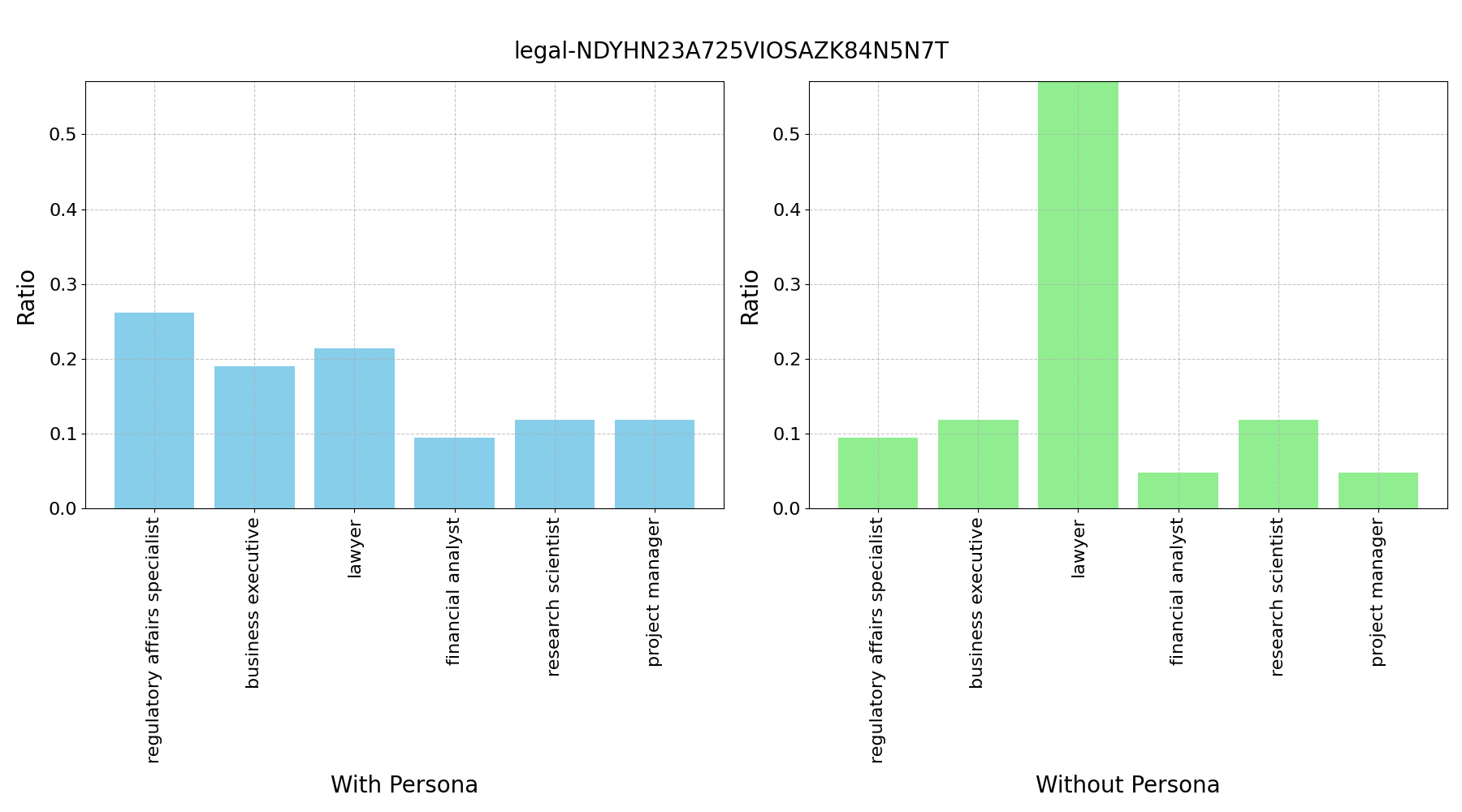}
    }
    
    \subfigure[Case 6]{
        \includegraphics[width=0.8\textwidth]{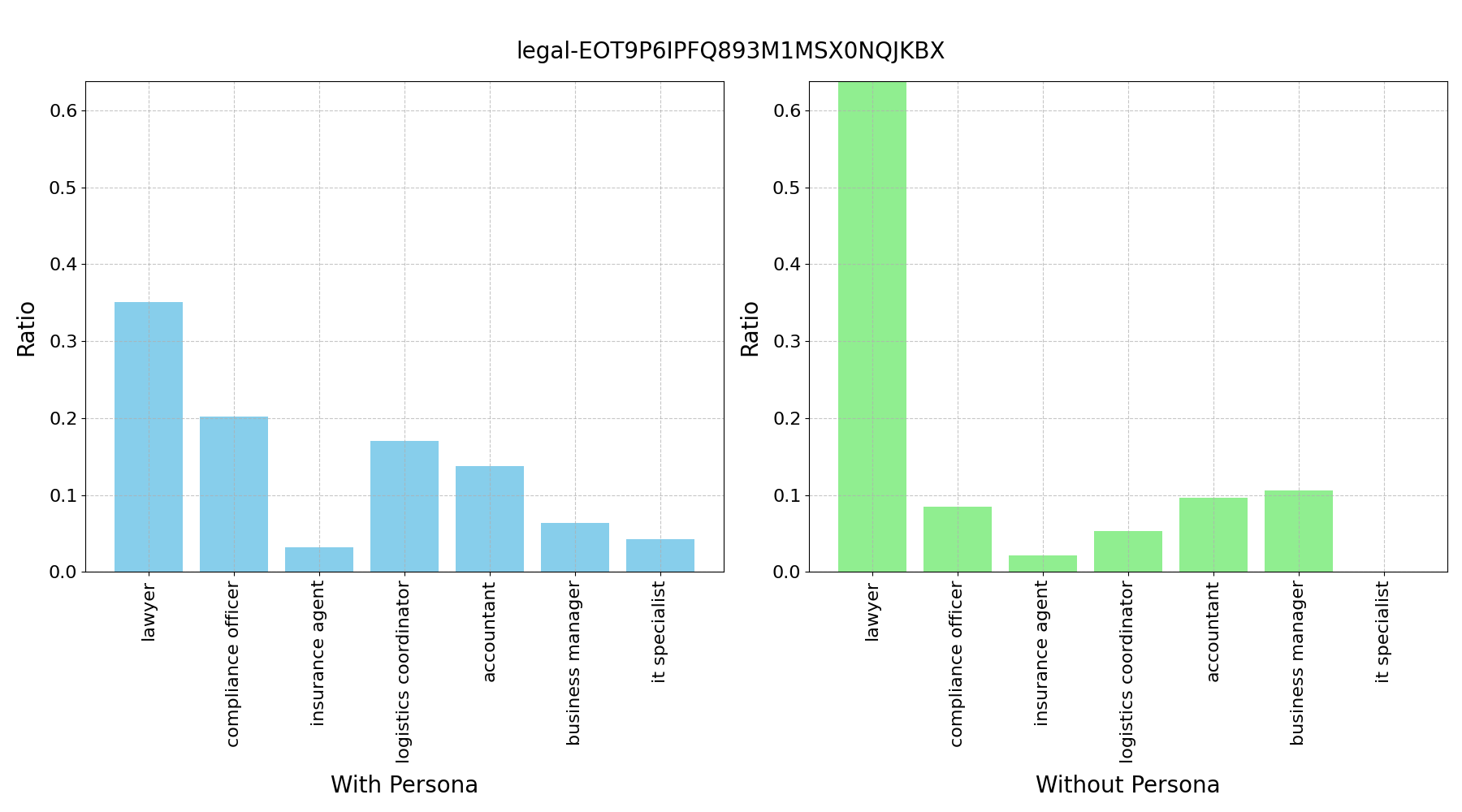}
    }
    
    \vspace{-5pt}
    \caption{Case 4-6: Document-level comparison of persona distribution between SQs generated with and without persona across three different cases in \textbf{legal} domain.}
    \label{fig:persona-distribution-legal-4-6}
\end{figure*}

\begin{figure*}[!ht]
    \centering
    \subfigure[Case 1]{
        \includegraphics[width=0.8\textwidth]{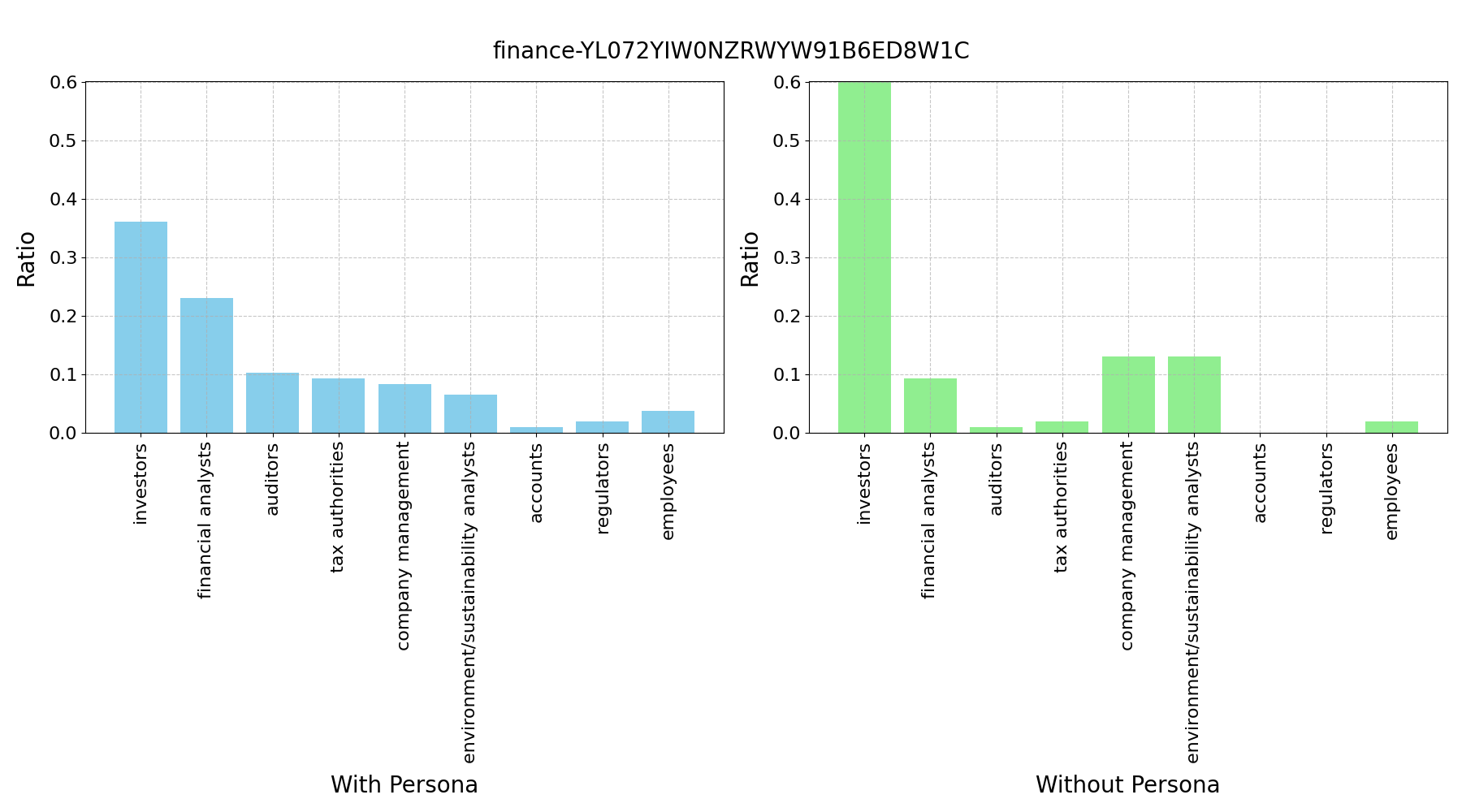}
    }
    \hfill
    \subfigure[Case 2]{
        \includegraphics[width=0.8\textwidth]{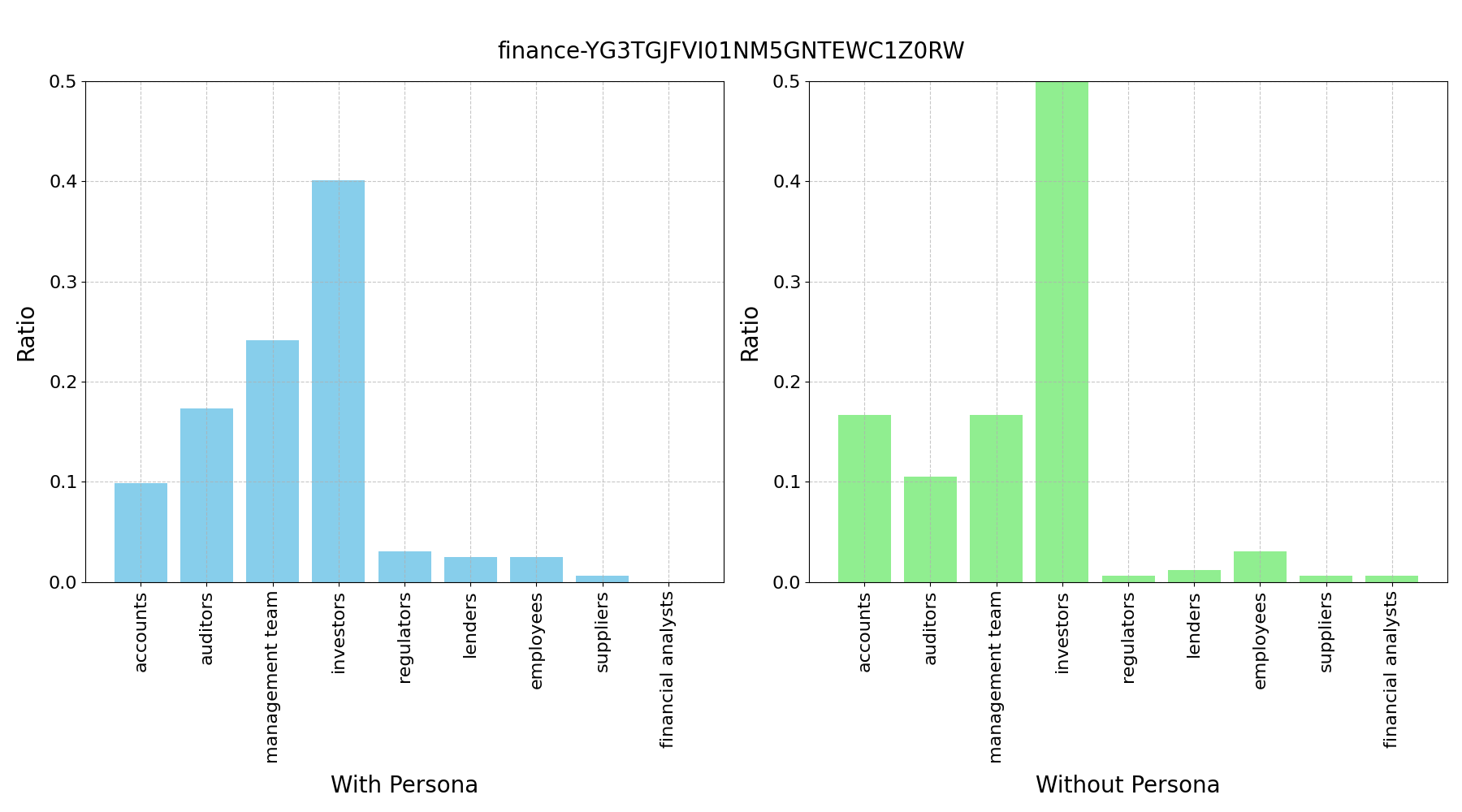}
    }
    
    \subfigure[Case 3]{
        \includegraphics[width=0.8\textwidth]{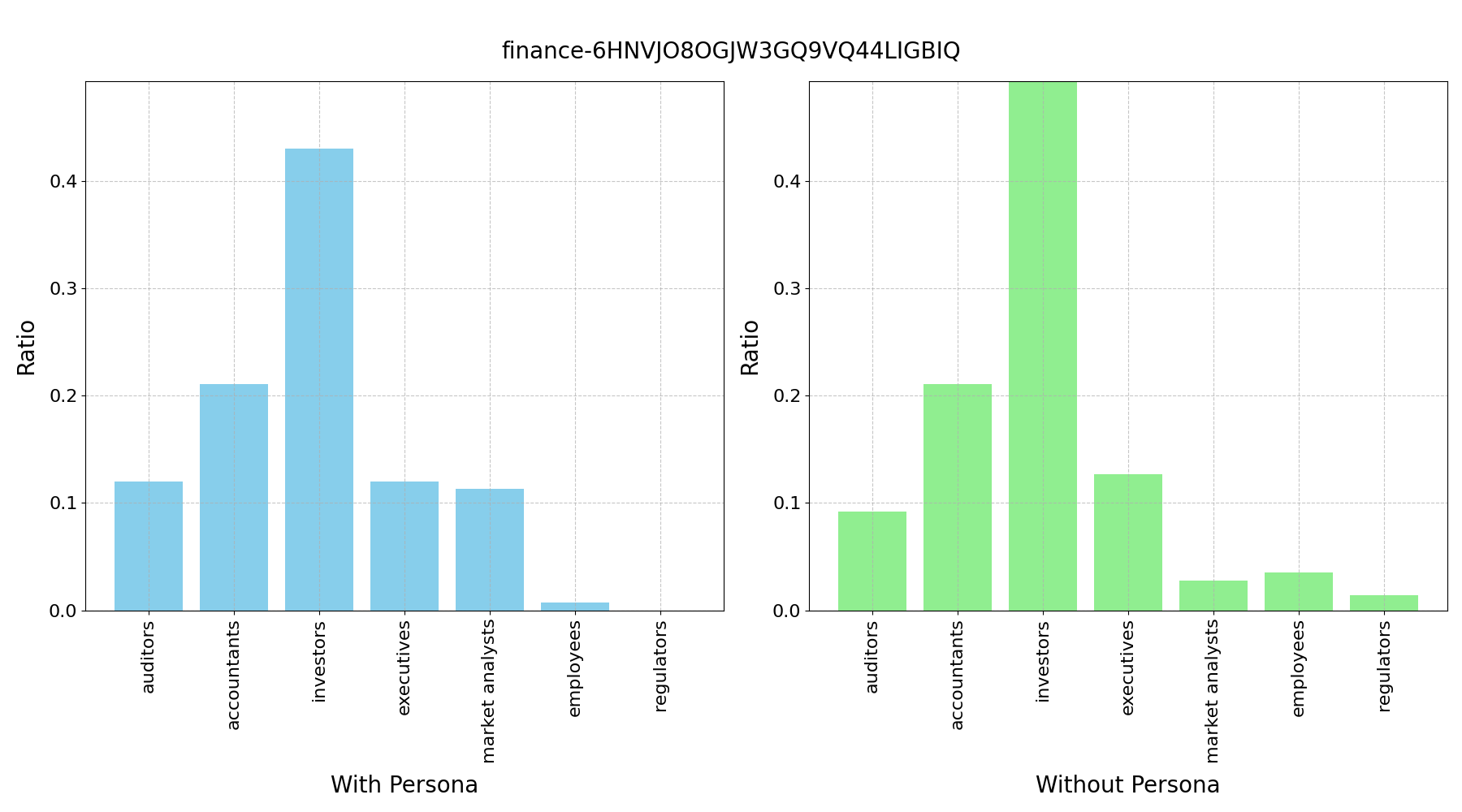}
    }
    
    \vspace{-5pt}
    \caption{Case 1-3: Document-level comparison of persona distribution between SQs generated with and without persona across three different cases in \textbf{finance} domain.}
    \label{fig:persona-distribution-finance-1-3}
\end{figure*}

\begin{figure*}[!ht]
    \centering
    \subfigure[Case 4]{
        \includegraphics[width=0.8\textwidth]{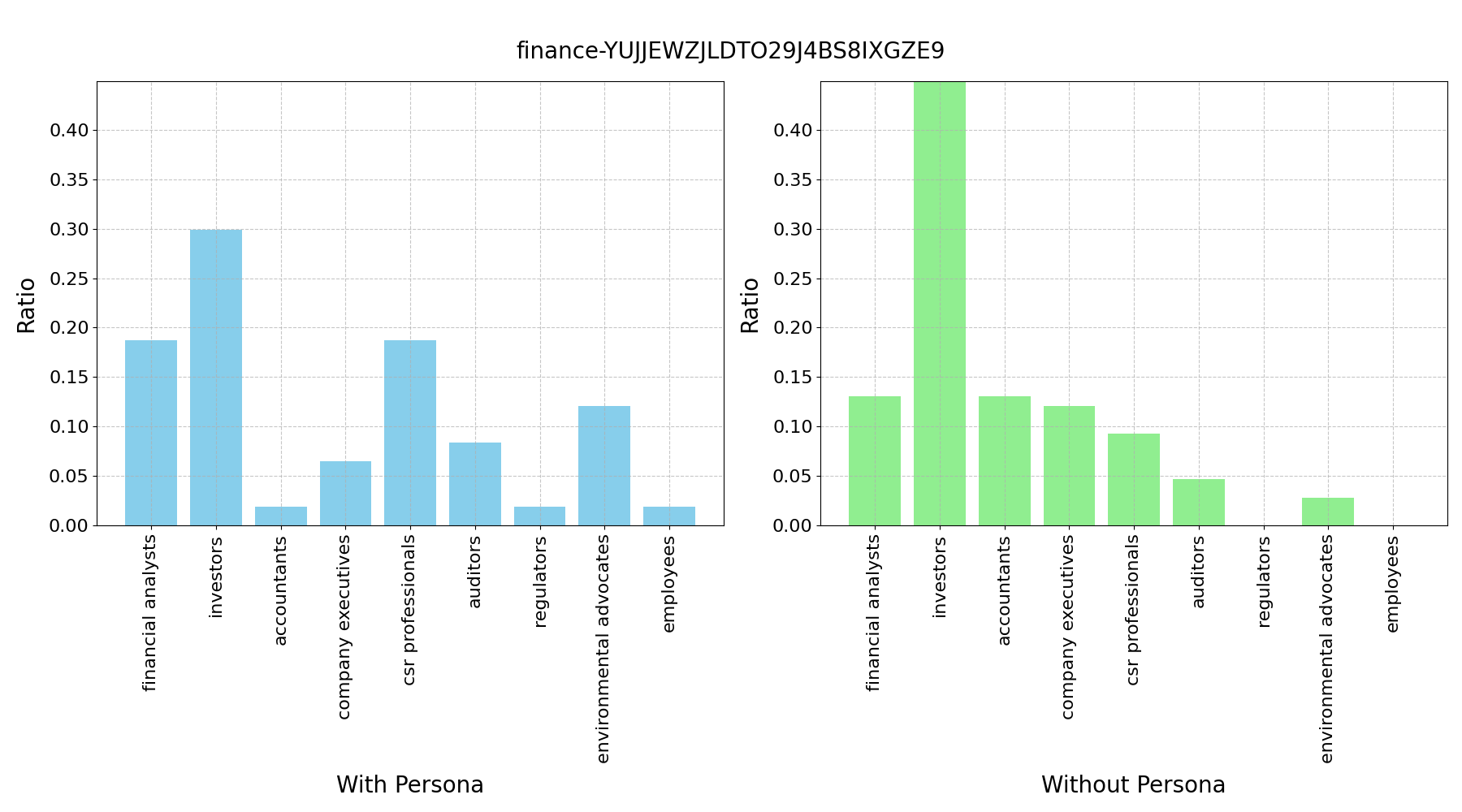}
    }

    \subfigure[Case 5)]{
        \includegraphics[width=0.8\textwidth]{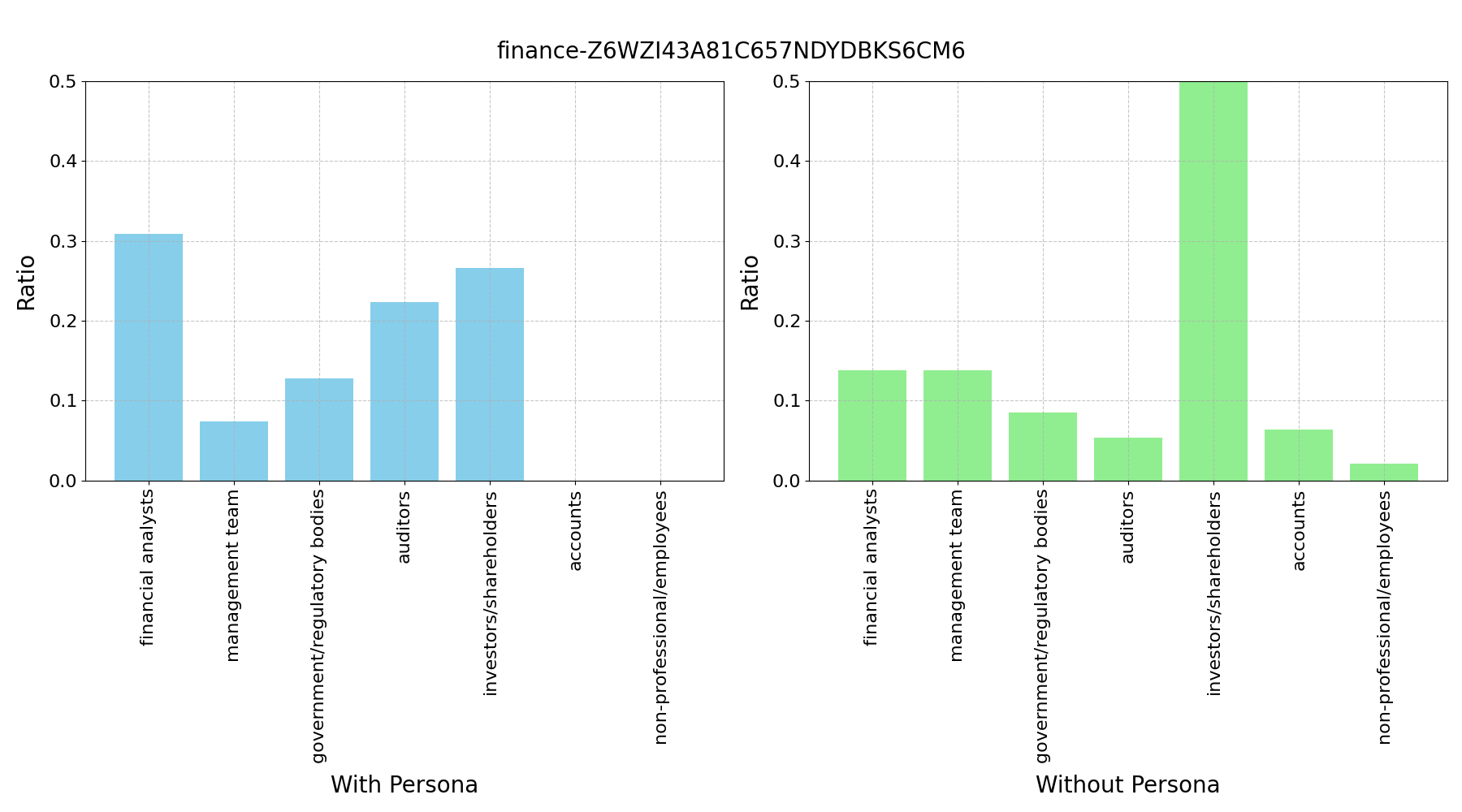}
    }
    
    \subfigure[Case 6]{
        \includegraphics[width=0.8\textwidth]{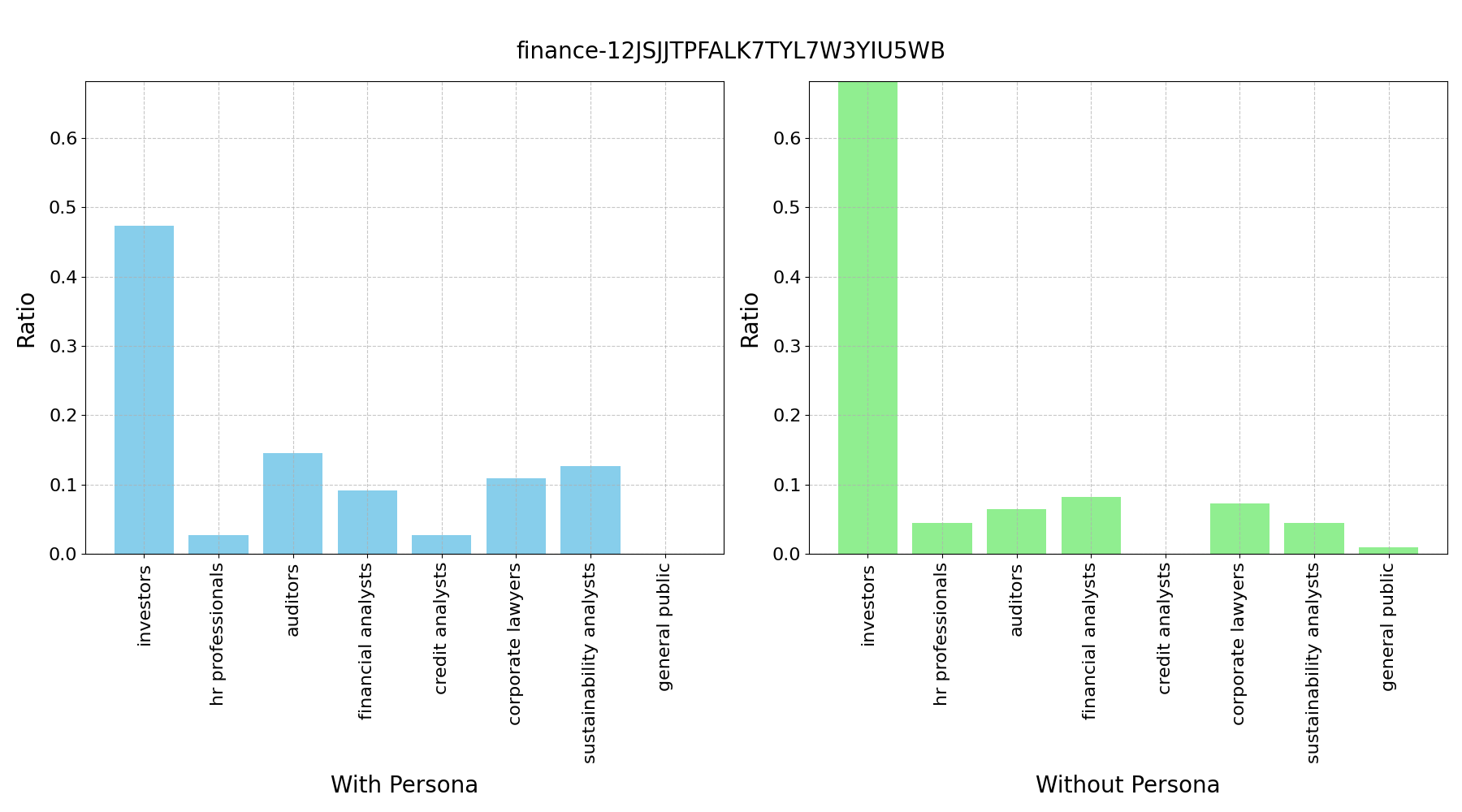}
    }
    
    \vspace{-5pt}
    \caption{Case 4-6: Document-level comparison of persona distribution between SQs generated with and without persona across three different cases in \textbf{finance} domain.}
    \label{fig:persona-distribution-finance-4-6}
\end{figure*}

\begin{figure*}[!ht]
    \centering
    \subfigure[Case 1]{
        \includegraphics[width=0.8\textwidth]{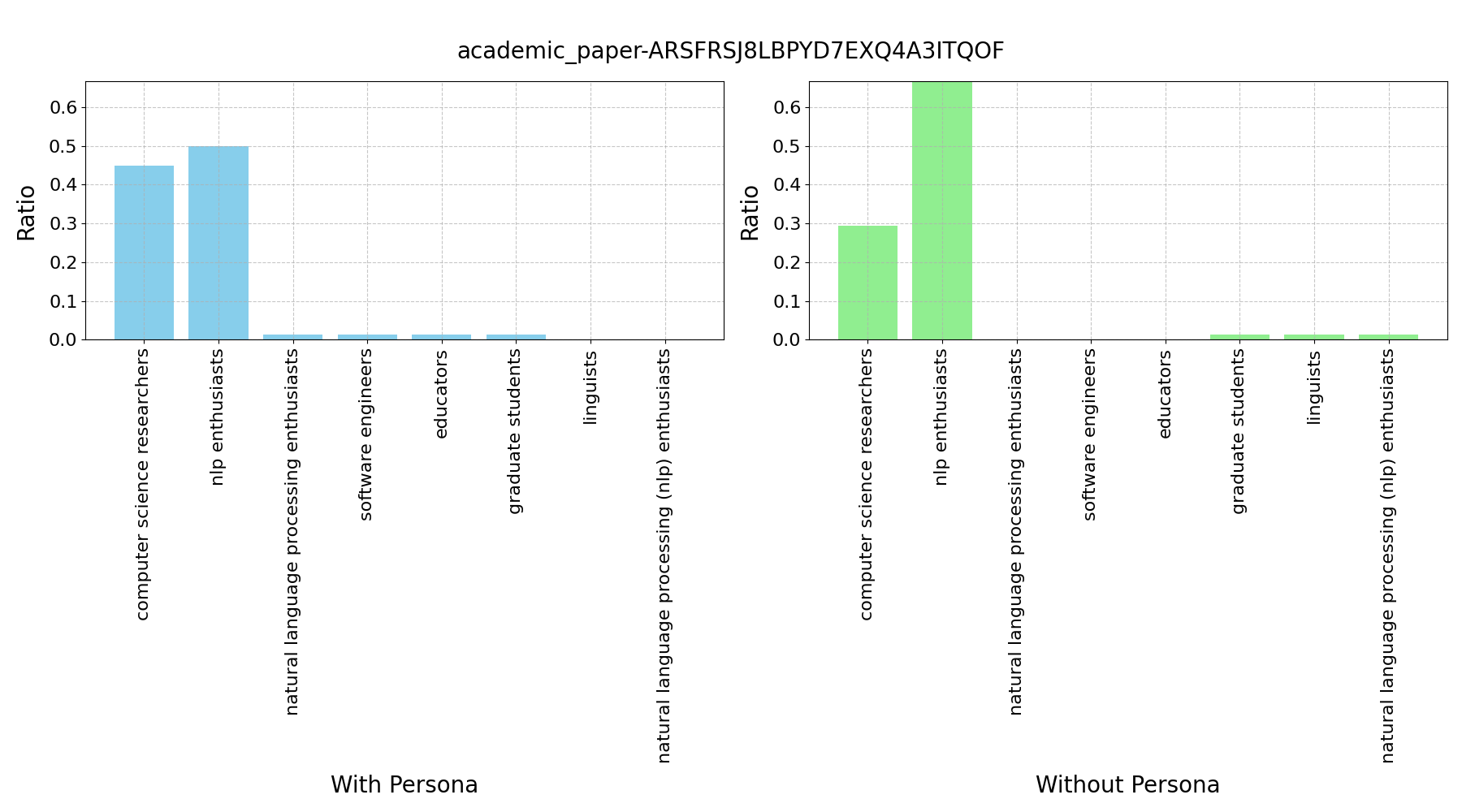}
    }
    \hfill
    \subfigure[Case 2]{
        \includegraphics[width=0.8\textwidth]{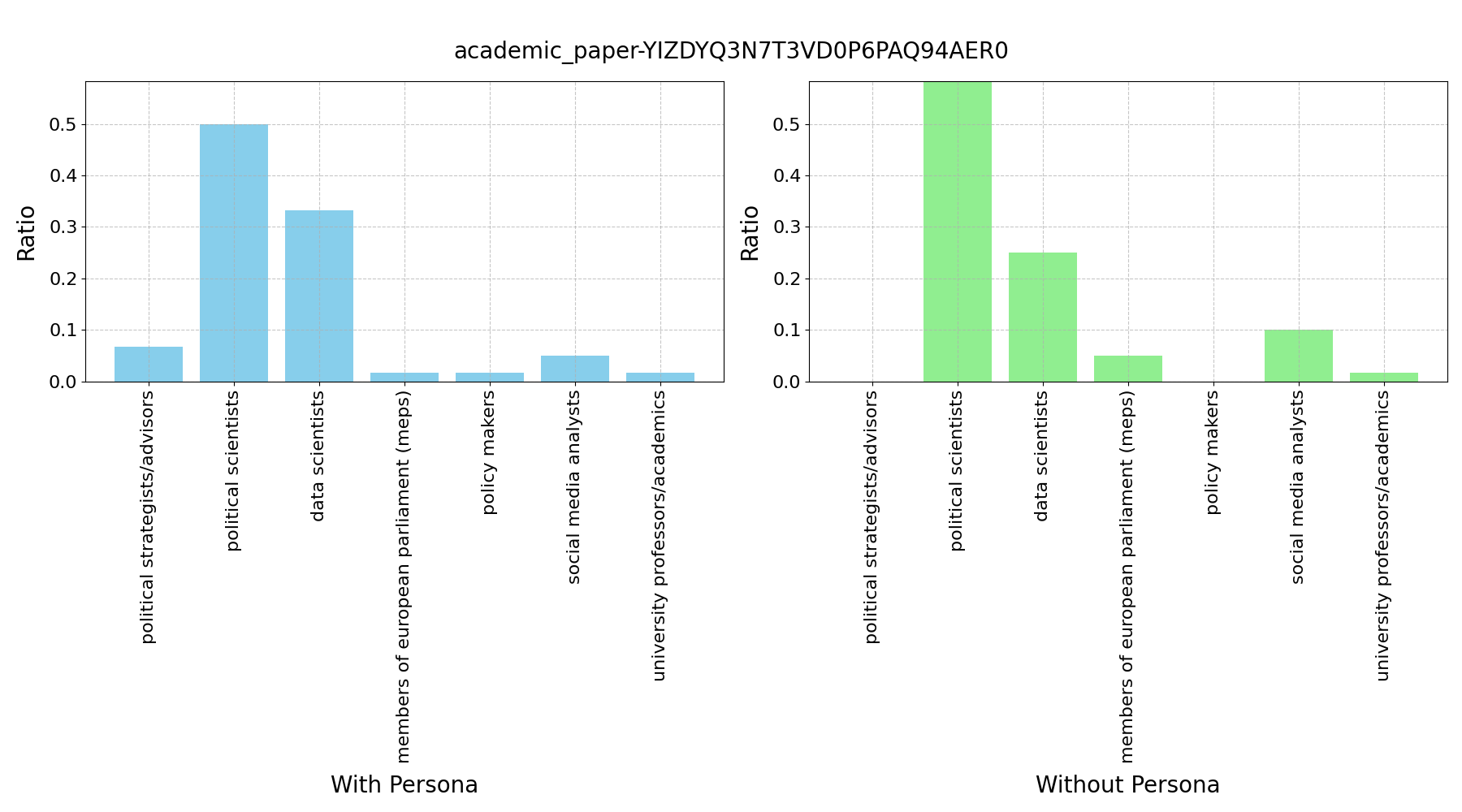}
    }
    
    \subfigure[Case 3]{
        \includegraphics[width=0.8\textwidth]{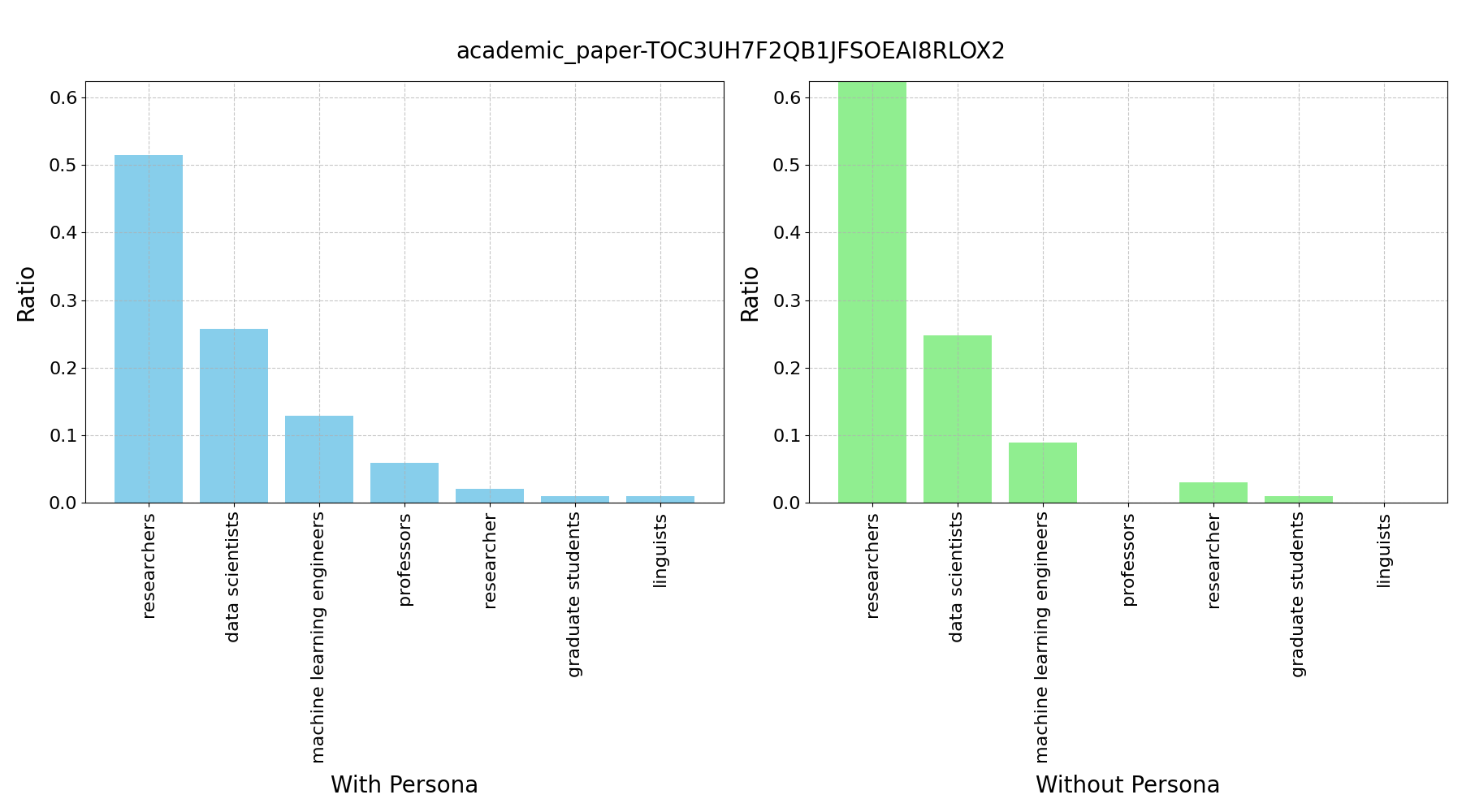}
    }
    
    \vspace{-5pt}
    \caption{Case 1-3: Document-level comparison of persona distribution between SQs generated with and without persona across three different cases in \textbf{academia} domain.}
    \label{fig:persona-distribution-academia-1-3}
\end{figure*}

\begin{figure*}[!ht]
    \centering
    \subfigure[Case 4]{
        \includegraphics[width=0.8\textwidth]{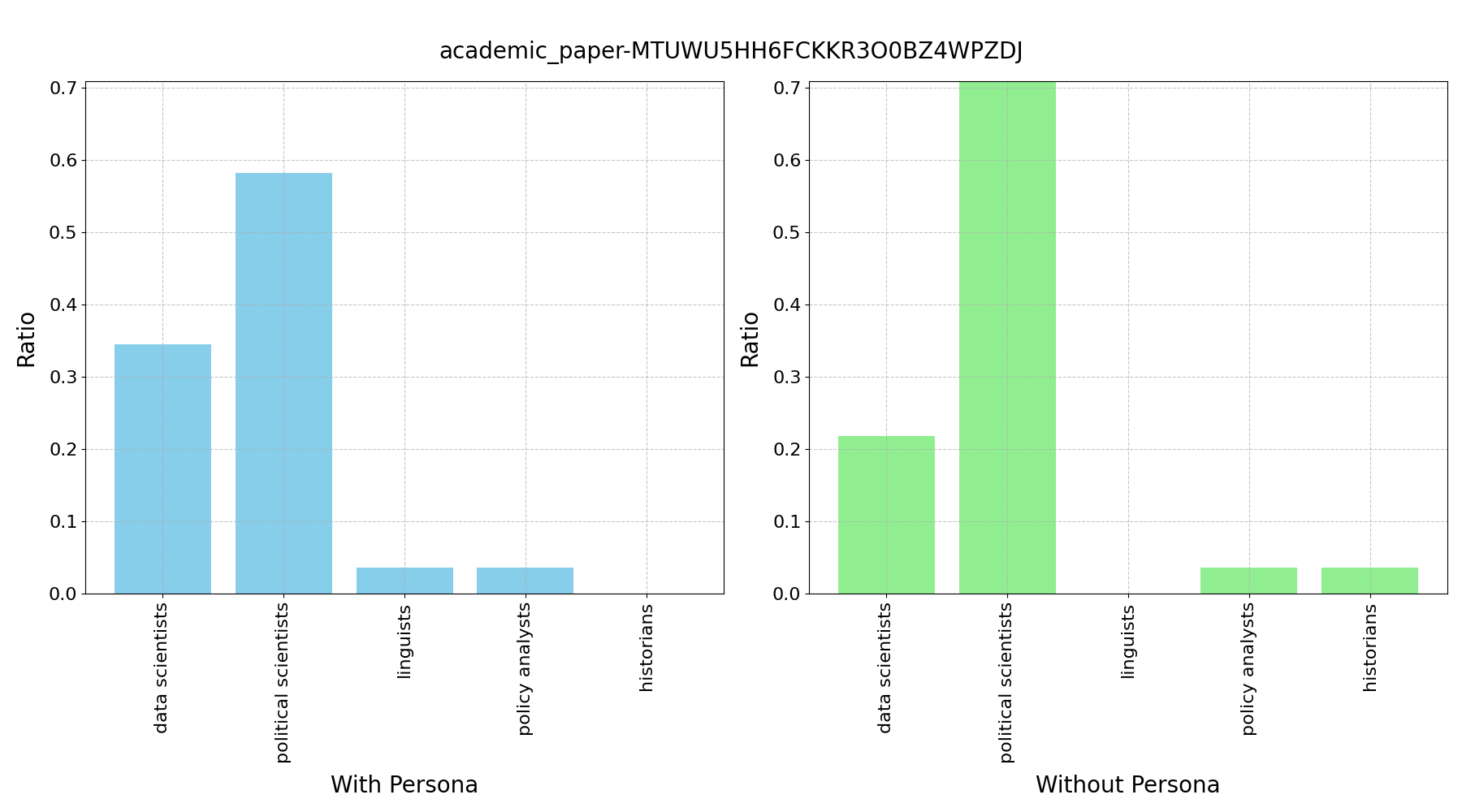}
    }

    \subfigure[Case 5)]{
        \includegraphics[width=0.8\textwidth]{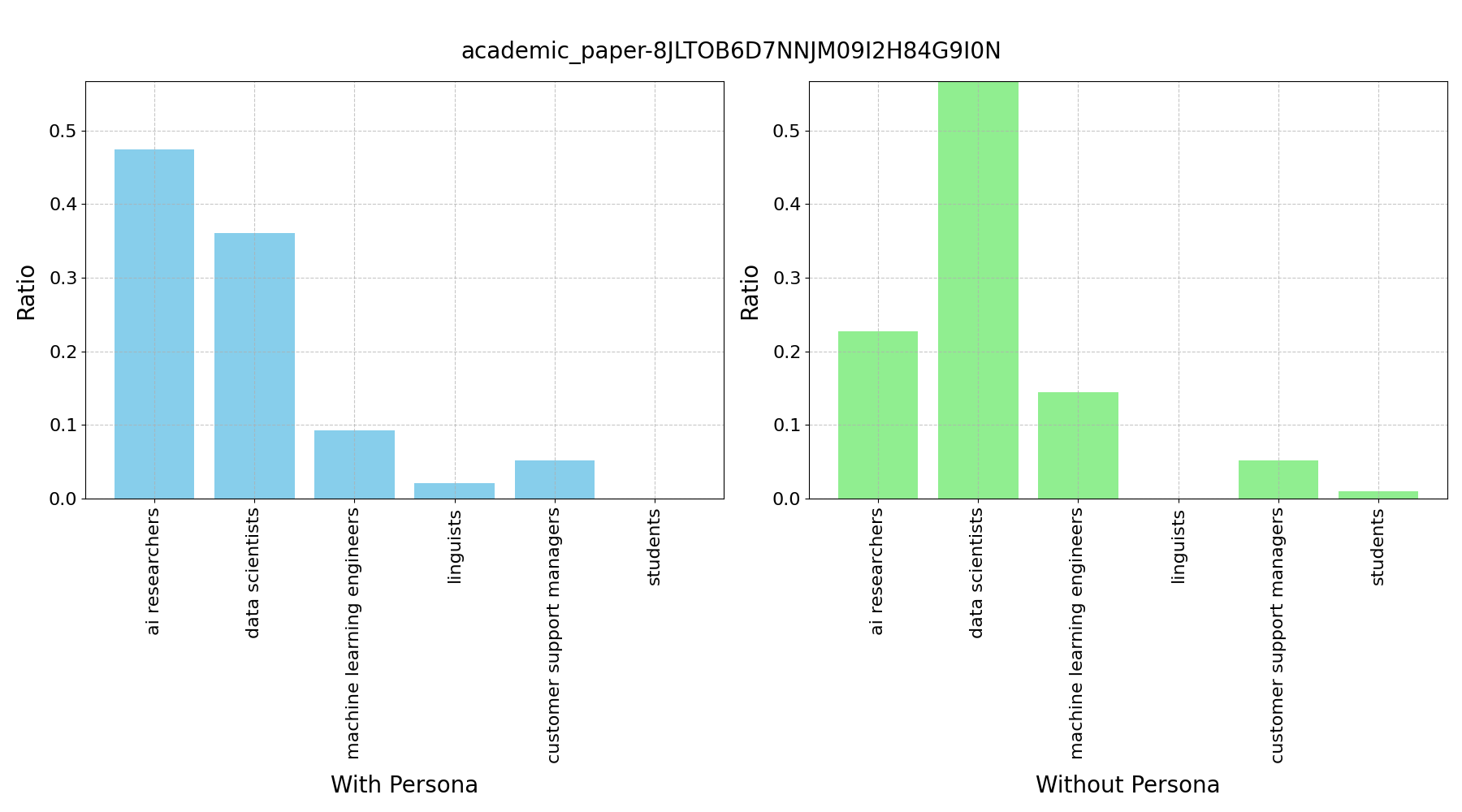}
    }
    
    \subfigure[Case 6]{
        \includegraphics[width=0.8\textwidth]{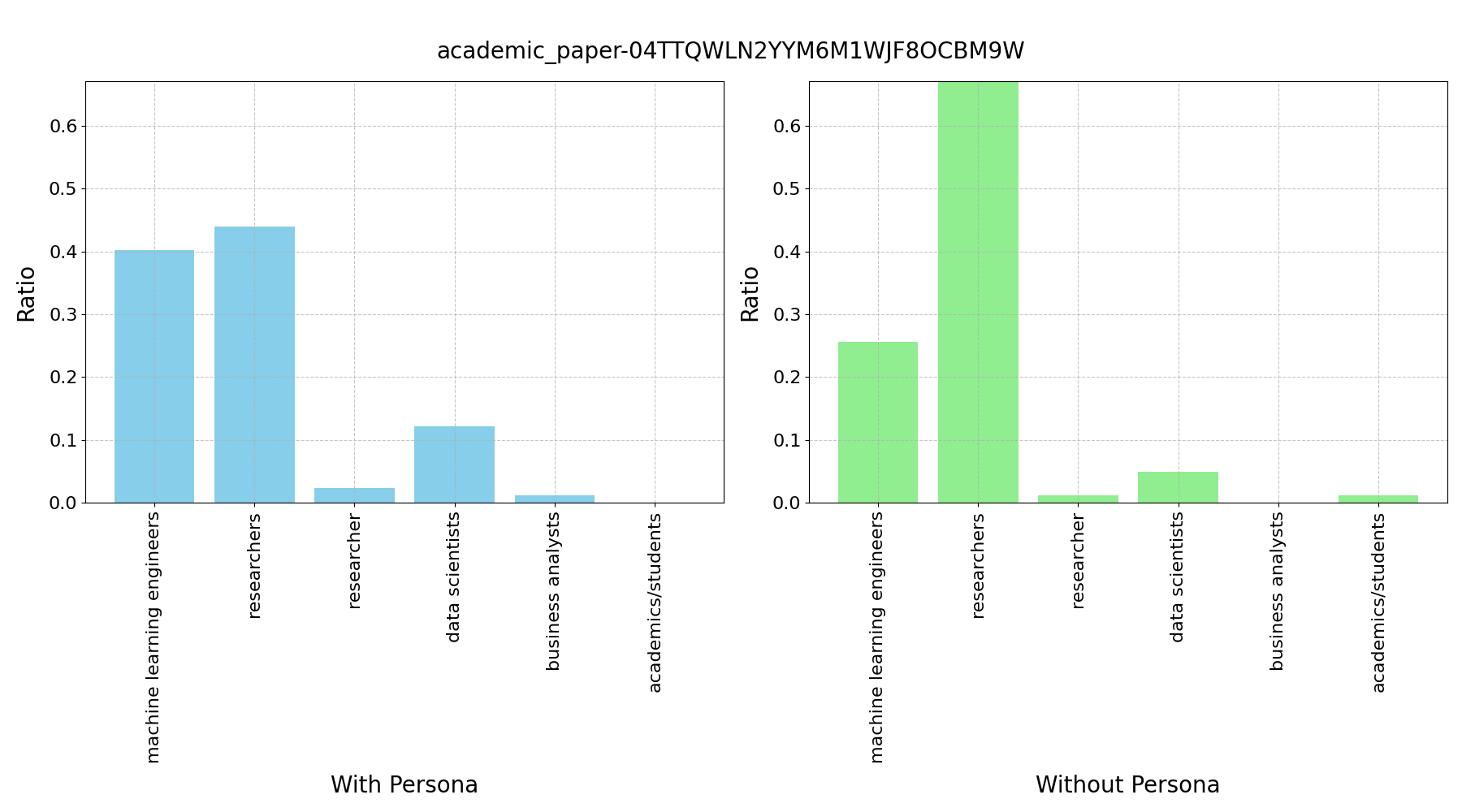}
    }
    
    \vspace{-5pt}
    \caption{Case 4-6: Document-level comparison of persona distribution between SQs generated with and without persona across three different cases in \textbf{academia} domain.}
    \label{fig:persona-distribution-academia-4-6}
\end{figure*}

We display more examples of persona distribution from the three domains of legal, finance, and academia in Figure 
\ref{fig:persona-distribution-legal-1-3},
\ref{fig:persona-distribution-legal-4-6}, Figure
\ref{fig:persona-distribution-finance-1-3},
\ref{fig:persona-distribution-finance-4-6}, Figure
\ref{fig:persona-distribution-academia-1-3},
\ref{fig:persona-distribution-academia-4-6} respectively.

\subsection{Question Persona Alignment}
\label{app:more-eval-results-coverage-ratio}
\begin{figure*}[!ht]
    \centering
    \subfigure[accountants]{
        \includegraphics[width=0.30\textwidth]{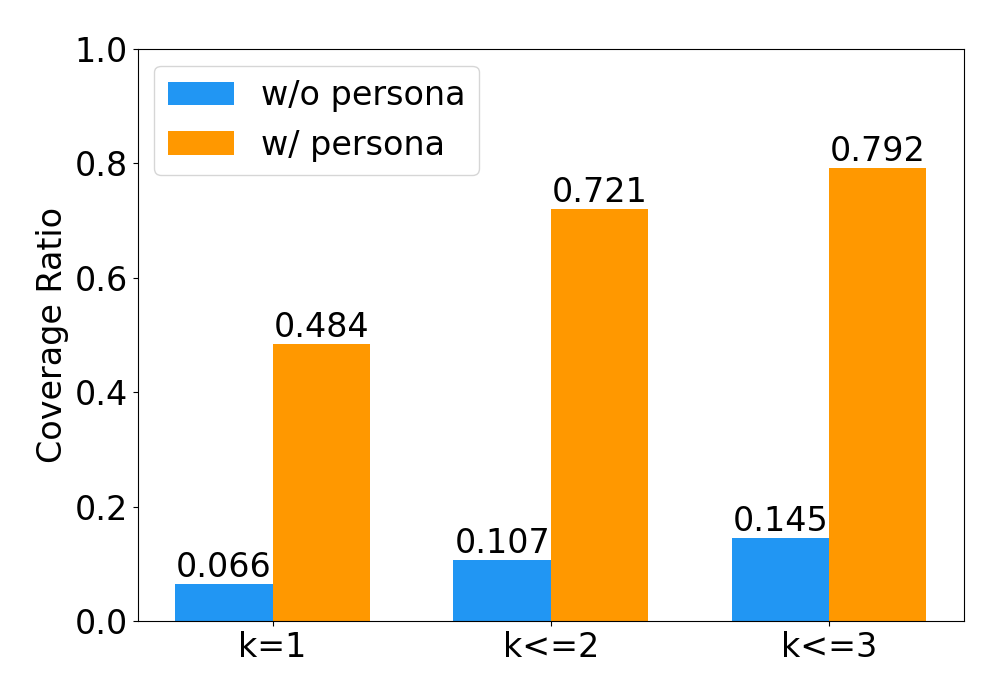}
    }
    \hfill  
    \subfigure[business owners]{
        \includegraphics[width=0.30\textwidth]{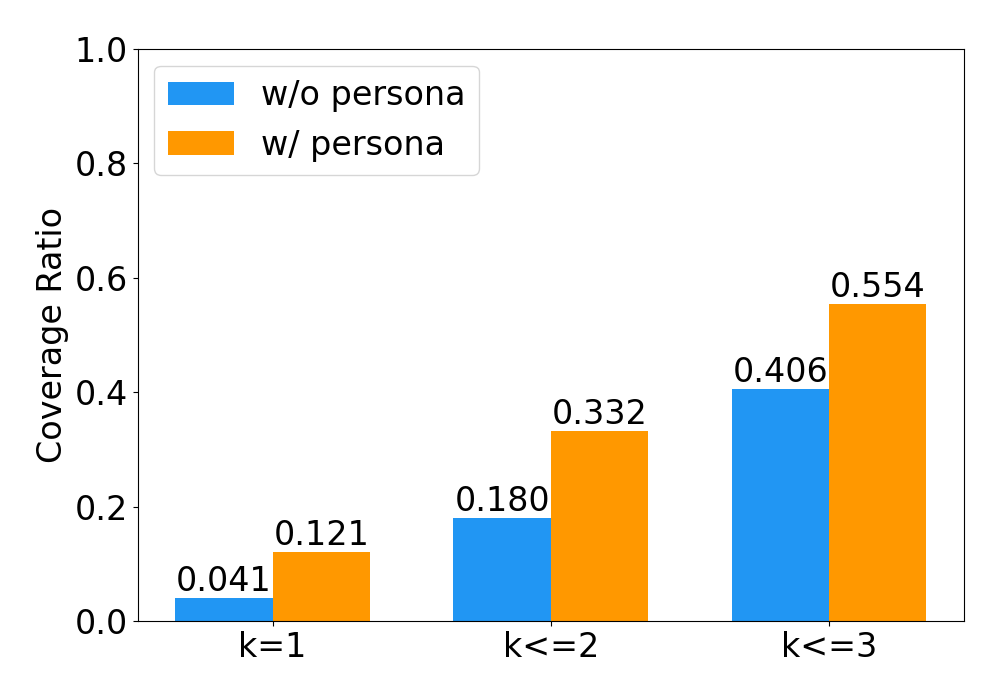}
    }
    \hfill
    \subfigure[consultants]{
        \includegraphics[width=0.30\textwidth]{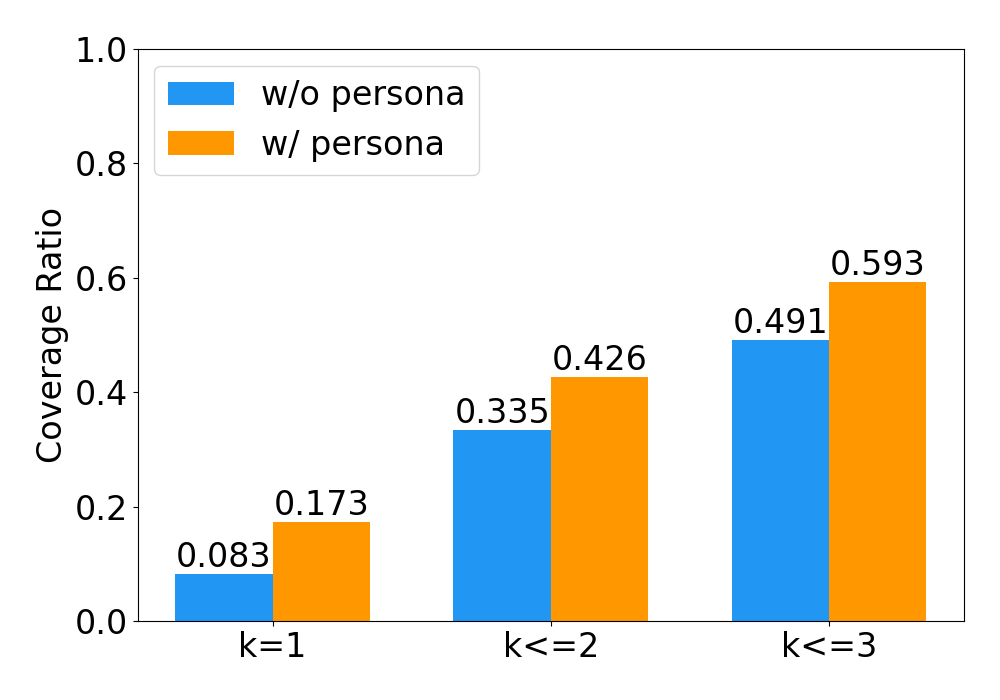}
    }

    \subfigure[data analysts]{
        \includegraphics[width=0.30\textwidth]{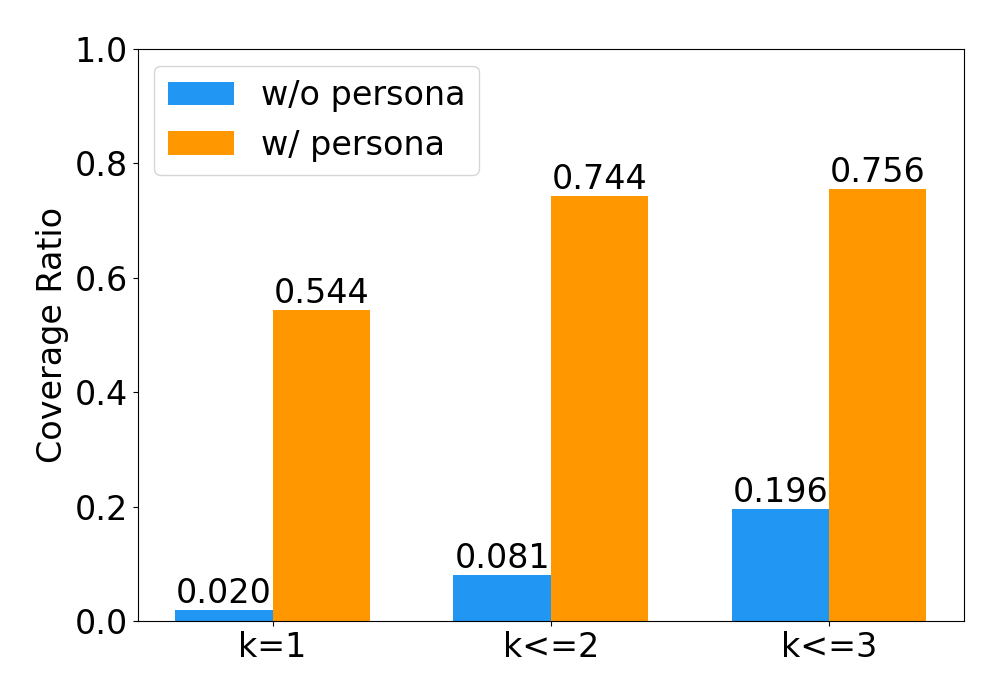}
    }
    \hfill  
    \subfigure[employees job candidates]{
        \includegraphics[width=0.30\textwidth]{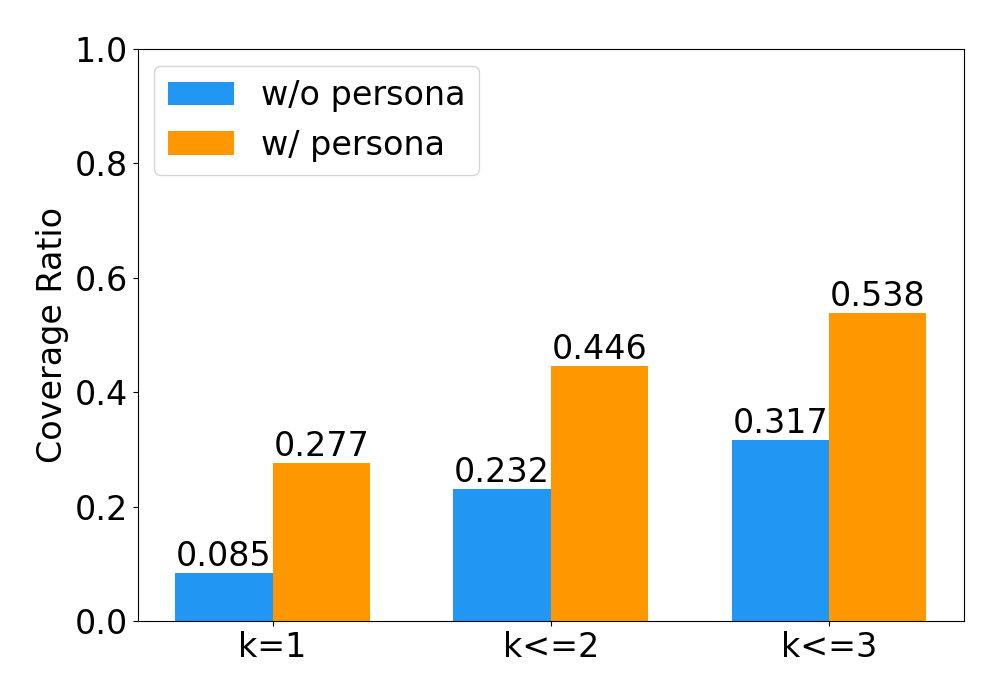}
    }
    \hfill
    \subfigure[environmental consultants]{
        \includegraphics[width=0.30\textwidth]{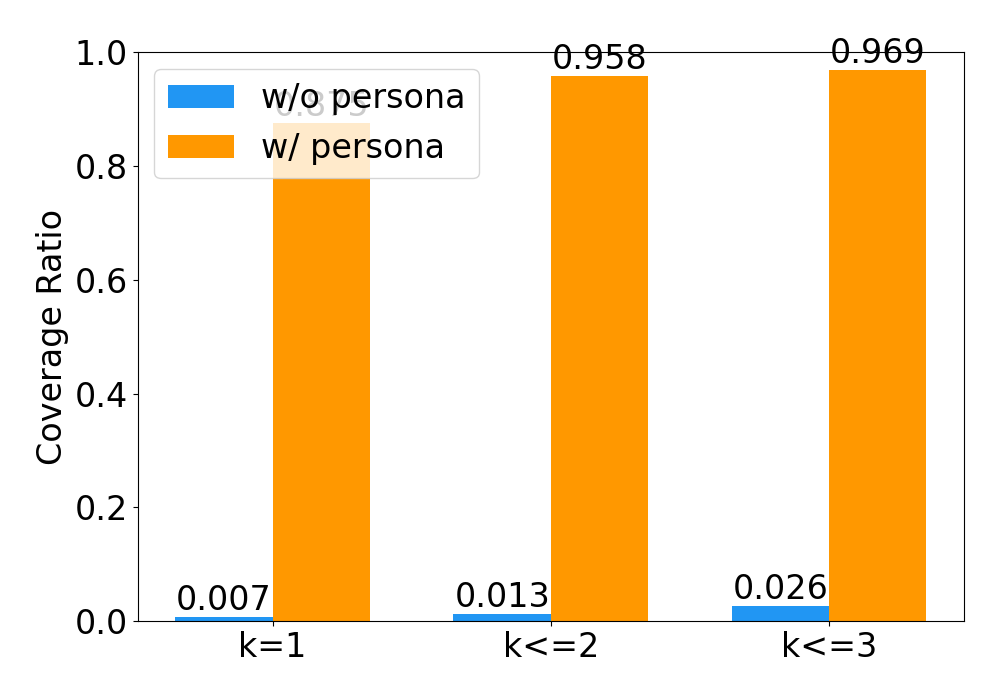}
    }

    \subfigure[event managers]{
        \includegraphics[width=0.30\textwidth]{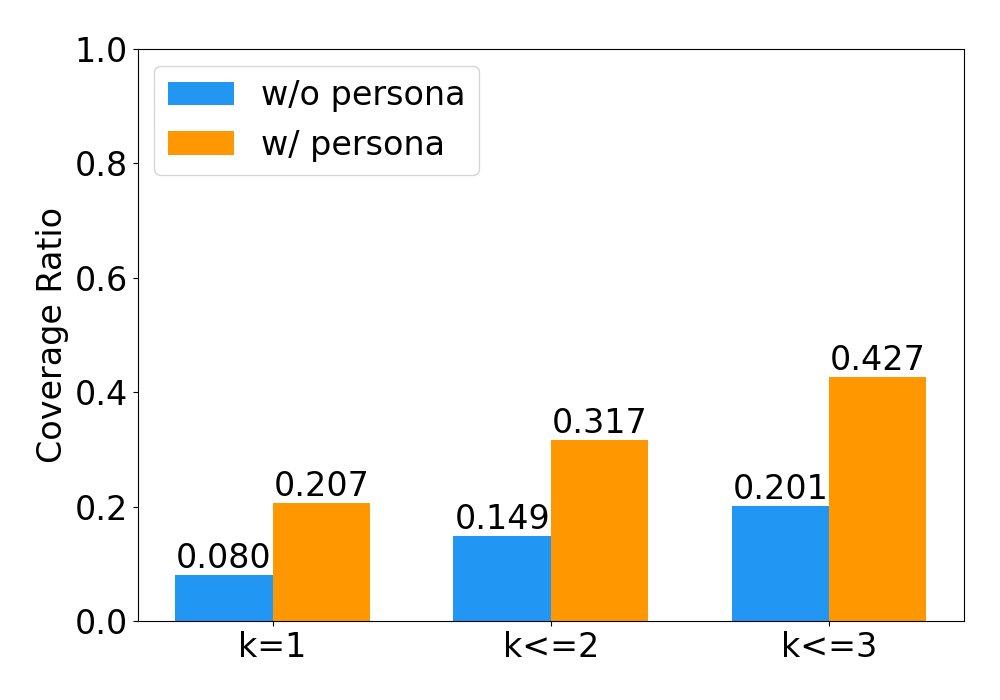}
    }
    \hfill  
    \subfigure[financial advisors]{
        \includegraphics[width=0.30\textwidth]{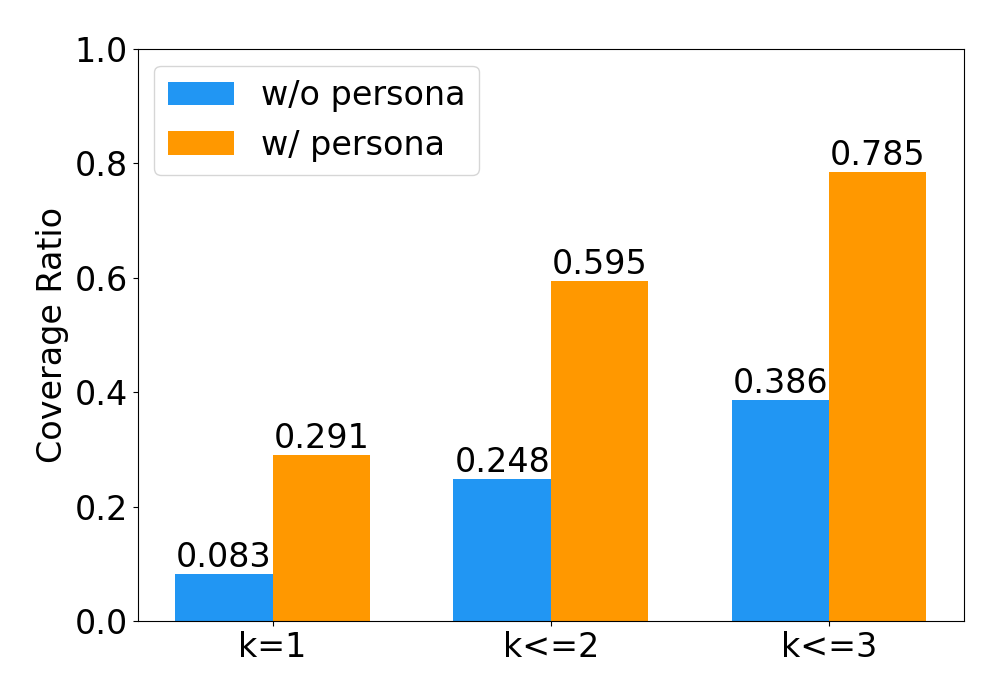}
    }
    \hfill
    \subfigure[health and safety officers]{
        \includegraphics[width=0.30\textwidth]{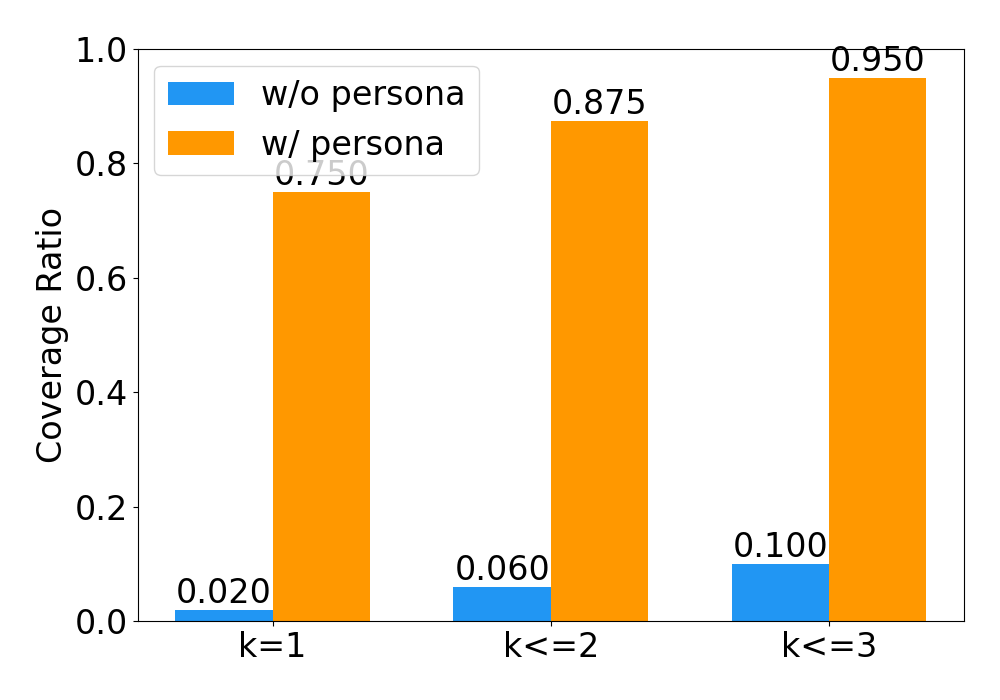}
    }
    \subfigure[insurance agents]{
        \includegraphics[width=0.30\textwidth]{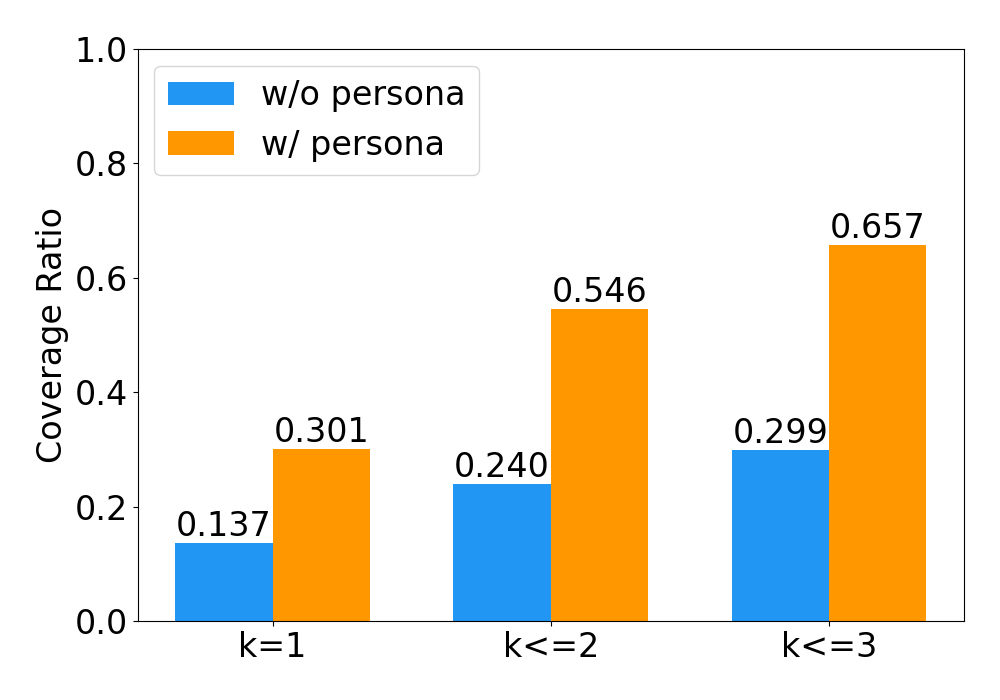}
    }
    \hfill  
    \subfigure[marketers]{
        \includegraphics[width=0.30\textwidth]{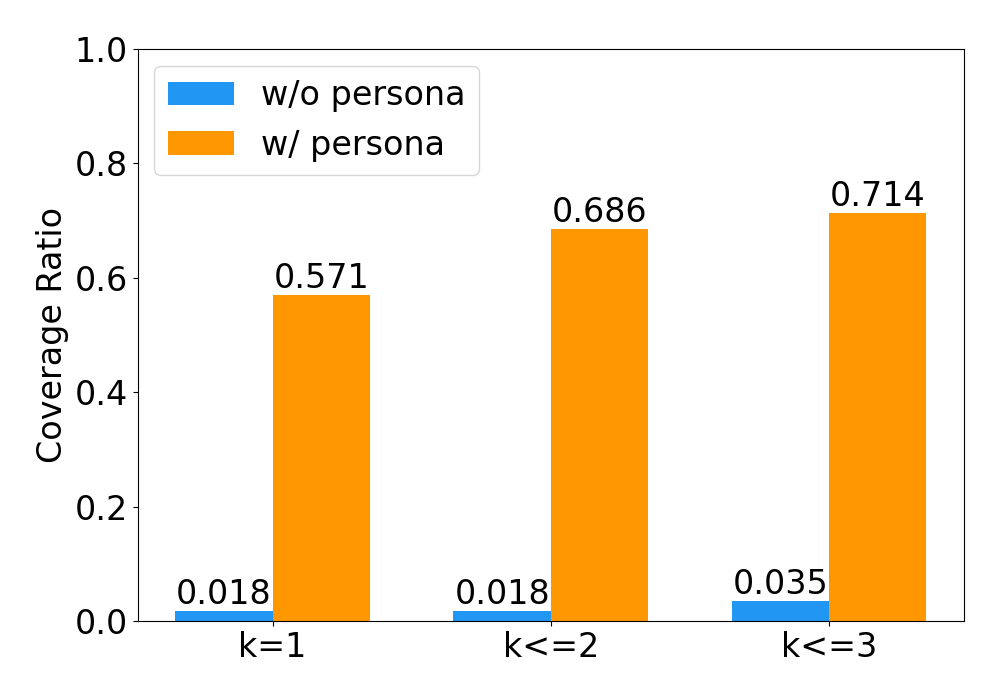}
    }
    \hfill
    \subfigure[plan administrators]{
        \includegraphics[width=0.30\textwidth]{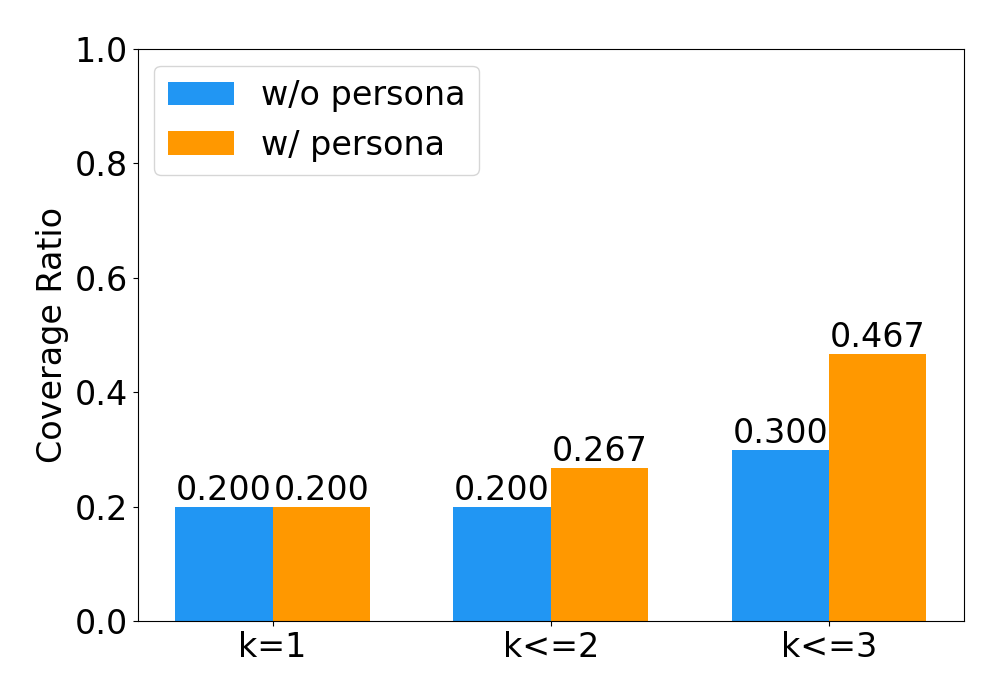}
    }
    \subfigure[procurement officers]{
        \includegraphics[width=0.30\textwidth]{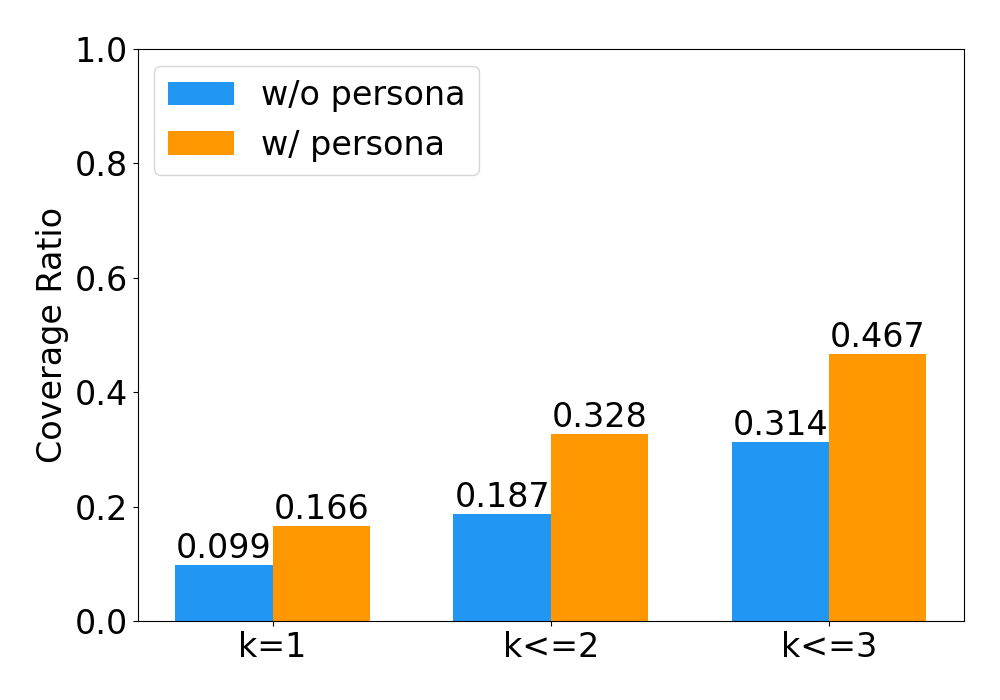}
    }
    \hfill  
    \subfigure[project managers]{
        \includegraphics[width=0.30\textwidth]{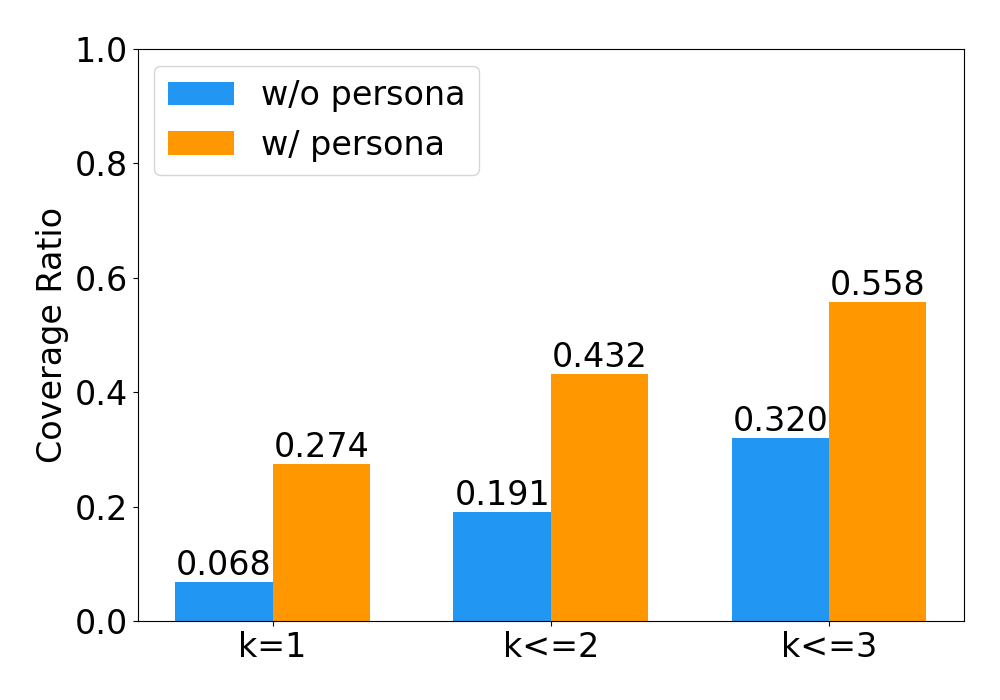}
    }
    \hfill
    \subfigure[risk managers]{
        \includegraphics[width=0.30\textwidth]{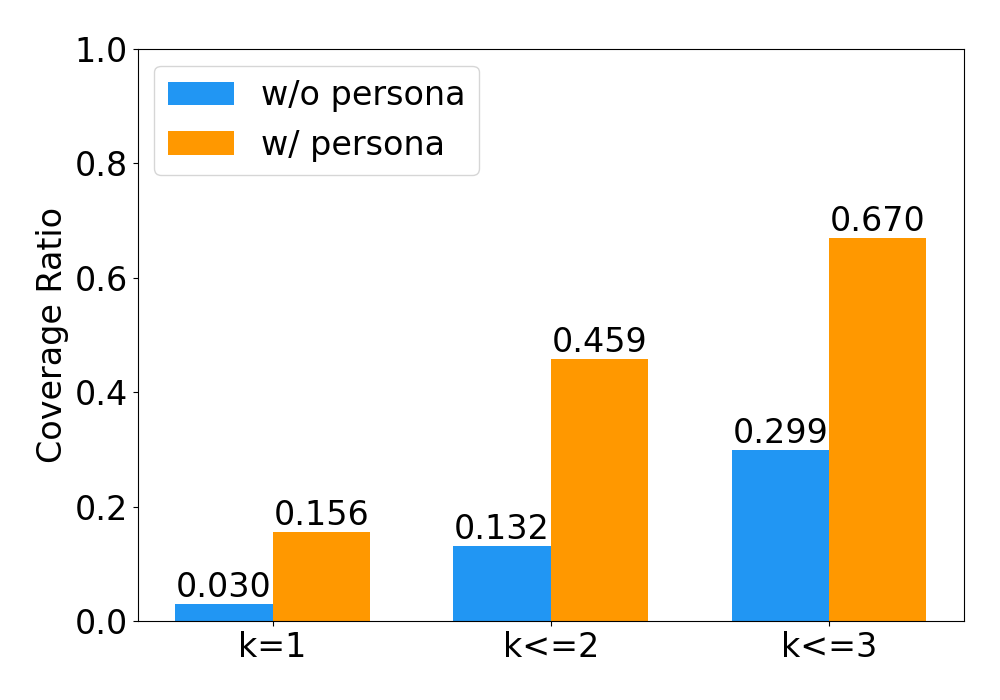}
    }
    \vspace{-5pt}
    \caption{The coverage ratio of 15 examples personas in the \textbf{legal} domain.}
    \label{fig-app:cover-ratio-legal}
\end{figure*}

\begin{figure*}[!ht]
    \centering
    \subfigure[accountants]{
        \includegraphics[width=0.30\textwidth]{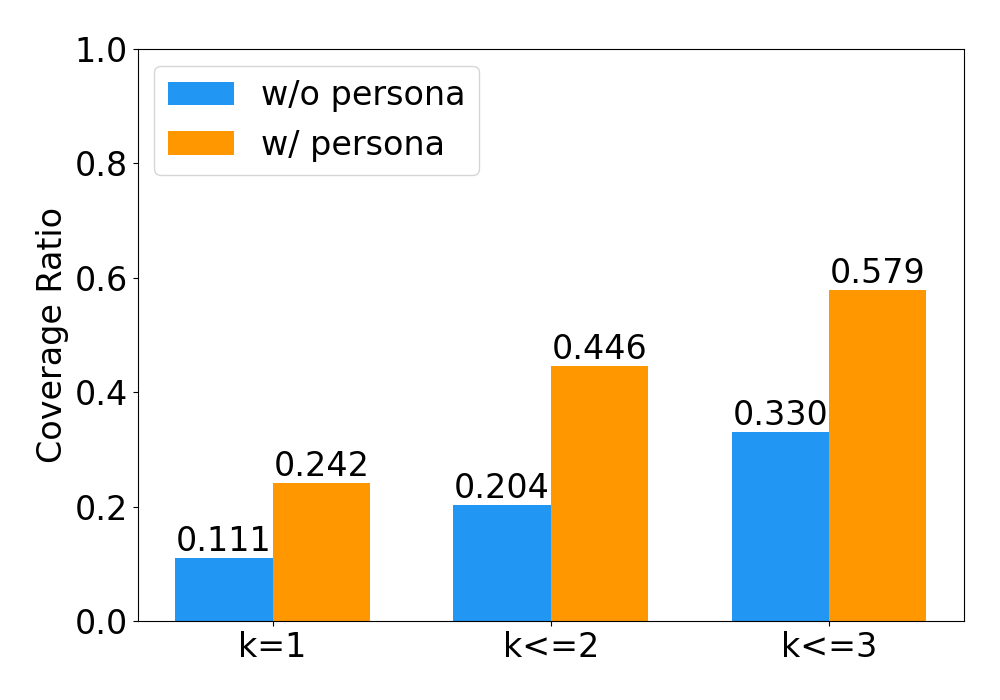}
    }
    \hfill  
    \subfigure[auditors]{
        \includegraphics[width=0.30\textwidth]{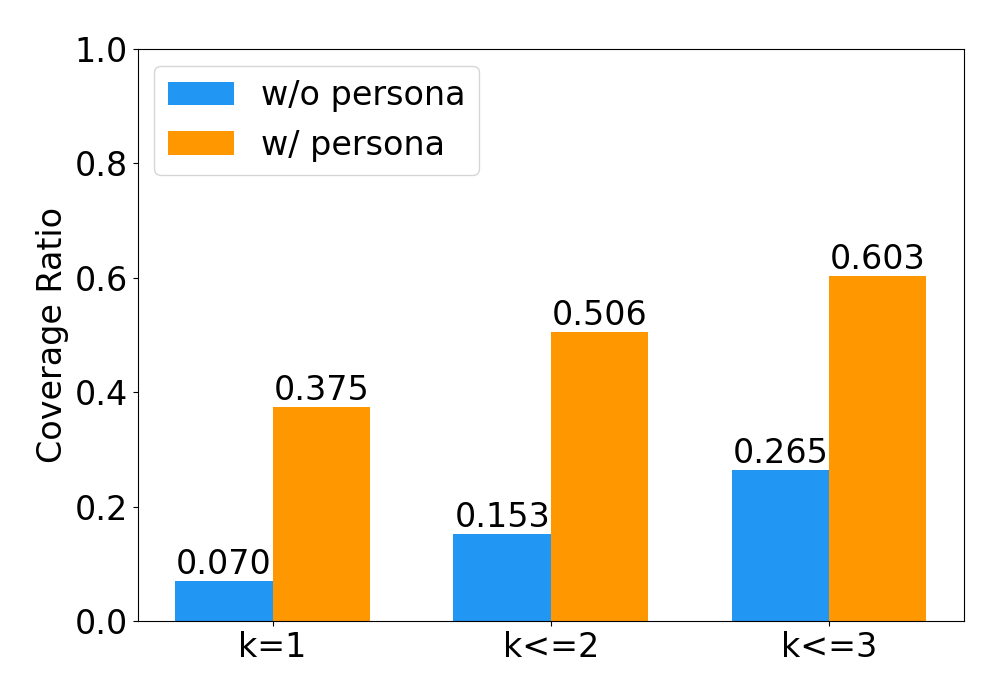}
    }
    \hfill
    \subfigure[board members]{
        \includegraphics[width=0.30\textwidth]{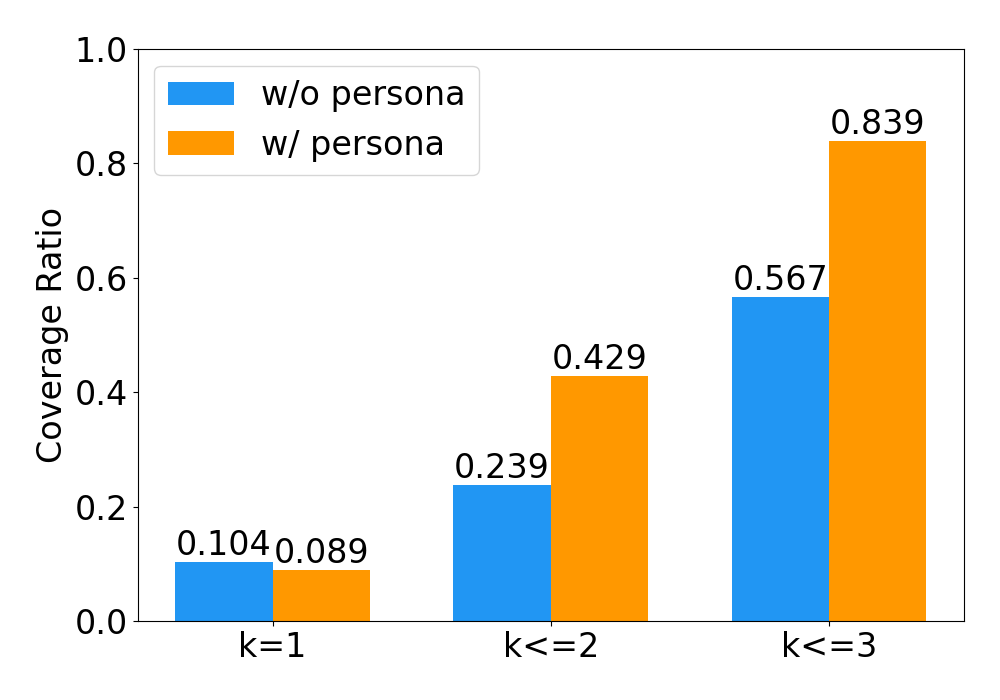}
    }

    \subfigure[business strategists]{
        \includegraphics[width=0.30\textwidth]{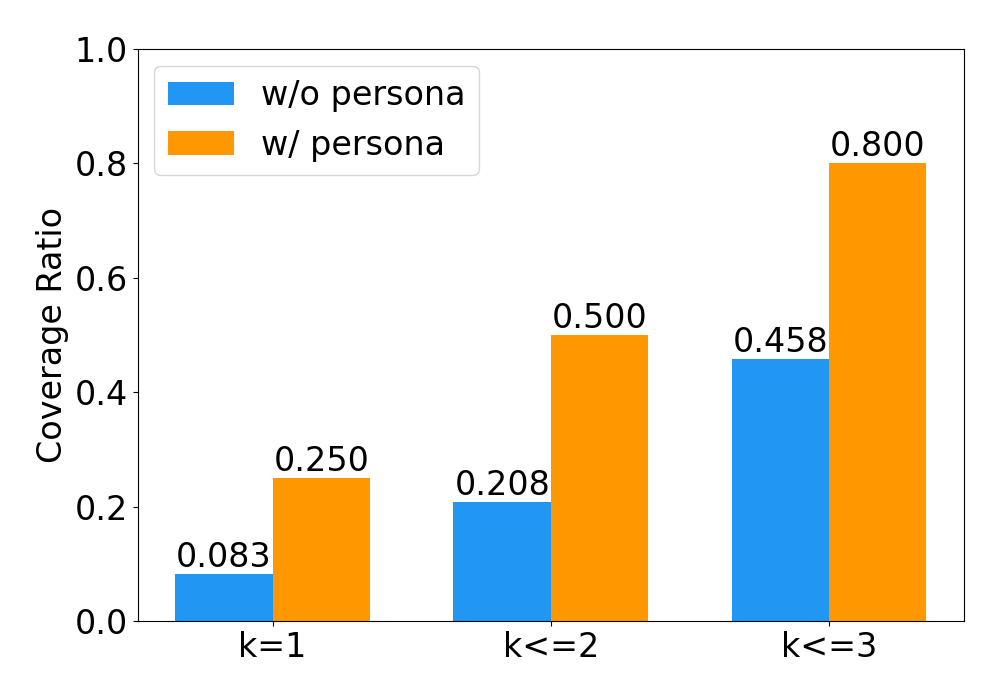}
    }
    \hfill  
    \subfigure[community stakeholders]{
        \includegraphics[width=0.30\textwidth]{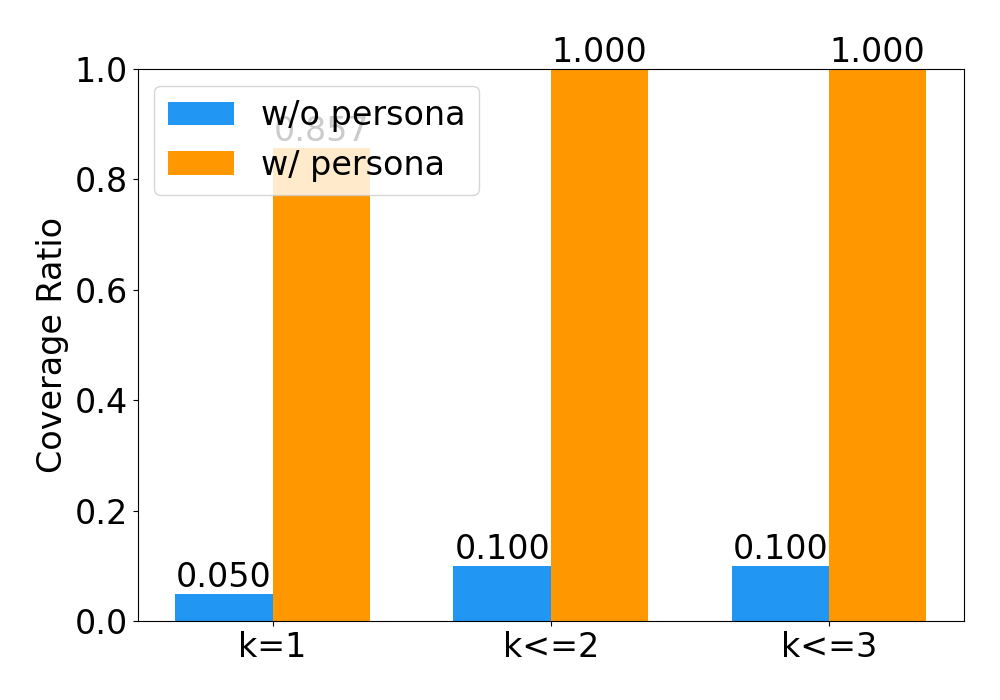}
    }
    \hfill
    \subfigure[company management]{
        \includegraphics[width=0.30\textwidth]{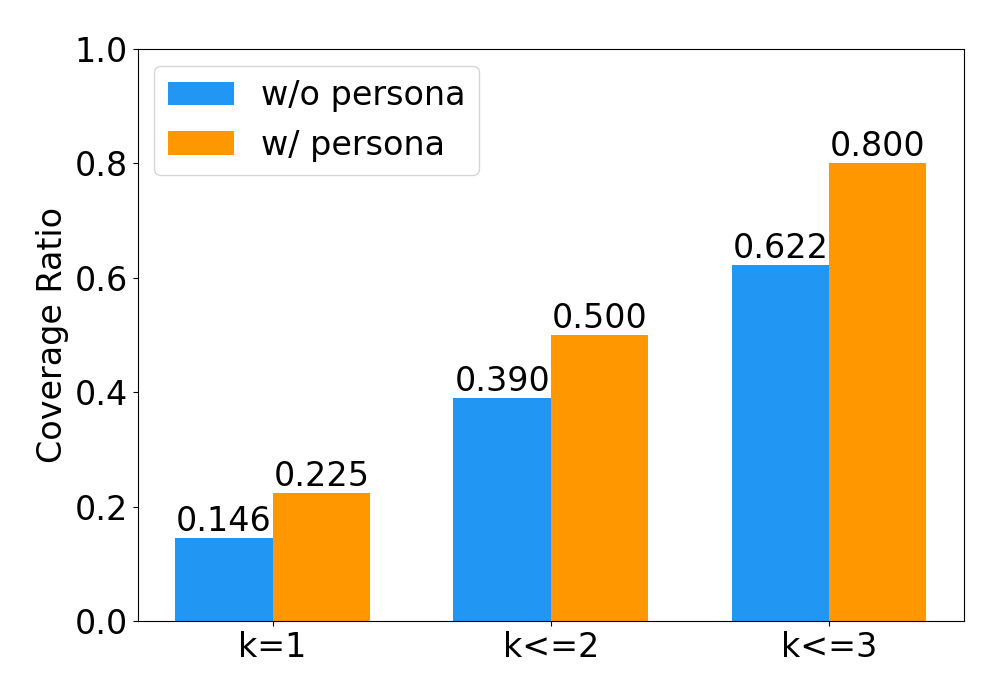}
    }

    \subfigure[executives]{
        \includegraphics[width=0.30\textwidth]{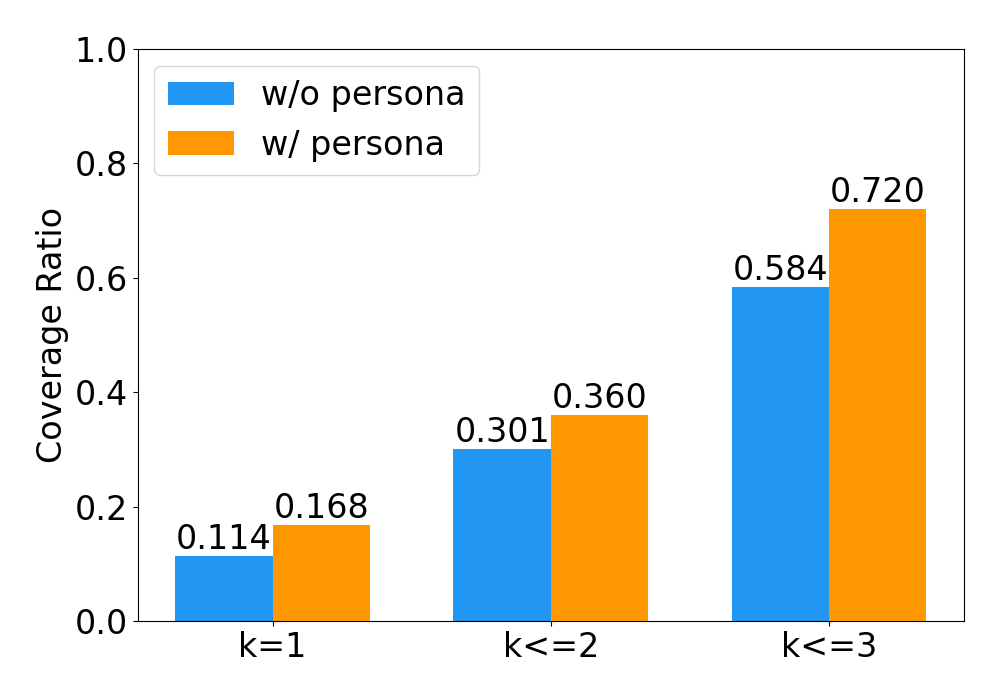}
    }
    \hfill  
    \subfigure[financial analysts]{
        \includegraphics[width=0.30\textwidth]{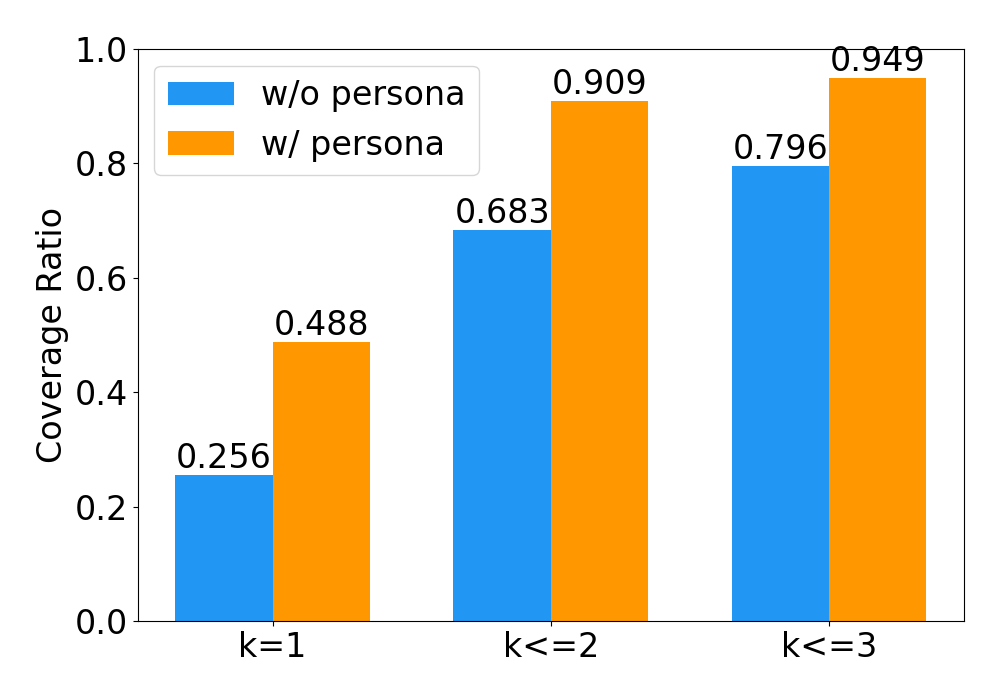}
    }
    \hfill
    \subfigure[hr professionalss]{
        \includegraphics[width=0.30\textwidth]{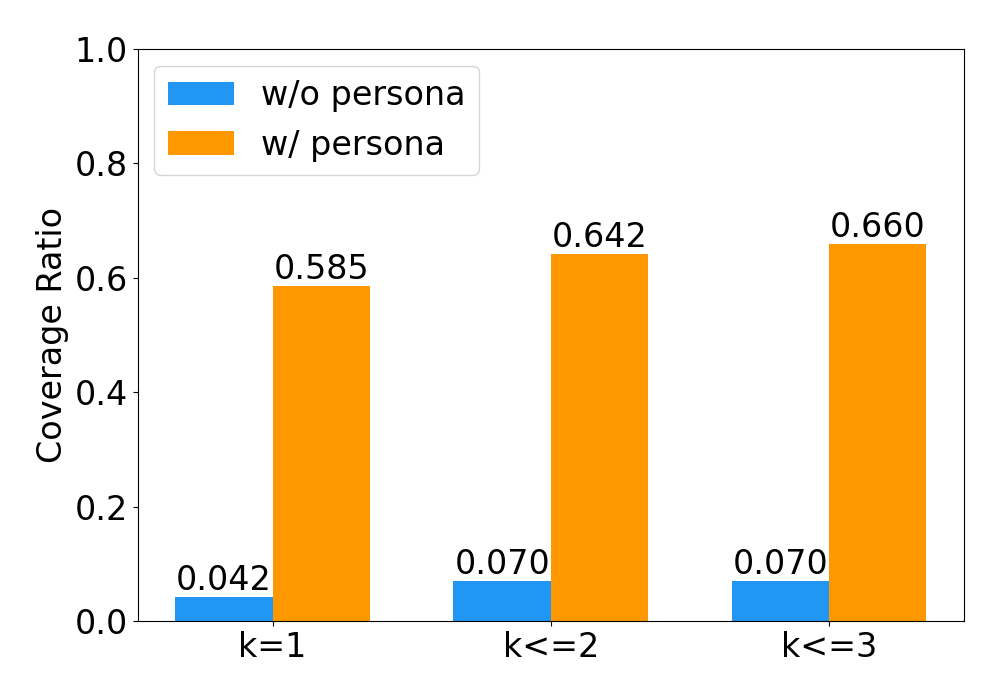}
    }
    \subfigure[investors]{
        \includegraphics[width=0.30\textwidth]{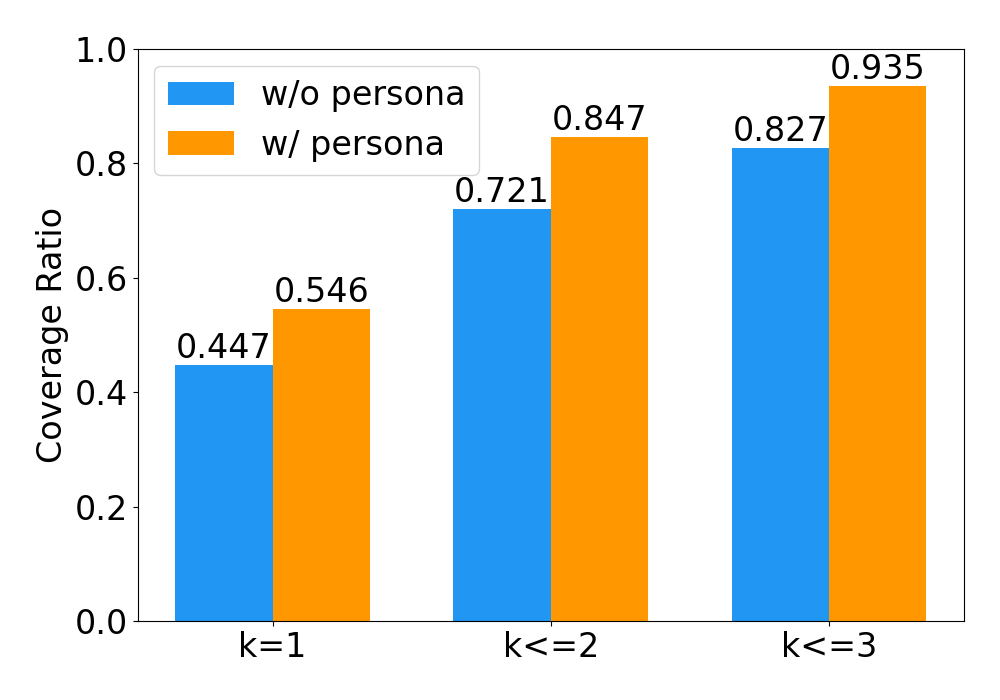}
    }
    \hfill  
    \subfigure[lawyer]{
        \includegraphics[width=0.30\textwidth]{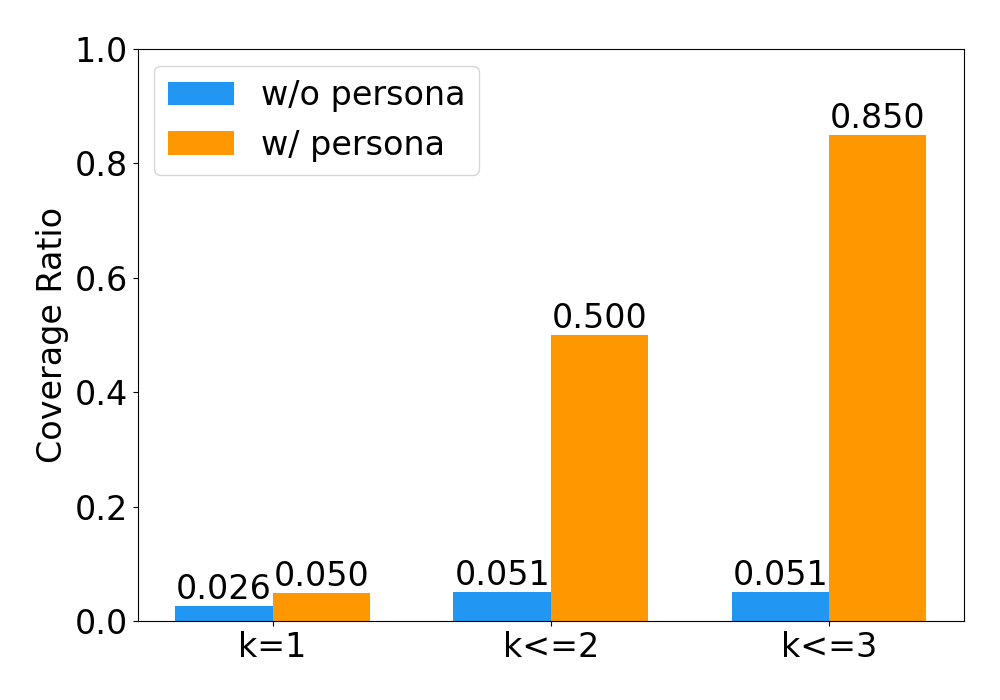}
    }
    \hfill
    \subfigure[market analysts]{
        \includegraphics[width=0.30\textwidth]{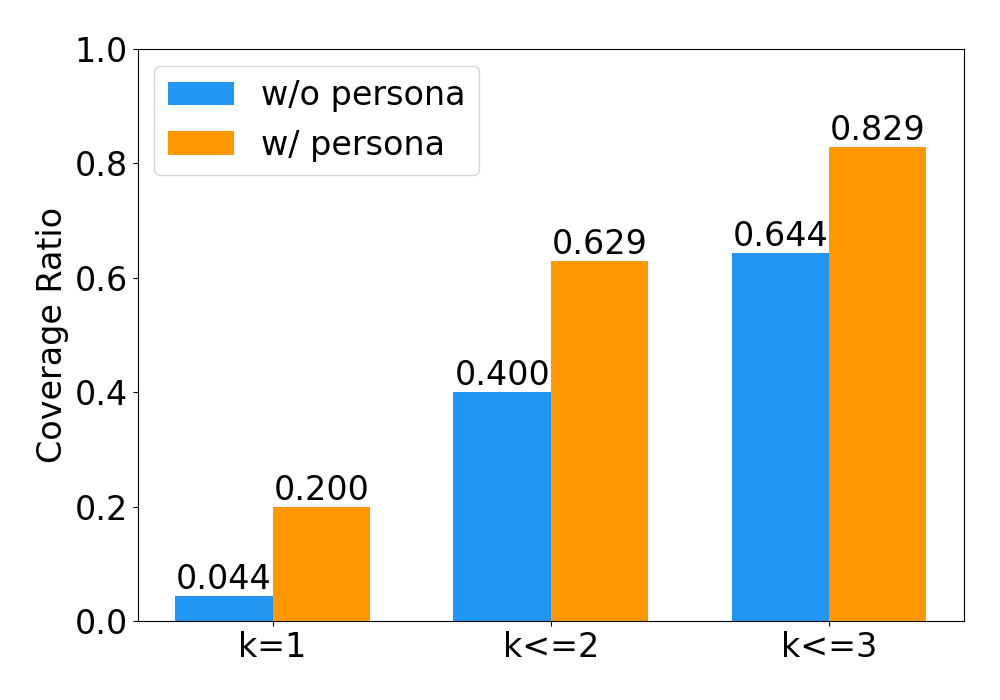}
    }
    \subfigure[marketing professionals]{
        \includegraphics[width=0.30\textwidth]{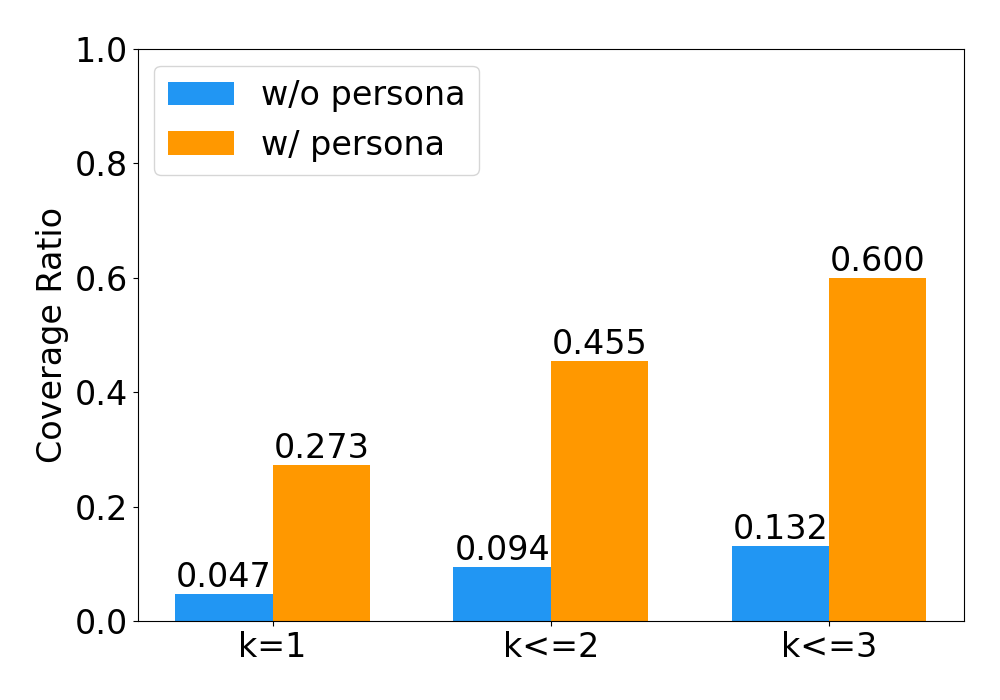}
    }
    \hfill  
    \subfigure[shareholders]{
        \includegraphics[width=0.30\textwidth]{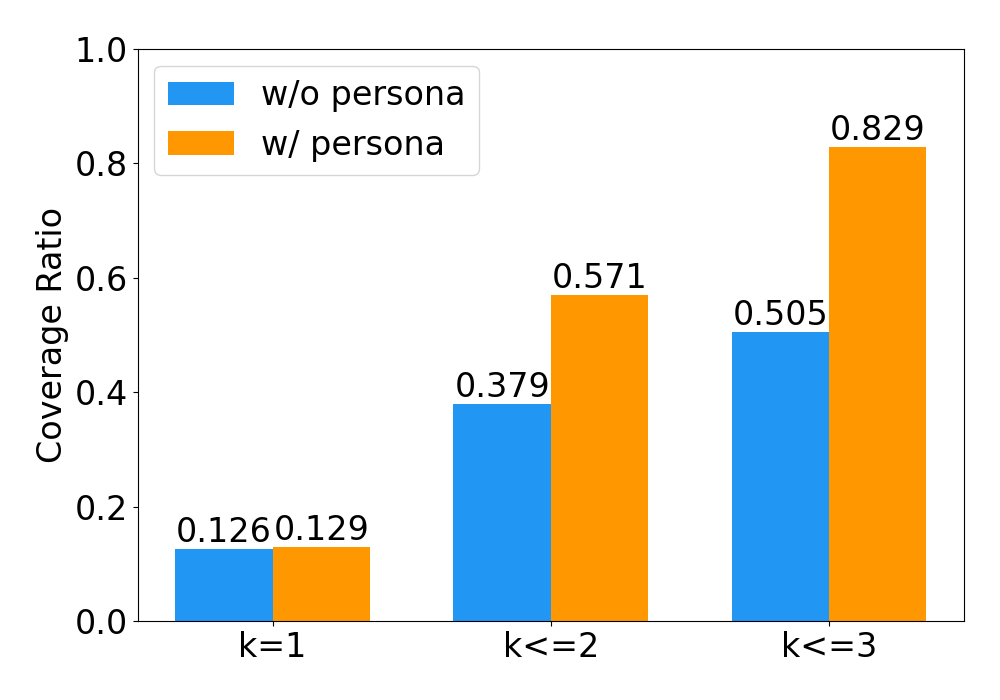}
    }
    \hfill
    \subfigure[tax authorities]{
        \includegraphics[width=0.30\textwidth]{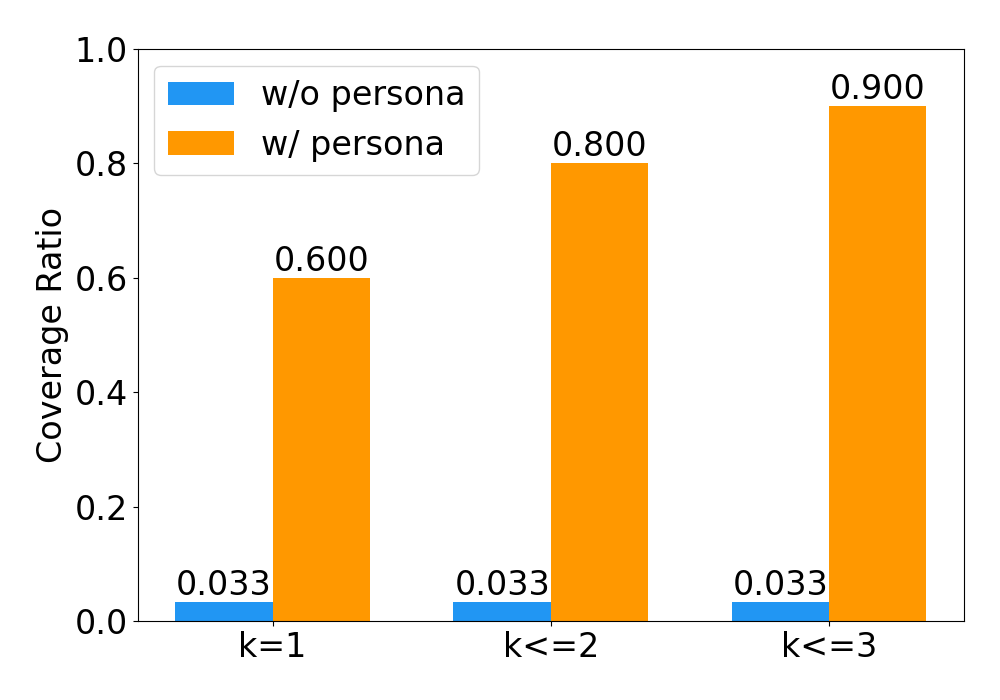}
    }
    \vspace{-5pt}
    \caption{The coverage ratio of 15 examples personas in the \textbf{finance} domain.}
    \label{fig-app:cover-ratio-finance}
\end{figure*}

\begin{figure*}[!ht]
    \centering
    \subfigure[academic professors]{
        \includegraphics[width=0.30\textwidth]{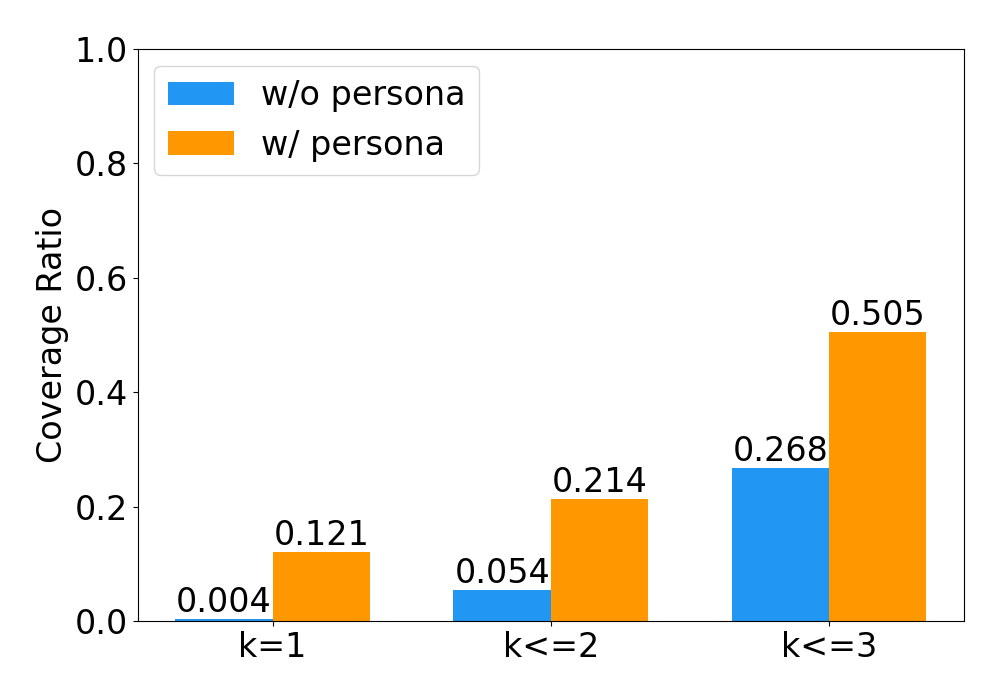}
    }
    \hfill
    \subfigure[ai researchers]{
        \includegraphics[width=0.30\textwidth]{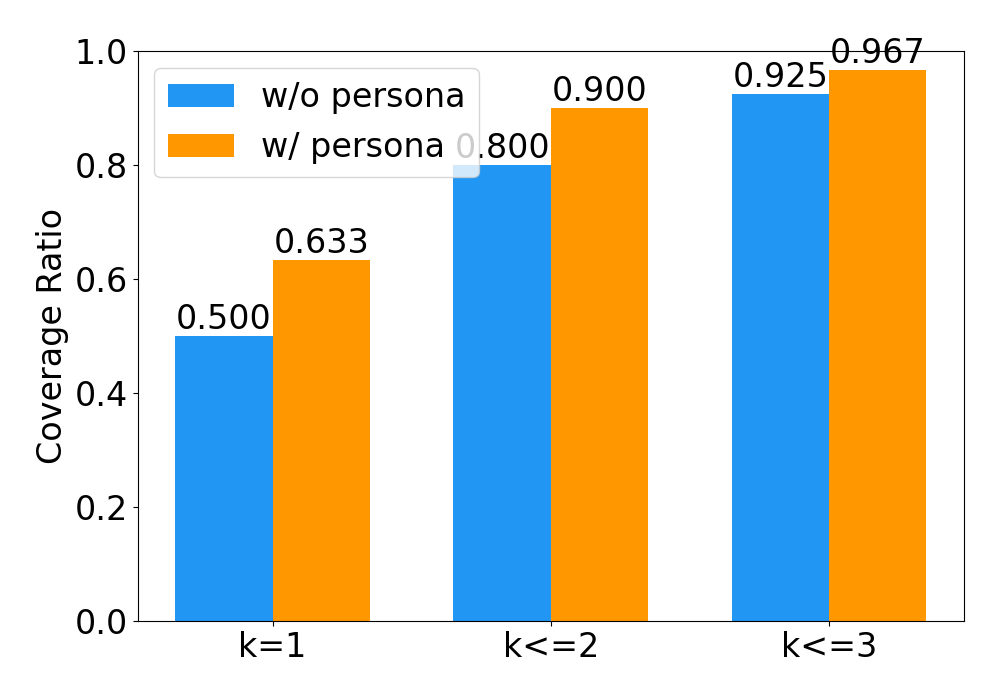}
    }
    \hfill
    \subfigure[computer vision experts]{
        \includegraphics[width=0.30\textwidth]{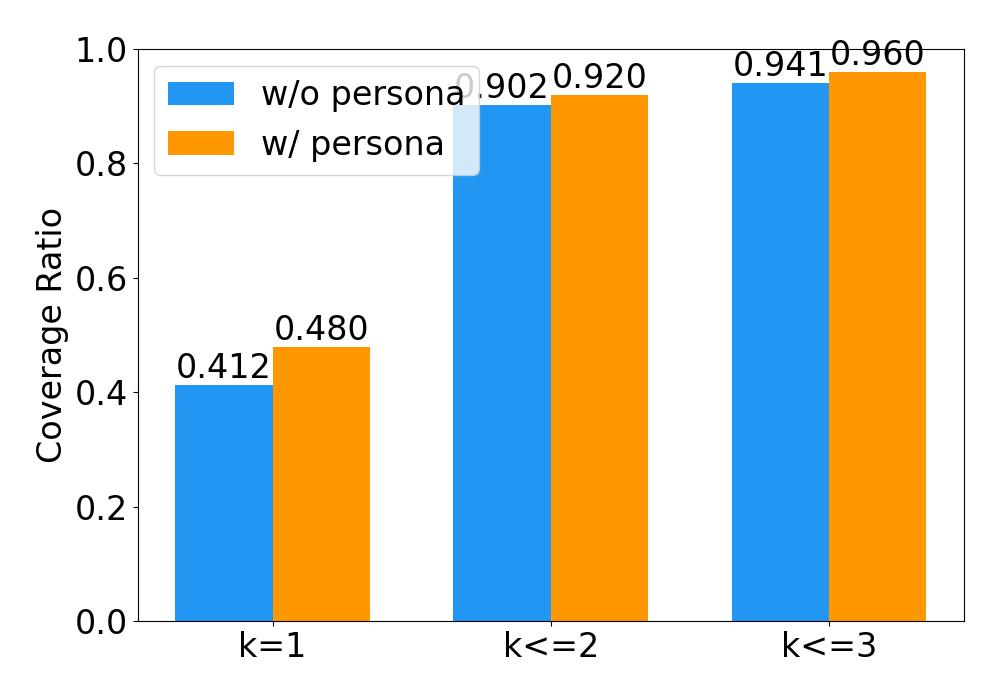}
    }
    \subfigure[educators]{
        \includegraphics[width=0.30\textwidth]{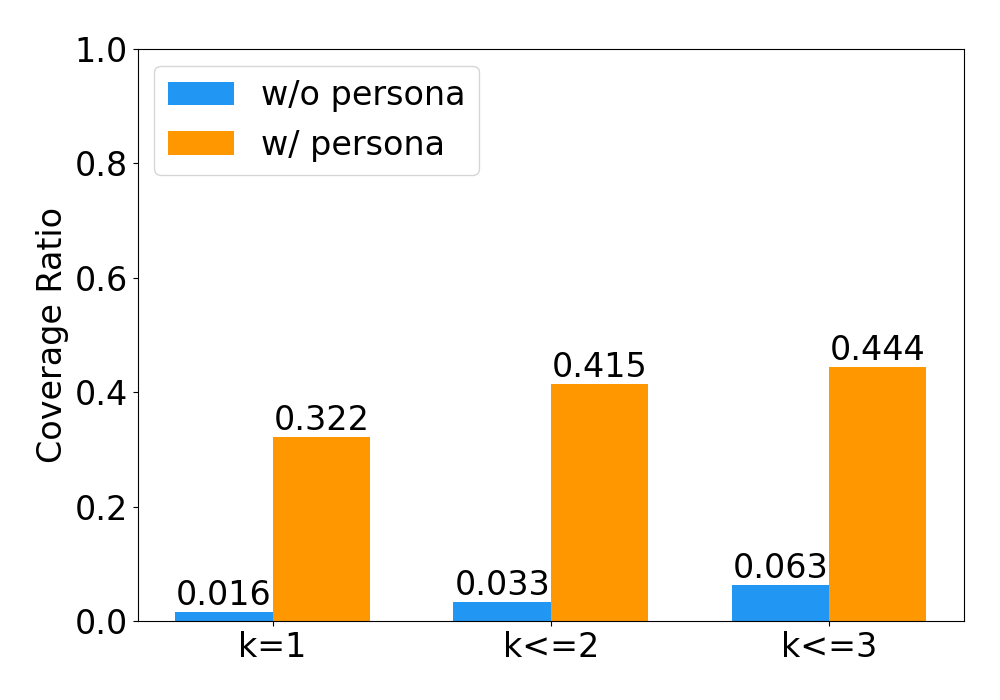}
    }
    \hfill
    \subfigure[general public]{
        \includegraphics[width=0.30\textwidth]{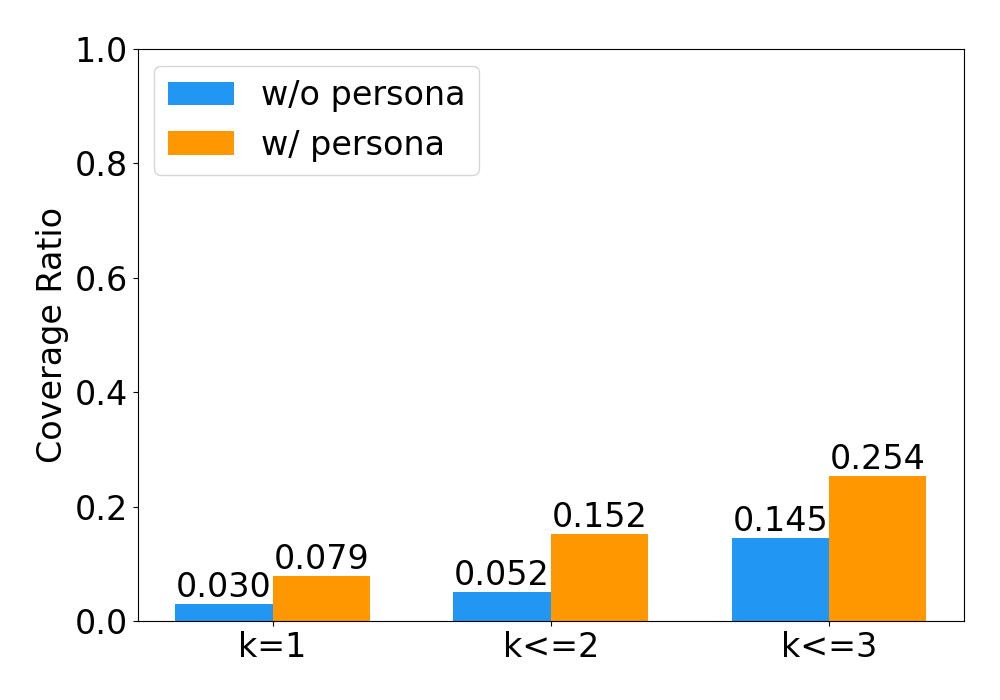}
    }
    \hfill
    \subfigure[linguists]{
        \includegraphics[width=0.30\textwidth]{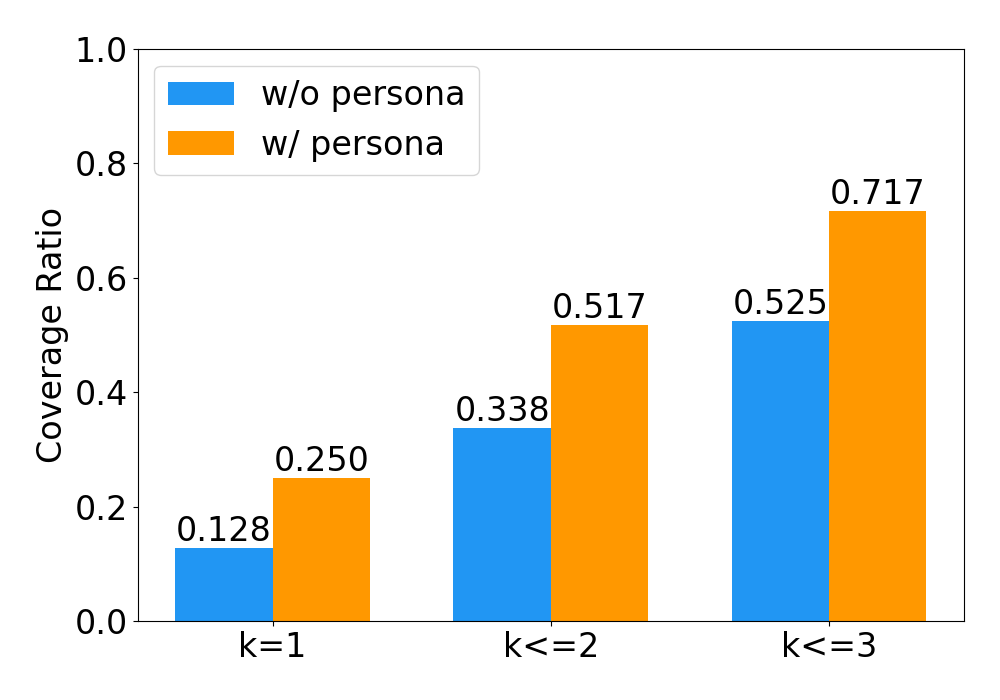}
    }
    \subfigure[machine learning engineers]{
        \includegraphics[width=0.30\textwidth]{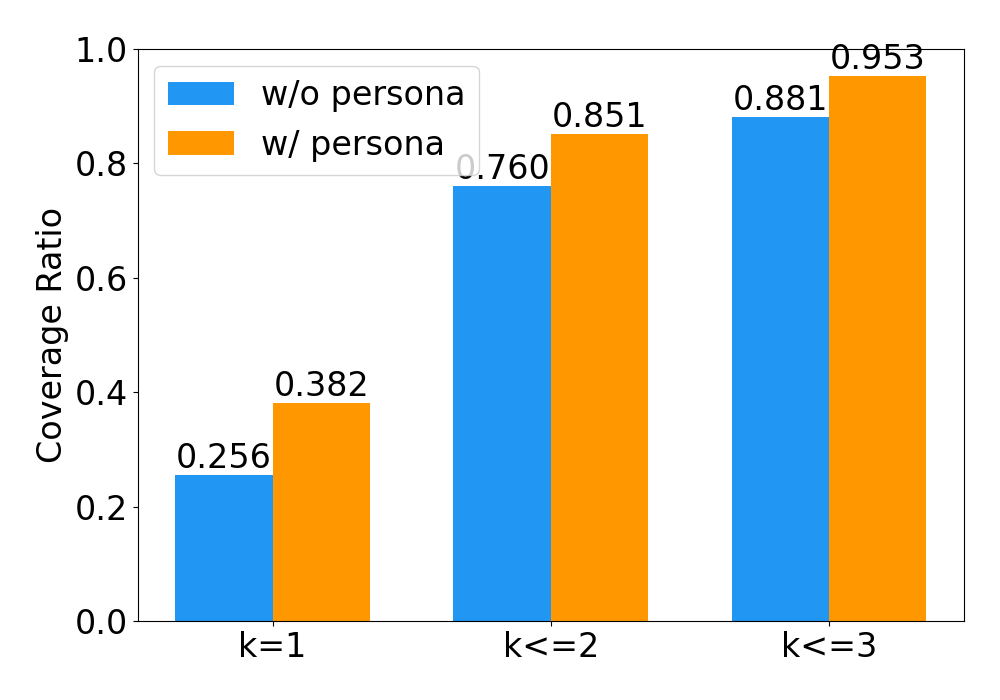}
    }
    \hfill
    \subfigure[privacy advocates]{
        \includegraphics[width=0.30\textwidth]{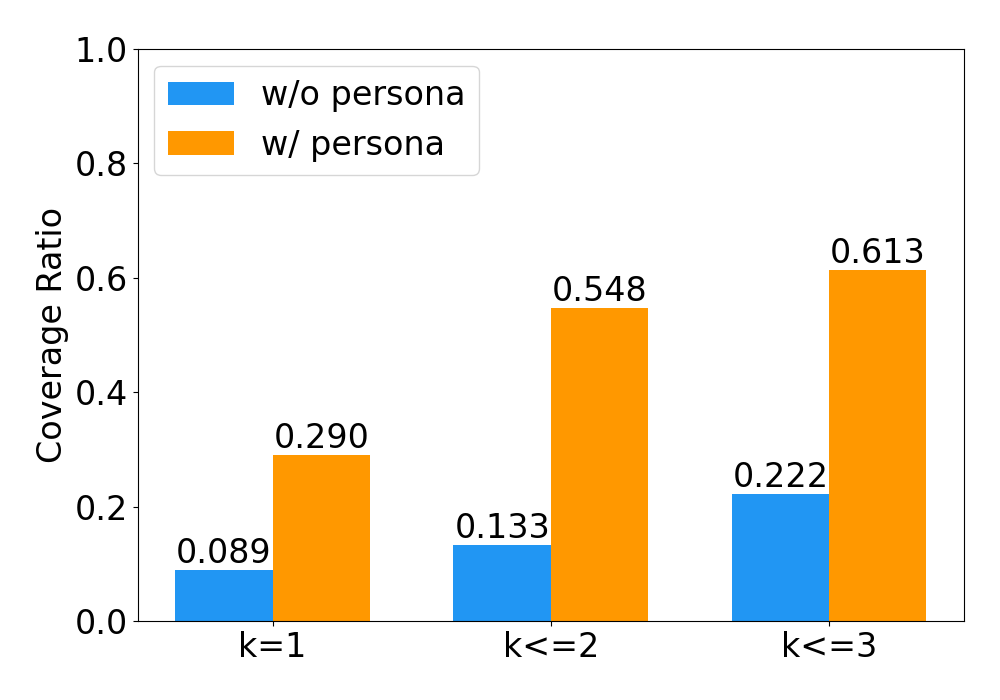}
    }
    \hfill
    \subfigure[product managers]{
        \includegraphics[width=0.30\textwidth]{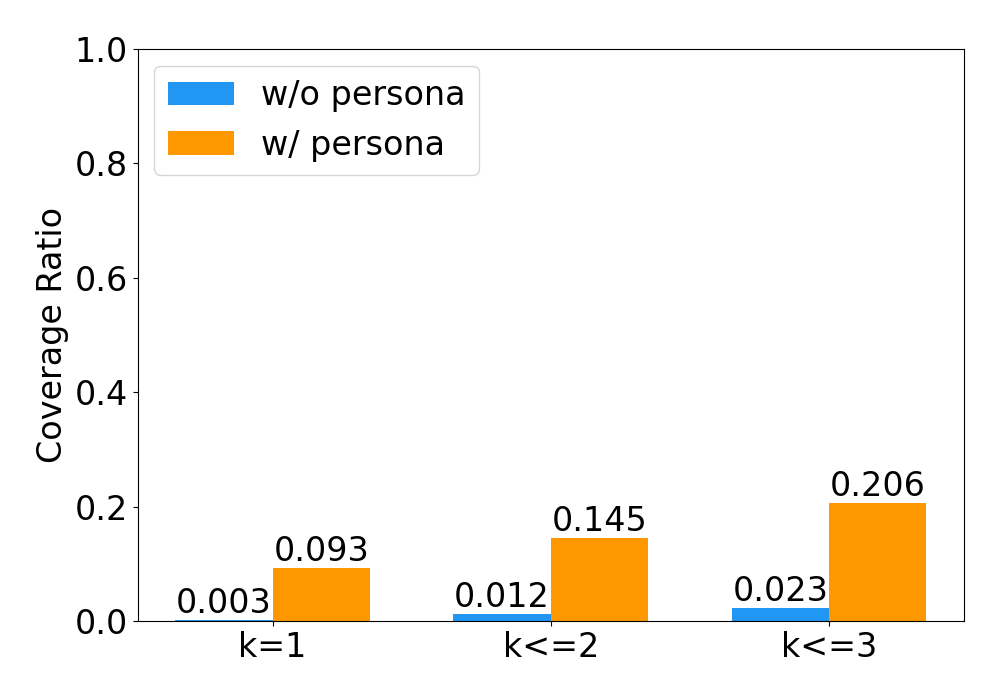}
    }
    \subfigure[researchers]{
        \includegraphics[width=0.30\textwidth]{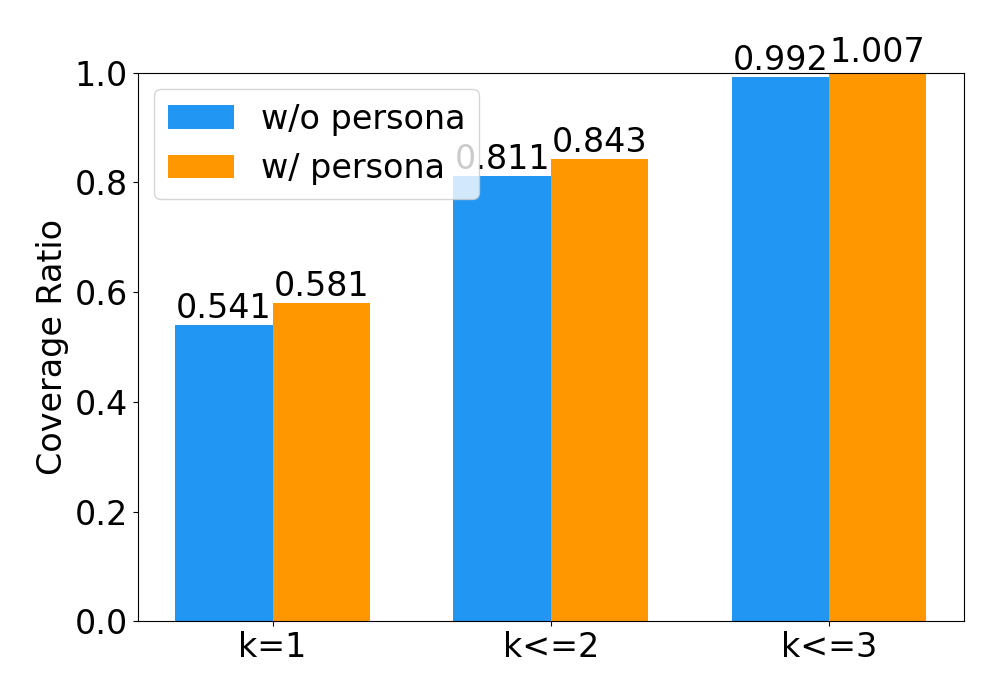}
    }
    \hfill
    \subfigure[search engine engineers]{
        \includegraphics[width=0.30\textwidth]{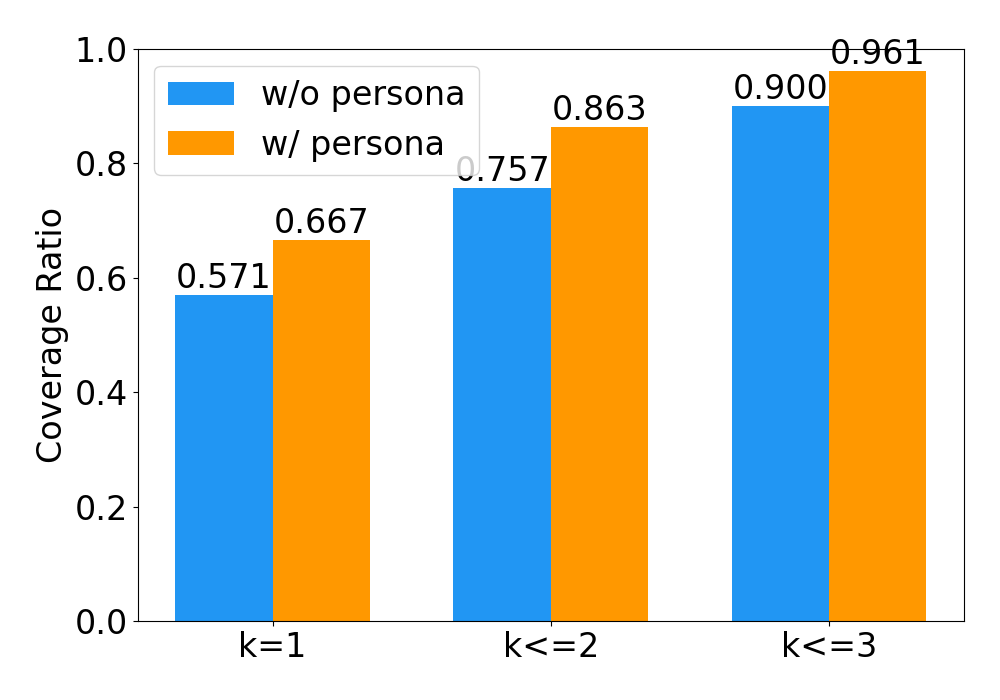}
    }
    \hfill
    \subfigure[social media analysts]{
        \includegraphics[width=0.30\textwidth]{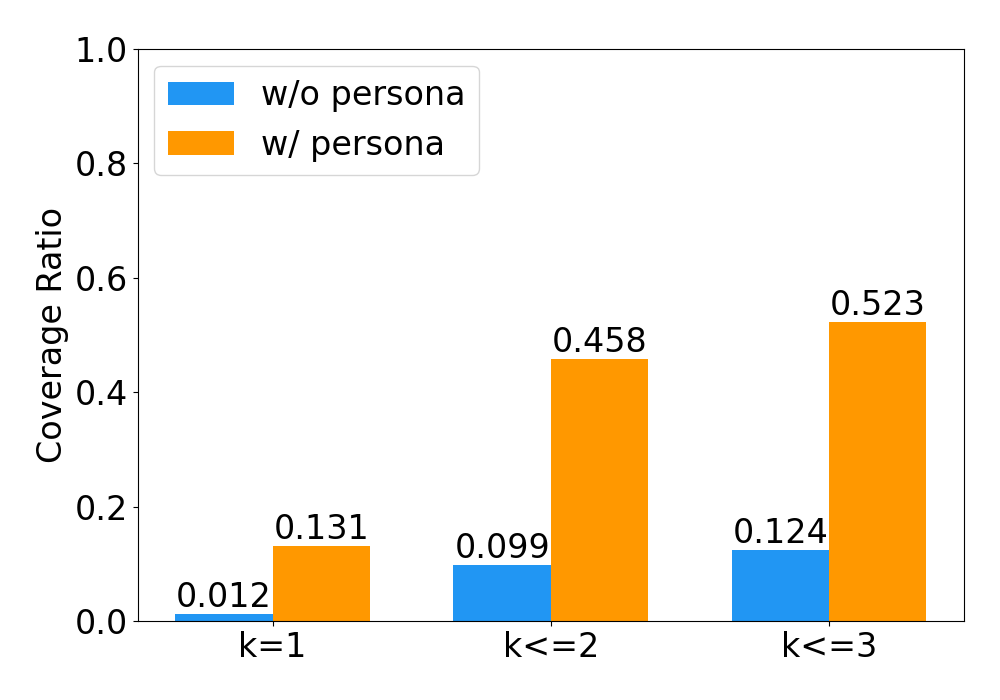}
    }
    \subfigure[software developers]{
        \includegraphics[width=0.30\textwidth]{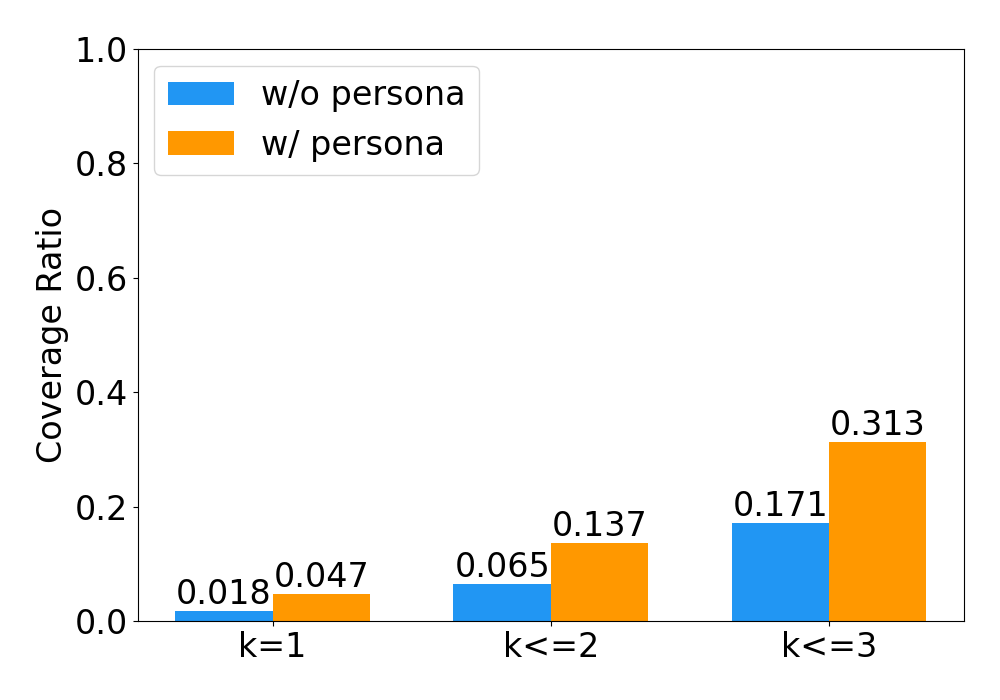}
    }
    \hfill
    \subfigure[students]{
        \includegraphics[width=0.30\textwidth]{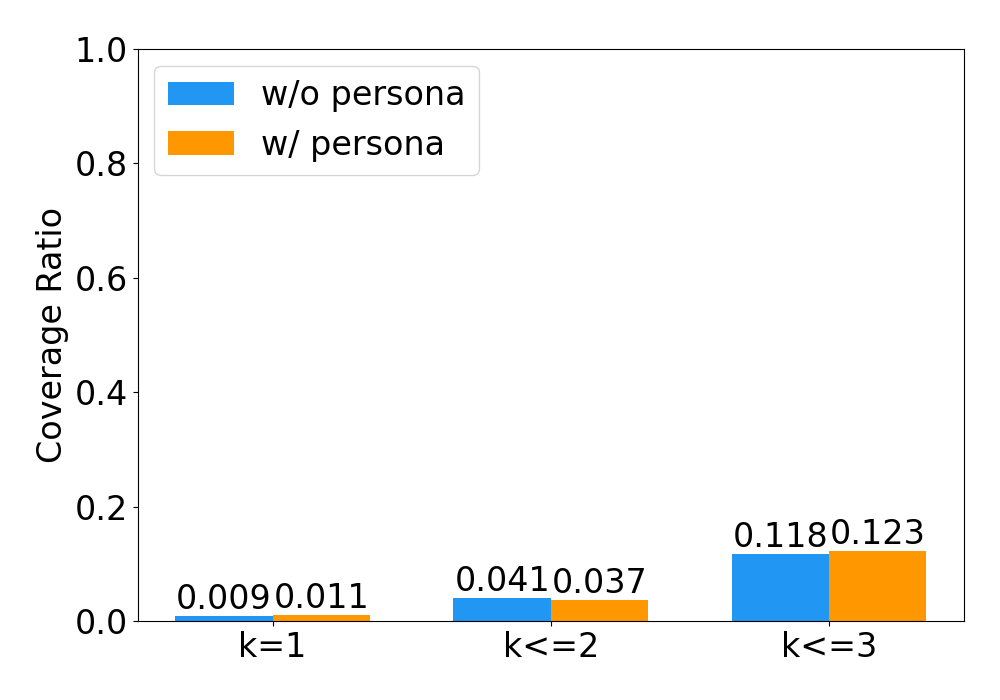}
    }
    \hfill
    \subfigure[voice assistant developers]{
        \includegraphics[width=0.30\textwidth]{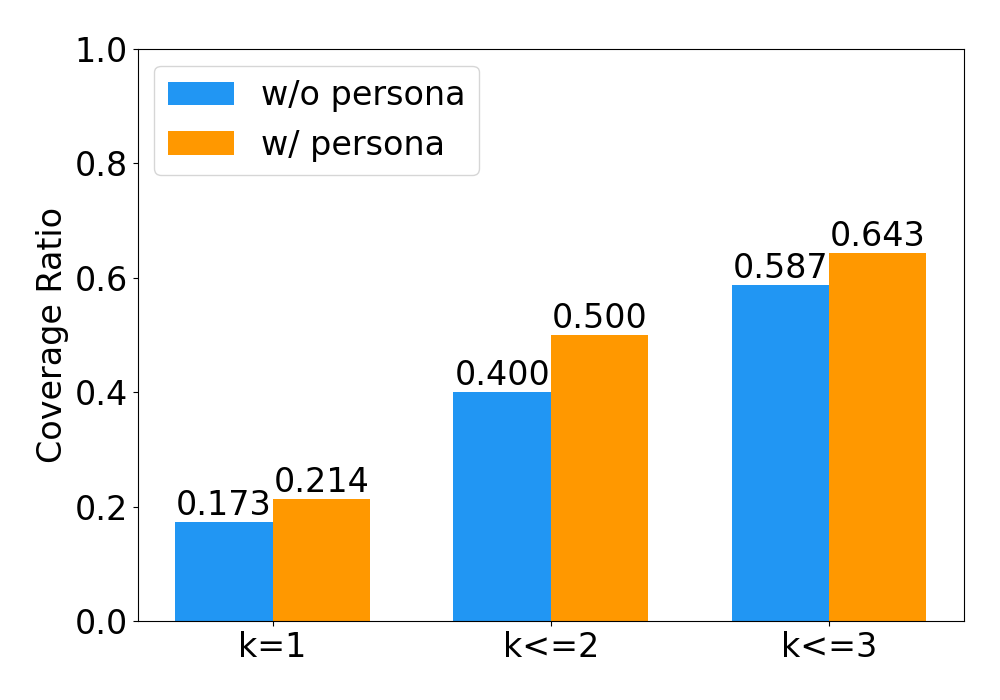}
    }
    \vspace{-5pt}
    \caption{The coverage ratio of 15 examples personas in the \textbf{academia} domain.}
    \label{fig-app:cover-ratio-academia}
\end{figure*}

We display more examples for the coverage ratio from the three domains of legal, finance, and academia in Figure \ref{fig-app:cover-ratio-legal}, Figure \ref{fig-app:cover-ratio-finance}, and Figure \ref{fig-app:cover-ratio-academia}.

\section{Auto-evaluation prompt}
\label{app:eval-prompt}
Below is an example prompt for evaluating question answerability:

{\ttfamily Your job is to evaluate the quality of a question generated based on the text of a document. The purpose of the question is to serve as a "suggested question" next to the document in a "smart" document reader software, in order to help the reader (user of the document reader software) better navigate the document and provide the reader a better reading experience.

Your job is to determine whether you believe the suggested question can be answered from the information contained in the document. Higher answerability means that the question can be directly answered based on the content available in the document.

You will reply with one of the following options : 'Strongly Disagree', 'Disagree', 'Undecided', 'Agree', 'Strongly Agree'.

For example, given the question below:

Question: \{sample\_question\}

If I were asked whether this question is answerable, I would reason as follows:
1. Reasoning : \{sample\_reasoning\}.
2. Answer : \{sample\_answer\}

Below is the text of a document the reader is reading:
\{document\}

Below is the question:
\{question\}

Read the document's content and then think step by step about whether the question can be answered based on the document's content. Then make an evaluation decision based on your reasoning.

You must format your response as follows:
1. Reasoning: [Your reasoning here]
2. Answer: [choose one of 'Strongly Disagree', 'Disagree', 'Undecided', 'Agree', 'Strongly Agree']}

The above prompt for evaluating answerability is different from the prompt in Table \ref{tab:prompt-question-check-answerability} directly returning the answer or None if not answerable, which is used for \textbf{filtering out} invalid questions. 
For other evaluation metrics, we can simply insert a different metric definition in the second paragraph, and use a different sample question, rationale, and answer as the guilding in-context example. Please note that this prompt is for \textbf{evaluating} the questions by giving scores. 

\section{Disk space considerations for on-device model deployment}
\label{app:on-device}
In contrast to the operating system that has the capacity to accept models with billions of parameters and gigabytes of disk space, an AI-powered reading application cannot, because a billion-parameter model within an application will not only significantly increase the application's download and the installer's size but also likely strain the device's memory and battery capacity when running such model in the application on top of the operating system. Models of hundreds of millions of parameters are an ideal choice because they can be quantized to take only several hundreds of megabytes, making them a more suitable option for deploying within an application.

\section{Qualitative examples}
Table~\ref{tab:examples} shows a few examples comparing our Persona-SQ fine-tuned model with SQuAD fine-tuned model and GPT4o prompting.

\begin{table*}[t!]
\caption{Qualitative results comparing the generated SQs among GPT4o, SQuAD baseline model, and Persona-SQ fine-tuned model.}
\label{tab:examples}
\resizebox{\linewidth}{!}{%
\begin{tabular}{@{}lp{4cm}p{4cm}p{4cm}p{4cm}@{}}
\toprule
\multirow{2}{*}{\bf Document link} & \multirow{2}{*}{\bf Document overview} & \multicolumn{3}{c}{\bf Generated questions (randomly selected 3)} \\ \cmidrule(l){3-5} 
 &  & \multicolumn{1}{c}{\bf SQuAD fine-tuned} & \multicolumn{1}{c}{\bf GPT4o} & \multicolumn{1}{c}{\bf Persona-SQ fine-tuned} \\ \midrule
\href{https://media.focusonthefamily.com/fotf/pdf/channels/marriage/marriage-and-conflict.pdf}{Document 1} & The document is about marriage and turning disagreement into growth, providing advice on conflict resolution in relationships. & 
    \begin{tabular}[t]{@{}p{4cm}@{}}What did the father give up to see if the boy would respond? 

    \vspace{5pt}
    
    What did Paul say regarding the issue of the dead? 

    \vspace{5pt}
    
    Who is the protagonist of the book and what is it's occupation?\end{tabular} & 
    \begin{tabular}[t]{@{}p{4cm}@{}}Who is the booklet dedicated to? 

    \vspace{5pt}
    
    How should disagreements be dealt with privately? 

    \vspace{5pt}
    
    What is the importance of using 'I' statements during conflicts?\end{tabular} & 
    \begin{tabular}[t]{@{}p{4cm}@{}}How to avoid getting stuck in unresolved conflicts? 

    \vspace{5pt}
    
    What are some healthy ways to express feelings? 

    \vspace{5pt}
    
    What are the effects of unresolved conflicts on marriage?\end{tabular} \\ \midrule
\href{https://springhillpremierchiro.com/wp-content/uploads/2019/01/Adult-Patient-Form.pdf}{Document 2} & This document is a Premier Chiropractic Adult Health History Form for patients to fill out before their appointment. & 
    \begin{tabular}[t]{@{}p{4cm}@{}}What does it mean to "ackchie" under HIPAA? 

    \vspace{5pt}
    
    How can you draw a line to represent your pain or discomfort? 

    \vspace{5pt}
    
    What is the most common way that chiropractic care is performed?\end{tabular} & 
    \begin{tabular}[t]{@{}p{4cm}@{}}What is the fee for copying X-Rays on a disc? 

    \vspace{5pt}
    
    How are spinal health problems identified in chiropractic? 

    \vspace{5pt}
    
    What risks are associated with chiropractic care?\end{tabular} & 
    \begin{tabular}[t]{@{}p{4cm}@{}}How does chiropractic care treat herniated discs? 

    \vspace{5pt}
    
    How long does the adjustment procedure take? 

    \vspace{5pt}
    
    What is my insurance coverage for chiropractic care?\end{tabular} \\ \midrule
\href{https://www.evelyntribole.com/wp-content/uploads/Celiac-Disease-Personal-Nutrition.LAT.02.pdf}{Document 3} & The document discusses the importance of choosing a qualified nutritionist for personalized nutrition advice and the potential risks of receiving advice from unqualified individuals in California. & 
    \begin{tabular}[t]{@{}p{4cm}@{}}What did Tribole help Barnes with? 

    \vspace{5pt}
    
    Who was the former professional football player who became discouraged after being forced to pack 290 pounds? 

    \vspace{5pt}
    
    How common is it for personal nutrition advice to be unreliable?\end{tabular} & 
    \begin{tabular}[t]{@{}p{4cm}@{}}Who monitors questionable nutrition practices in California?

    \vspace{5pt}
    
    What is the role of registered dietitians in nutrition?

    \vspace{5pt}
    
    How did a nutritionist help a woman with celiac disease?\end{tabular} & 
    \begin{tabular}[t]{@{}p{4cm}@{}}What are some common myths about nutrition?

    \vspace{5pt}
    
    What motivates clients to make healthy choices? 

    \vspace{5pt}
    
    What qualifications are necessary to practice nutrition therapy?\end{tabular} \\ \bottomrule
\end{tabular}
}
\end{table*}

\begin{table*}[h!]
\centering
\caption{Main results of various fine-tuned SmolLM 360M comparing to SQs generated by GPT3.5 Turbo. The model fine-tuned on Persona-SQ generated dataset achieves the best performace among all model variants and is the first best tiny model to match the performance of GPT3.5 Turbo.}
\vspace{-5pt}
\label{tab:results}
\resizebox{0.8\linewidth}{!}{%
\begin{tabular}{@{}lcccc@{}}
\toprule
\textbf{Model}    & \textbf{Win}  & \textbf{Tie}  & \textbf{Lose} & \textbf{avg win+match rate} \\ \midrule
SQuAD model       & 12.67 (3.06)  & 36.33 (7.02)  & 147 (5.29) & 25\%  \\ 
non-filtered, non-persona model & 95.33 (5.03) & 8.33 (2.08) &	98.33 (3.21) & 51.32\%  \\
non-filtered, persona model & 104.33 (6.11) &	10 (3.46) &	87.67 (4.16) & 56.60\% \\
filtered, non-persona model & 106.33 (2.52) & 8.67 (4.04)   & 87 (3.46)  & 56.93\%  \\\midrule
{\bf TinyDocLM-SQ}      & {\bf 107.67 (6.43)} & 10.67  (1.53) & 83.67 (7.51) & {\bf 58.58\%} \\
\bottomrule
\end{tabular}%
}
\end{table*}

\section{Model details}
\label{app:model-details}
To construct the fine-tuning dataset, we assemble each data point in the synthetic training data into the chat format consisting of a ``user'' turn and an ``assistant'' turn compatible with the instruction fine-tuning input style. For the Persona-SQ dataset, the user turn is the following: \texttt{Please read the document below and then do the following: 1) make some predictions about the reader who is likely to read it, including the reader's profession, the reader's intent of reading this document, and what this reader might already know related to this document; and  2) generate a guiding question such that the answer to this question will be interesting and informative to the reader you just predicted. \#\#\#Document:\{document\}}. The assistant's turn is formatted as follows: \texttt{\#\#\#Reader profile: \{persona\} \#\#\#Question: \{question\}}.
For the non-persona SQ dataset, the user turn is formatted as follows:
\texttt{Please read the document below and then generate a guiding question such that the answer to this question will be interesting and informative to the reader who is reading this document. \#\#\#Document: \{document\}}. And the assistant's turn is formatted as follows: \texttt{\#\#\#Question: \{question\}}. For dealing with long documents, we split it into chunks of 1500 tokens with 200 tokens overlap as the ``document'' input in the above prompts. documents with less than 500 tokens are discarded. We use the chunked documents as input for Persona-SQ to generate synthetic training data as well as for fine-tuning. We leave extending the model to handle longer context for future research. For the SQuAD baseline, we use the non-persona chat format to assemble the document and questions into the non-persona SQ training data.

For all experiments, we fine-tune for one epoch, with 4 A100 80G GPUs, batch size of 4 per device, gradient accumulation step of 1, and learning rate of $1\times10^{-5}$.

\section{Synthetic training data statistics}
\begin{figure}
    \centering
    \includegraphics[width=1\linewidth]{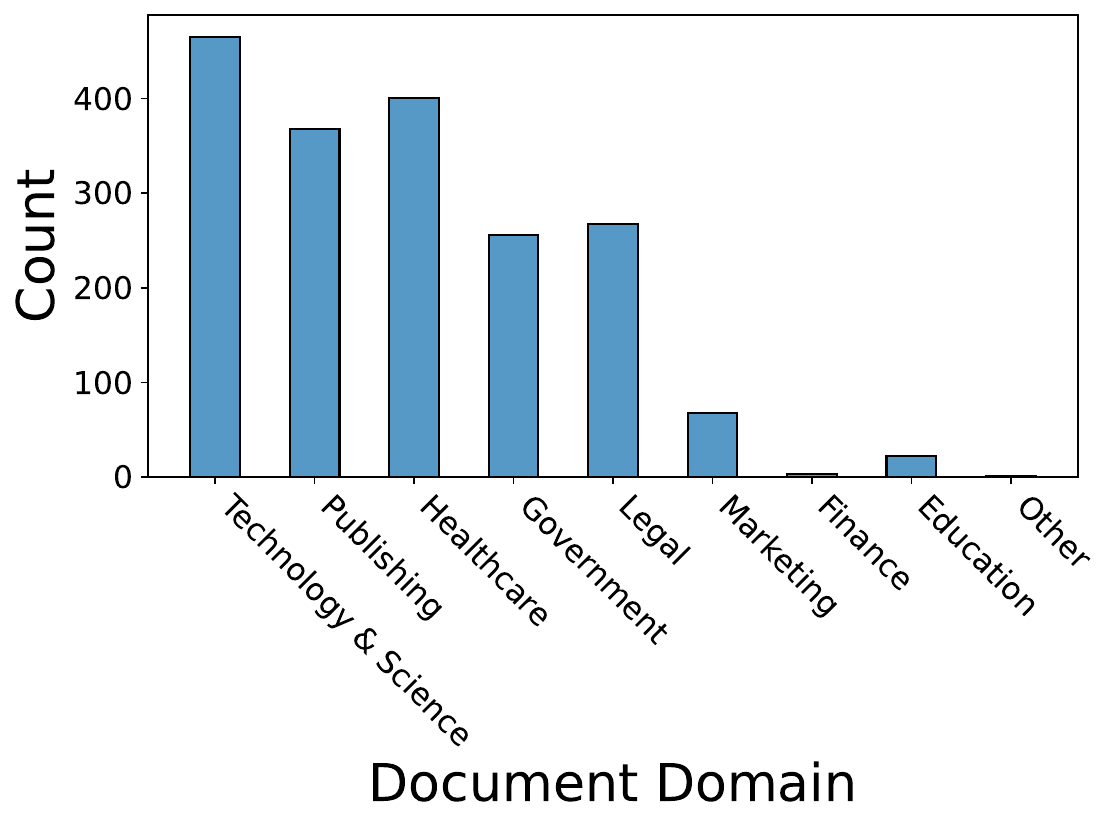}
    \caption{Number of documents according to the document domains, in the Persona-SQ synthetic fine-tuning dataset.}
    \label{fig:doc-domain-count}
\end{figure}

\begin{figure}
    \centering
    \includegraphics[width=\linewidth]{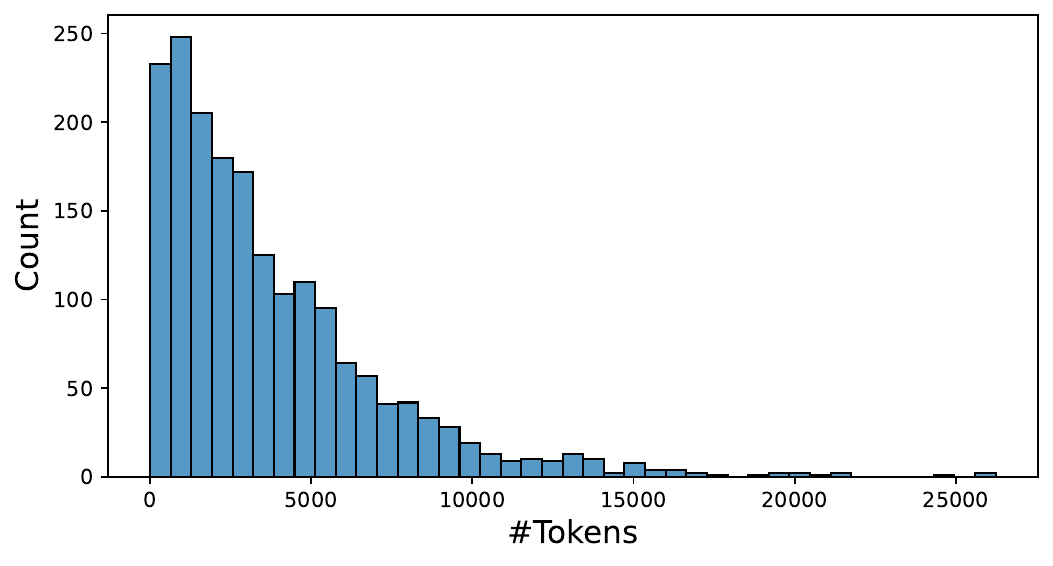}
    \caption{Distribution of token counts for all documents, in the Persona-SQ synthetic fine-tuning dataset.}
    \label{fig:doc-token-count}
\end{figure}

Table~\ref{tab:data-stat} and figures~\ref{fig:doc-domain-count} and~\ref{fig:doc-token-count} shows the statistics of the document domains, document toke counts, and the number of questions generated per domain. In total, we synthetically generated about 23k questions from around 1600 documents across a variety of professional documents.

\section{Additional results on Persona-SQ fine-tuned models}
\label{app:additional-smollm-results}
\subsection{Automatic evaluations}
We additionally conduct an automatic evaluation by comparing SQs generated by the fine-tuned models with those generated by GPT3.5 Turbo. We present the two sets of three questions, one set from one of our fine-tuned models and the other set from GPT3.5 Turbo, along with the document from which the questions are generated, to an evaluator, who judges which set is better, or if both sets are equally good or bad. 
The evaluation criteria emphasizes the naturalness and attractiveness of these questions when users see them at the very beginning of reading a document. 
In practice, we use GPT4o for this evaluation task.
We compute a ``win/tie rate'', i.e., the proportion of documents which the fine-tuned model is judged to be either better than GPT3.5 Turbo, equally good, or equally bad.
Table~\ref{tab:results} shows win/tie rate using GPT4o as the evaluator. The model fine-tuned on the dataset synthesized by Persona-SQ achieves highly competitive performance against GPT3.5 Turbo. Comparisons among models fine-tuned on other datasets demonstrate 1) the usefulness of quality filters and persona in producing a higher quality dataset and thus a better performing model, and 2) public QA dataset, when used as fine-tuning dataset for SQ generation, is undesirable for real-world documents.

\subsection{Qualitative examples}
Table~\ref{tab:examples} displays a few qualitative examples comparing the generated SQs comparing various models, showcasing that our Persona-SQ fine-tuned model's outstanding performance compared to other fine-tuned models and its competitiveness against prompting much larger models.

\section{Human evaluation procedure}
\begin{figure*}
    \centering
\includegraphics[width=0.7\linewidth]{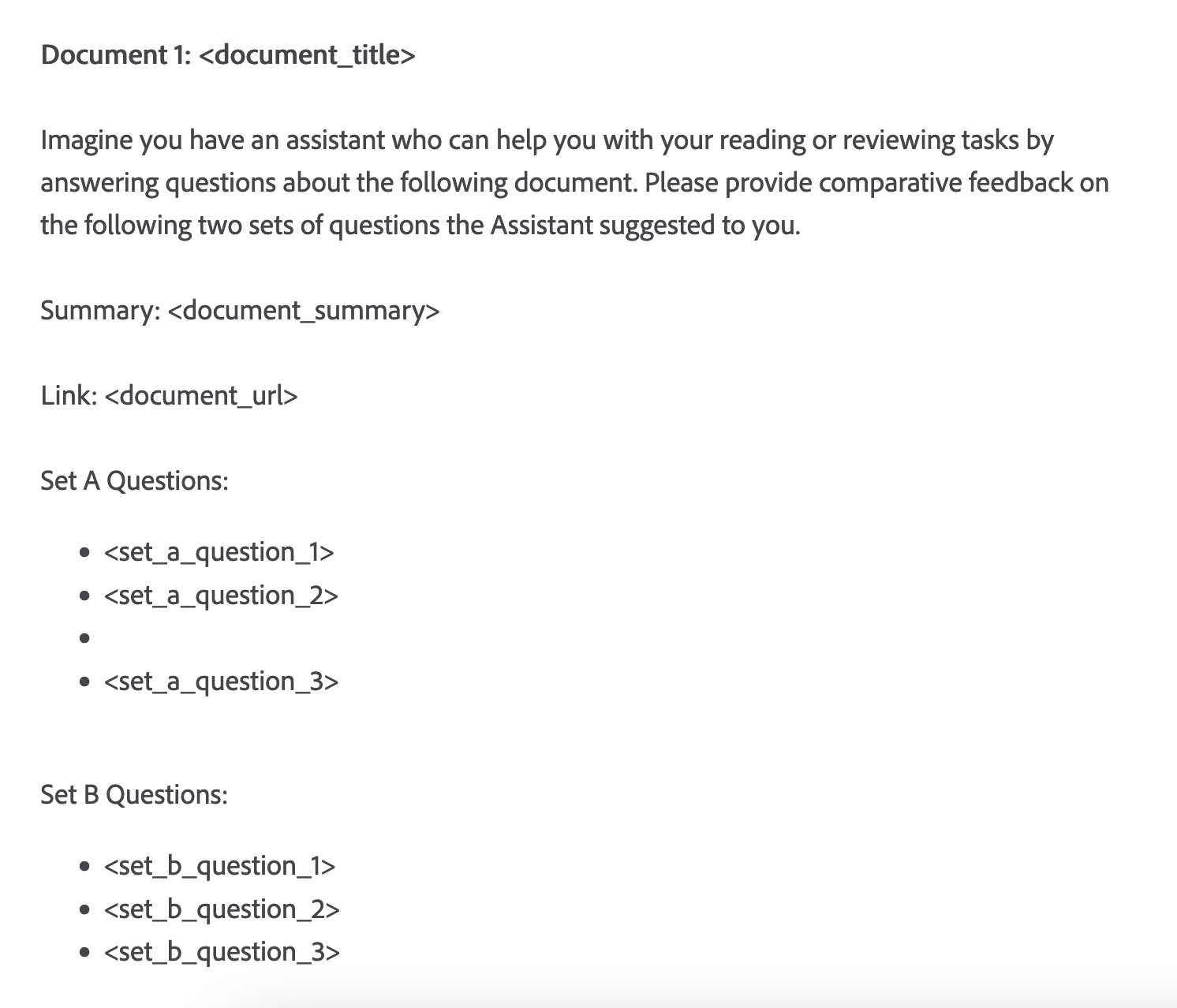}
    \caption{A screenshot of the user evaluation survey, where the document title, document summary (automatically generated), the full document link, and the two sets of questions are shown to the user. Content in brackets are placeholders for the actual content.}
    \label{fig:human-interface-1}
\end{figure*}

\begin{figure*}
    \centering
\includegraphics[width=0.7\linewidth]{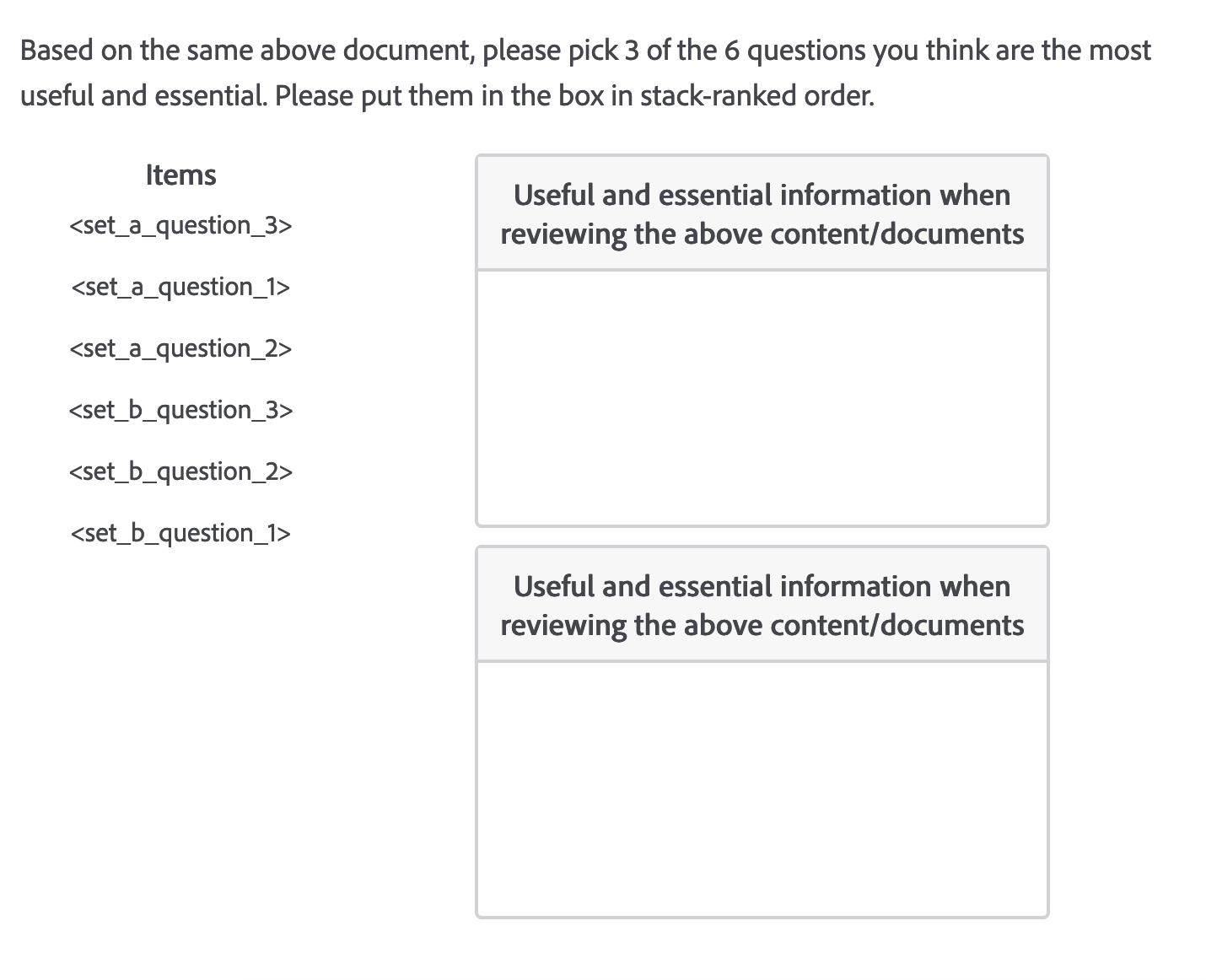}
    \caption{A screenshot of the user evaluation survey, where we ask the evaluators to rank, via drag and drop, the questions in terms of their preference. The order of the questions presented to the evaluator is randomized by Qualtrics.}
    \label{fig:human-interface-3}
\end{figure*}

We conduct our human evaluation on on Prolific. We design the survey using Qualtrics; an example is shown in Figure~\ref{fig:human-interface-1} and~\ref{fig:human-interface-3}. We show the evaluator the task, the document title, document summary, and document URL from which the full original document can be accessed. Then we present the evaluators two sets of three questions. The evaluator will rate the quality along several axes and then rank the questions in terms of preference. When ranking for preference, the question ordering is randomized.

\begin{table*}[t!]
    \centering
    \caption{The statistics of the documents and persona-based SQs.}
    \vspace{-5pt}
    \resizebox{2\columnwidth}{!}
    {
    \begin{tabular}{c|ccccccc}
        \toprule
        Domain & Subdomain & Dataset & \#. Doc & \makecell[c]{Avg. \\Length} & \#. Persona & \makecell[c]{\#. Gen. \\ Question} & \makecell[c]{\#. Gen. Ques. after \\ Quality Control} \\
        \midrule
        Finance & Annual Report & Fns2020 \cite{el-haj-etal-2020-financial} & 50 & 42583 & 68 & 9214 & 7621 \\
        Legal & Contract & CUAD \cite{hendrycks2021cuad} & 100 & 9622 & 73 & 12902 & 9262 \\
        academia & Paper & qsper \cite{dasigi-etal-2021-dataset} & 100 & 4355 & 41 & 12311 & 7708 \\
        \bottomrule
    \end{tabular}
    }
    \label{tab:stats-dataset}
\end{table*}

\begin{table*}[h!]
\centering
\caption{Statistics of the synthetic Persona-SQ fine-tuning dataset. Note that some documents do not have the vertical tag; in those cases, we use GPT4o to give the document a tag and include the LLM-produced tags in the statistics computation. Some verticals do not belong to the seven major verticals, which we group them together into the ``Unknown'' vertical.}
\label{tab:data-stat}
\begin{tabular}{@{}lccc@{}}
\toprule
\textbf{vertical} & \textbf{\#documents} & \textbf{avg \#questions per document} & \textbf{median \#words in question} \\ \midrule
Publishing        & 334                  & 16                                    & 7                                   \\
Healthcare        & 374                  & 14                                    & 7                                   \\
Research          & 308                  & 14                                    & 8                                   \\
Legal             & 244                  & 10                                    & 8                                   \\
Government        & 195                  & 12                                    & 7                                   \\
Marketing         & 73                   & 18                                    & 7                                   \\
Science           & 111                  & 11                                    & 8                                   \\
Unknown           & 25                   & 9                                     & 7                                   \\ \bottomrule
\end{tabular}
\end{table*}

\label{app:train-data-stat}

\begin{table*}[!ht]
    \centering
    \caption{Prompts for generating personas and questions. }
    \resizebox{\textwidth}{!}
    {
    \begin{tabular}{l|m{14cm}}
    \toprule
        \textbf{Step} & \textbf{Prompt} \\ \midrule
         Generate Personas & \quad In some professional setting, for some document domains, people with different backgrounds would read them with very different purposes/goals and ask very different questions.

        \quad Your job is to predict what profession would read this document, and what goals they want to achieve. 
        
        \quad The goals should be closely related to the profession. Your prediction should try to be various. The statement describing the goal can be either first-person or a general declarative sentence.

        \quad You should think step by step and try your best to be creative. One profession can have different number of goals. The goals should be very diverse but related to the corresponding profession.

        \quad The profession can also be non-professional.

        \quad The following is a document from \$DOMAIN\$ \$SUBDOMAIN\$:

        \quad \$DOCUMENT CONTENT\$.
        
        \quad You should generate output in the following JSON format, for example:
        
       \{
       
            \quad "domain": \{
            
            \quad \quad "subdomain": \{
                
            \quad \quad \quad "profession": ["goal 1.", "goal 2."]
            
        \quad \quad \}
        
            \quad\}
            
        \}
        
        According to the document from the domain \$DOMAIN\$ \$SUBDOMAIN\$, your answer is:
         \\
        \midrule
        Normalize Personas & 
        \quad You are an AI helper to help users to classify professions into different groups.

        \quad The professions are as follows: \$PERSONAS\$.
        
        \quad You should return in a JSON format. The key is profession and the value is a list of given professions. For example:

        \{
        
            \quad "Accountants": ["Accountants", "Financial Accountants"],
            
            \quad "Auditors": ["Auditors", "auditors"]
            
        \}

        Based on the given professions, your answer for the groups of personas is: \\
        \midrule
        Generate Questions & \quad You are a PDF Reader AI Assistant. You will be given a long PDF document, a user profession, and several goals of the user. Your task is to generate a series of questions that users with the specified profession and goals might be interested in. 

        \quad The user's profession and goals are provided below:
        
        \quad **Profession:** \$PROFESSION\$
        
        \quad **Goals:** \$GOALS\$
        
        \quad Please generate questions that meet the following criteria:
        
        \quad 1. **Personalized:** The questions should align with the user's interests and profession.
        
        \quad 2. **Logical:** The questions should follow a logical order.
        
        \quad 3. **Comprehensive:** The questions should cover as much useful information as possible to ensure the user can achieve their goals.
        
        \quad Output the questions in a JSON format. For example:
        
        \{
        
            \quad "Question 1": "xxx",
            
            \quad "Question 2": "xxx",
        
        \}
        
        \quad Ensure that the output is in a JSON format without any additional text or errors.
        
        \quad Ensure that generate a series of questions as various as possible.
        
        \quad The following is the document:
        
        \quad \$DOCUMENT CONTENT\$
        
        \quad The generated questions are: \\
        \bottomrule
    \end{tabular}
    }
    \label{tab:prompt-generation}
\end{table*}

\begin{table*}[!ht]
    \centering
    \caption{Prompts for persona quality control. }
    \resizebox{\textwidth}{!}
    {
    \begin{tabular}{l|m{14cm}}
    \toprule
        \textbf{Step} & \textbf{Prompt} \\ \midrule
         Goals Quality & \quad You are an AI assistant to help user to finish the task. You will be provided with one persona, and many goals candidates corresponding to the persona. The goals are the purposes of a user want to achieve by reading a document.

        \quad Your job is to score the goals based on the consistency between the goals, persona and the domain of the document. 
        
        \quad Provide your rating on a scale from 1 to 5 based on the criteria below:
        
        \quad - **Rating 1**: The goal quality is extremely poor. The generated goal is not described in a valid format with ovbious grammar error or it is not a goal but a question or something else.
        
        \quad - **Rating 2**: The goal quality is somewhat poor. The generated goal is in a valid format but it is totally unrelated to the persona or the document domain.
        
        \quad - **Rating 3**: The goal quality is good. The generated goal is related to both the document and the persona, but the connection is not very strong. The goal is somewhat meaningful. Sometimes, the persona might want to achieve the goal but sometimes not.
        
        \quad - **Rating 4**: The goal quality is very good. The generated goal is closely related to both the document and the target persona. For most cases, the persona may have the goal when they read the document.
        
        \quad - **Rating 5**: The goal quality is excellent. The generated question is highly relevant to both the document and the target persona. The persona always have the goal when they read the document.
        
        \quad Here is the persona: \$PERSONA\$
        
        \quad Here are the goals that are separated by ";":
        
        \quad \$GOALS\$
        
        \quad You should return in a JSON format. The key is the repeat of the goal, and the value is the score. For example:
        
        \{
            \quad \quad "I want to understand the document in details.": 5
        \}
        
        \quad Based on the provided persona and goals, your scores for the goals are:
         \\
        \bottomrule
    \end{tabular}
    }
    \label{tab:prompt-persona-quality-control}
\end{table*}

\begin{table*}[!ht]
    \centering
    \resizebox{\textwidth}{!}
    {
    \begin{tabular}{l|m{14cm}}
    \toprule
        \textbf{Step} & \textbf{Prompt} \\ \midrule
        Score Questions & 
        \quad You will be given a long document, a target persona with specific goals, and several questions that the target persona might ask. Your task is to evaluate the quality of these generated questions based on the document and the target persona's goals.

        \quad Here is the document: \$DOCUMENT\$
        
        \quad In this task, you need to evaluate the quality of the generated questions based on the document and the persona's goals. The quality of the generated questions depends on how meaningful, valuable, and relevant they are to the document and persona's goals.
        
        \quad Provide your rating on a scale from 1 to 5 based on the criteria below:
        
        \quad - **Rating 1**: The question quality is extremely poor. The generated question is completely unrelated to the document and persona's goals.
        
        \quad - **Rating 2**: The question quality is somewhat poor. The generated question is related only to the document or only to persona, but not both. The question may also be meaningless in helping persona achieve their goals.
        
        \quad - **Rating 3**: The question quality is good. The generated question is related to both the document and the target persona, but the connection is not very strong. The question is somewhat meaningful and can help the persona partially achieve one of their goals. The persona might ask the question, but not always.
        
        \quad - **Rating 4**: The question quality is very good. The generated question is closely related to both the document and the target persona. However, compared to the target persona, the question is more likely to be asked by one of OTHER PERSONAS.

        \quad - **Rating 5**: The question quality is excellent. The generated question is highly relevant to both the document and the target persona. The persona will definitely ask the question about the reference document. Compared to "OTHER PERSONAS", the question is more likely to be asked by the target persona.
        
        \quad For each question, conduct the evaluation as described above. If you provide score of 4, also reply which "other persona" is more likely to ask the question compared to the target persona; if you provide other scores, reply none for this. Your response should be in JSON format, with the question as the key and the score with other persona as the value.
        
        \quad Here is the target persona: \$PERSONA\$.
        
        \quad Here are the goals of the target persona: \$GOALS\$.
        
        \quad Here are the generated questions separated by semicolons: \$QUESTIONS\$.
        
        \quad Here are OTHER PERSONAS: \$OTHER\_PERSONA\$.
        
        \quad Ensure that the key is an exact copy of the question and the score is between 1 and 5. Ensure the output follows a VALID JSON FORMAT!
        
        \quad Given the example questions: "Question A?; Question B?", the example output is:
        
        ```json
        \quad {
        
            \quad \quad "Question A?": [4, "other\_persona"],
            
            \quad \quad "Question B?": [3, "None"]
            
        \quad }
        ```
        
        \quad The score you give for each question is: \\
        
        \bottomrule
    \end{tabular}
    }
    \caption{Prompts for question quality control. }
    \label{tab:prompt-question-quality-control}
\end{table*}

\begin{table*}[!ht]
    \centering
    \caption{Prompts for checking the answerability of generated questions. }
    \resizebox{\textwidth}{!}
    {
    \begin{tabular}{l|m{14cm}}
    \toprule
        \textbf{Step} & \textbf{Prompt} \\ \midrule
        Score Questions & 
        \quad You will be given a long document and several questions related to the document. Your task is to evaluate whether these questions can be answered based on the content of the document. 

        \quad Here is the document: \$DOCUMENT\$
        
        \quad For each question:
        
        \quad 1. If the document contains the answer, provide the answer and the exact reference text from the document. The answer should not be a direct copy from the original document. You should answer the question in your own words but refer to the document contents. The reference text should contain enough information to answer the question. If the reference texts contain different parts, concatenate every parts together.
        
        \quad 2. If the document does not contain the answer, return "None" for both the answer and the reference.
        
        \quad You will be given several questions to evaluate. Conduct the task described above for each question. Your response should be in JSON format, with each question as the key and the answer and reference as the values.
        
        \quad Ensure that the key is an exact copy of the question and the reference is an exact copy of a text span in the given document. Ensure the output follows a VALID JSON FORMAT!
        
        \quad Example of two questions (the first question is answerable, while the second one is not answerable):
        
        \quad **Questions:**
        
        \quad 1. Question 1?
        
        \quad 2. Question 2?
        
        \quad **Answers:**
        
        ```json
        
        \quad \{
        
            \quad \quad "Question 1?": { "Answer": "xxx", "Reference": "yyy" },
            
            \quad \quad "Question 2?": { "Answer": "None", "Reference": "None" },
            
        \quad \}
        
        ```
        
        \quad **Questions:**
        \quad \$QUESTIONS\$
        
        \quad **Answers:**\\
        
        \bottomrule
    \end{tabular}
    }
    \label{tab:prompt-question-check-answerability}
\end{table*}

\begin{table*}[!ht]
    \centering
    \caption{Prompts for predicting the related persona given on generated question. }
    \resizebox{\textwidth}{!}
    {
    \begin{tabular}{l|m{14cm}}
    \toprule
        \textbf{Step} & \textbf{Prompt} \\ \midrule
        Predict related personas & 
        \quad You will be given a summary of a document, one question and several personas. Your task is to conduct a multiple choice to choose the personas that might be interested in the given question that is related to the document. You should respond in a JSON format.

\quad Here is an example. In  this example, four personas are given to you, and the persona3 is the most one to be interested in the question, while the persona2 is the second one. Persona1 and persona4 are not interested in the question. Example of the INPUT and OUTPUT:

\quad **INPUT**:

\quad **Document**: \quad Document content.

\quad **Question**: \quad Question?

\quad **Personas**: \quad Persona1, persona2, persona3, persona4.

\quad **OUTPUT**:

```json

\quad \{

    \quad \quad "order 1": "persona3,
    
    \quad \quad "order 2": "persona2"
    
\quad \}

```

\quad **INPUT**:

\quad **Document**: \quad \$DOCUMENT\$

\quad **Question**: \quad \$QUESTION\$

\quad **Personas** \quad \$PERSONA\$

\quad **OUTPUT**\\
        
        \bottomrule
    \end{tabular}
    }
    \label{tab:prompt-question-check-answerability}
\end{table*}

\section{More details on the demonstrations}
For the model demonstration, to speed up the generation from GPT4o and save costs, we prompt it with the same prompt as the Persona-SQ fine-tuned model.

\end{document}